\documentclass[11pt]{article}
\usepackage[utf8]{inputenc}
\DeclareUnicodeCharacter{02BB}{'}

\usepackage[preprint]{acl}

\usepackage{times}
\usepackage{latexsym}

\usepackage[T1]{fontenc}

\usepackage[utf8]{inputenc}

\usepackage{microtype}

\usepackage{inconsolata}

\usepackage{graphicx}
\usepackage{booktabs}
\usepackage{soul}
\usepackage{xcolor}
\usepackage{makecell}
\usepackage{adjustbox}
\usepackage{float}
\usepackage{multirow} 
 \usepackage{pgfplots}
 \usepackage{tikz}
\usetikzlibrary{arrows.meta}
\pgfplotsset{compat=1.18}
\usepackage{subcaption}
\usepackage[table]{xcolor}
\usepackage{arydshln}
\usepackage{amsmath}
\usepackage{longtable}
\usepackage{listings}
\usepackage{xcolor}
\usepackage{xfp}   

\newcommand{\posneg}[1]{%
  \ifnum\fpeval{#1>0}=1
    {\color{green!60!black}+#1}%
  \else
    {\color{orange!80!black}#1}%
  \fi
}

\newcolumntype{C}{>{\centering\let\newline\\\arraybackslash\hspace{0pt}}m{1cm}}

\title{Multilingual and Cross-Lingual Citation Needed Detection on Wikipedia for Lower-Resource Languages}

\author{%
  Gerrit Quaremba$^{1}$, Amy Rechkemmer$^{1}$, Elizabeth Black$^{1}$, Denny Vrandečić$^{2}$, Elena Simperl$^{1}$ \\
  $^{1}$King's College London, 
  $^{2}$Wikimedia Foundation \\
  \texttt{\{gerrit.quaremba,amy.rechkemmer,elizabeth.black,elena.simperl\}@kcl.ac.uk} \\\texttt{denny@wikimedia.org}
}

\begin{document}
\maketitle

\begin{abstract}
In automated fact-checking (AFC), check-worthiness detection identifies claims requiring verification based on domain-specific criteria.
On Wikipedia, this task instantiates as Citation Needed Detection (CND), which flags claims lacking supporting citations.
However, existing research has largely overlooked lower-resource languages, and recent AFC pipelines rely on large language models (LLMs), which are inaccessible to low-resource organizations.
We introduce MCN, a multilingual CND corpus spanning 18 languages across three resource levels, on which we conduct an extensive study of small decoder-based language models (SLMs).
Our experiments show that SLMs fine-tuned with an encoder-style objective substantially outperform prompted LLMs across languages.
We further present one of the first studies on cross-lingual CND, demonstrating that SLMs fine-tuned solely on English claims surpass LLMs, even with little to no target-language adaptation.
Our findings have important implications for lower-resource Wikipedia communities and suggest that compact, task-specific models are preferable to LLMs for CND.
We release all data and code at \url{https://github.com/gerritq/mcn} 

\end{abstract}

\section{Introduction}

The surge of mis- and disinformation has increased the demand for fact-checking~\cite{vlachos2014}.
As manual fact-checking is laborious~\cite{nakov2021}, automated fact-checking (AFC) has emerged as a critical approach to support fact-checkers.
A crucial early-stage task in AFC is claim check-worthiness detection (CWD), identifying claims that require verification based on domain-specific criteria~\cite{guo2022}.

\begin{figure}[t]
    \centering
        \includegraphics[width=1\linewidth]{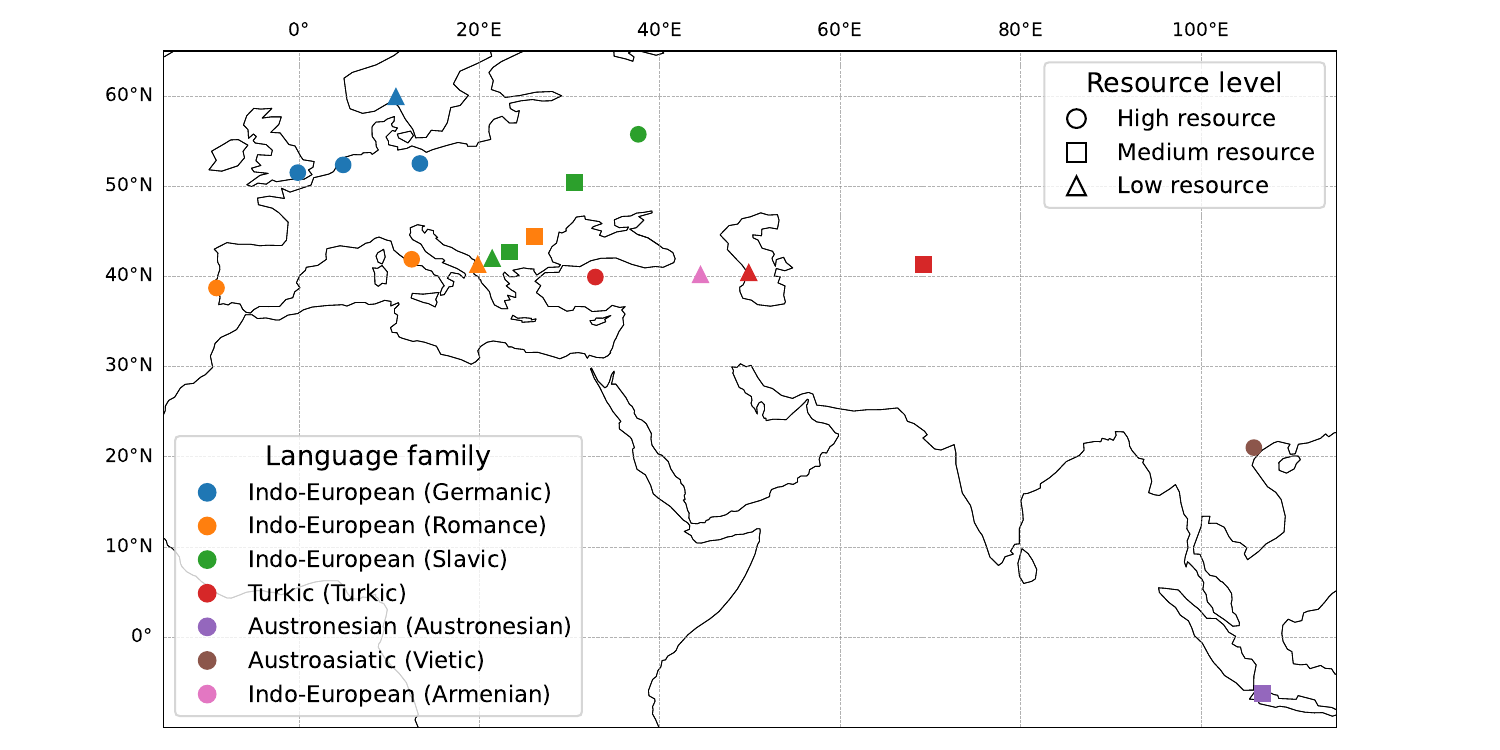}
    \caption{MCN Language Coverage.}
    \label{fig:world_map}
\end{figure}

\begin{figure}[t]
    \centering
        \includegraphics[width=1\linewidth]{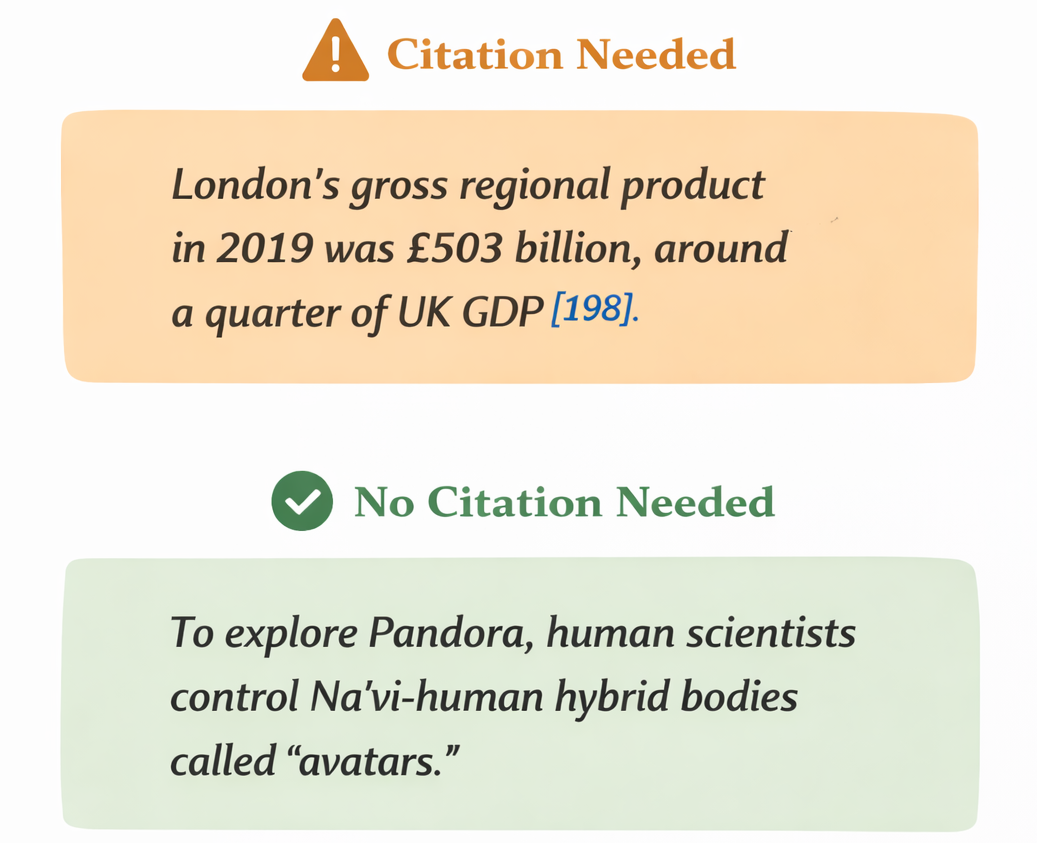}
    \caption{Dataset examples from the Wikipedia articles ``London'' and ``Avatar (2009 film)''. The first sentence requires a citation because it presents a statistical claim, whereas the second does not, as its source (the film itself) is evident to the reader.}
    \label{fig:example}
\end{figure}

On Wikipedia, CWD operationalizes as citation needed detection (CND) \citep{redi2019,guo2022}. 
We present examples of such claims in Figure~\ref{fig:example} and Appendix~\ref{app:data}.
Wikipedia editors collectively judge and determine the check-worthiness of claims by applying Wikipedia’s verifiability policy.\footnote{The \href{https://en.wikipedia.org/wiki/Wikipedia:Verifiability}{verifiability policy} states that \textit{all claims that are challenged or likely to be challenged must be supported by reliable sources}. We treat each sentence as a claim.}
When editors encounter claims that violate the policy, they tag them with the \textsc{Citation Needed} template to signal the need for an authoritative citation.
This maintenance work is critical for preserving Wikipedia’s knowledge integrity, which has made Wikipedia an essential resource for the artificial intelligence community~\cite{johnson2024}.
Yet, CND remains a largely manual task, motivating research that frames it as a supervised binary classification problem~\cite{redi2019,halitaj2024,wmf_mcd}.


However, existing work has primarily studied CND in high-resource languages.\footnote{We adopt the language resource taxonomy of~\citet{joshi2020}.}
This is a critical limitation, as lower-resource language communities typically have fewer active editors and therefore face a disproportionate burden in tagging claims that require citations~\citep{lewoniewski2017}.
Moreover, recent AFC pipelines have often relied on large language models (LLMs) for claim detection \citep[e.g.,][]{li2024self,wang2024}.
Their proprietary nature and high inference costs, however, make LLMs inaccessible to low-resource organizations~\citep{warren2025}.

In this work, we address these gaps by introducing \textbf{M}ultilingual \textbf{C}itation \textbf{N}eeded (MCN), a large-scale CND dataset derived from high-quality Wikipedia articles covering 18 linguistically diverse languages across multiple resource levels~\cite{joshi2020} (Figure~\ref{fig:world_map}).
MCN advances prior work by extending coverage to lower-resource languages and by substantially increasing scale (Table~\ref{tab:data_comp}).
On these data, we study small decoder-based language models (SLMs)\footnote{LM taxonomy is not mutually exclusive~\cite{minaee2025}. We refer to encoder-based pretrained language models (e.g., BERT-based models) as \textit{PLMs}, to decoder-based language models with fewer than 10B parameters as \textit{SLMs}~\cite{belcak2025}, and to larger decoder-based models as \textit{LLMs}.} as cost-effective, open-source alternatives to LLMs for CND. 
To this end, we specialise SLMs for CND through various supervised fine-tuning (SFT)~\cite{wei2021} variants with QLoRA~\cite{dettmers2023}, to meet low-resource constraints.

We structure our study around three research questions for multilingual CND.
First, initial work demonstrates strong performance of SLMs for English CWD~\cite{majer2024,li2024,bell2025}.
However, their effectiveness in lower-resource languages remains unexplored, and existing work relies on standard SFT.
This motivates:
(\textbf{RQ1}) \textit{How do different SFT variants affect SLM performance across languages?}
Second, recent AFC work relies predominantly on LLMs~\cite{dmonte2024}, despite their prohibitive cost for low-resource organizations.
We assess whether smaller models offer a viable alternative:
(\textbf{RQ2}) \textit{How do fine-tuned SLMs compare to prompted LLMs and encoder-based PLMs across languages?}
Third, to our knowledge, no prior work has evaluated cross-lingual CND (X-CND).
This setting is particularly relevant for lower-resource languages, where labelled data is scarce and supervised training is often infeasible.
We therefore ask:
(\textbf{RQ3}) \textit{How well do English-trained models transfer to lower-resource target languages under zero-shot and few-shot settings?}

Our experiments show that encoder-style fine-tuning of SLMs~\cite{suganthan2025} yields gains of up to 8 percentage points over generative SFT variants (RQ1).
Across all resource levels, fine-tuned SLMs and PLMs substantially outperform prompted LLMs, with PLMs remaining a competitive baseline (RQ2).
For cross-lingual transfer (RQ3), English-trained SLMs achieve strong zero-shot performance on parallel data but accuracy drops by about 20 percentage points under real-world conditions.
Notably, encoder-style SLMs fine-tuned solely on English outperform prompted LLMs---even with little to no target-language supervision.
We further demonstrate that our findings obtained on high-quality FA articles generalize to randomly sampled Wikipedia claims, underscoring the practical utility of our models.

Our findings have important implications for practitioners aiming to support lower-resource language communities.
Crucially, fine-tuning SLMs on high-resource languages alone enables strong cross-lingual transfer, providing a viable path to deployment even when target-language data is scarce.
Beyond Wikipedia, our findings reinforce emerging evidence~\citep[e.g.,][]{bucher2024,majer2024} that compact, task-specific models are often preferable to general-purpose LLMs.

\begin{table*}[t]
\centering
    \begin{adjustbox}{max width=.8\linewidth}\centering

        \begin{tabular}{l llll}
    \hline
    \textbf{Dataset} & \textbf{Task} & \textbf{Domain} &\textbf{Languages (resource-level)} &\textbf{Size}\\
    
    \hline
    
    NLP4IF 2021~\cite{shaar2021} &
    CD/CWD &
    Covid-19/Politics &
    3 (2 high, 1 medium) &
    1.3K--4K \\
    
    CitationNeeded~\cite{redi2019} &
    CWD &
    Wikipedia &
    3 (3 high) &
    20K \\

    \citet{kazemi2021} &
    CD &
    Covid-19, Politics &
    5 (2 high, 2 medium, 1 low) &
    5K \\

    \citet{dutta2023} &
    CD &
    Politics &
    3 (2 high, 1 medium)&
    600--1.4K \\
    
    \citet{halitaj2024} &
    CWD &
    Wikipedia &
    3 (2 high, 1 medium) &
    31K-1.1m \\

    CheckThat 2024~\cite{hasanain2024} &
    CD/CWD &
    Politics, Covid-19 &
    3 (3 high) &
    1K--23K \\

    \citet{wmf_mcd} &
    CWD &
    Wikipedia &
    5 (5 high) &
    20K \\

    \addlinespace[0.2em]
    \hdashline
    \addlinespace[0.2em]
    
    \textbf{MCN (ours)}&
    CWD &
    Wikipedia &
    \textbf{18 (8 high, 5 medium, 5 low)} &
    \textbf{2}K--\textbf{1.1}m \\
    \hline
    \end{tabular}

    \end{adjustbox}
        
    \caption{Comparison of multilingual claim detection (CD) and claim check-worthiness detection (CWD) datasets.
    Resource-level definitions follow~\citet{joshi2020}. MCN substantially expands both linguistic and resource-level coverage compared to prior CD/CWD datasets.}
    \label{tab:data_comp}
\end{table*}

Our contributions are as follows: 
\begin{itemize}
    \item We introduce \textbf{M}ultilingual \textbf{C}itation \textbf{N}eeded (MCN), a large-scale CND corpus covering 18 linguistically diverse languages across three resource levels sourced from high-quality Wikipedia articles. MCN extends prior datasets by including lower-resource languages and providing substantially larger samples.

    \item We study the impact of different SFT variants on adapting SLMs for multilingual CND, and benchmark them against encoder-based PLMs and LLMs. Our results show that fine-tuned models consistently outperform LLM baselines.

    \item To our knowledge, we present the first evaluation of cross-lingual CND. Our results show that SLMs can transfer effectively across languages in zero- and few-shot regimes, highlighting their practical potential for deployment in low-resource settings.

\end{itemize}

\section{Related work}

\paragraph{Citation Needed Detection (CND)}  
CWD is an early-stage task in AFC that aims to identify claims requiring verification based on domain-specific criteria~\cite{guo2022}. 
Unlike verifiable claim detection (CD), CWD is inherently domain dependent~\cite{konstantinovskiy2021,panchendrarajan2024}. 
Accordingly, prior work has studied CWD across a range of domains, including rumor detection~\cite{zubiaga2018}, politics~\cite{hassan2015}, and Wikipedia~\cite{redi2019}. 
\citet{redi2019} formalized the CND task for Wikipedia as a binary supervised classification problem that identifies English claims requiring citations. \citet{halitaj2024} extended CND to three smaller non-English Wikipedias, and subsequent work proposed a sample-efficient CND framework~\cite{halitaj2025}.\footnote{Following~\cite{joshi2020}, only Albanian qualifies as a low-resource language in their work.} 
Recently, the Wikimedia Foundation expanded CND to several high-resource languages, including Spanish and Italian~\cite{wmf_mcd}. 
However, existing work largely focuses on high-resource languages, leaving lower-resource language communities underexplored.

\paragraph{Multilingual Claim Detection}
The majority of claim detection research has focused on English~\cite[e.g.][]{arslan2020,gangi2022}. 
The most prominent non-English claim detection efforts originate from the NLP4IF~\cite{shaar2021} and CheckThat!~\cite{hasanain2024} shared task series. 
Although a small number of additional multilingual efforts exist (Table~\ref{tab:data_comp}), several studies have emphasized the need for multilingual claim detection resources~\cite{panchendrarajan2024,dmonte2024,guo2022}. 
Regarding cross-lingual approaches, only \citet{panda2021} investigate cross-lingual transfer on the NLP4IF dataset~\cite{shaar2021} and show that mBERT generalizes well to high-resource languages. Our work addresses the critical lack of multilingual CWD data and enables both multilingual and cross-lingual experiments across languages with varying resource levels.

\section{Data}
\label{sec:data}

We introduce \textbf{M}ultilingual \textbf{C}itation \textbf{N}eeded (MCN), a CND corpus that spans 18 linguistically diverse languages across three resource levels~\cite{joshi2020}, sampled from high-quality Wikipedia articles. MCN advances prior datasets by incorporating medium- and low-resource languages and by containing a substantially larger number of claims (see Table~\ref{tab:data_comp}).

\begin{table*}[t]
    \centering
    \footnotesize
    \begin{adjustbox}{max width=1\linewidth}\centering

    \begin{tabular}{l cccc ccc ccc cc}
 & \multicolumn{4}{c}{\textbf{Language Statistics}} & \multicolumn{3}{c}{\textbf{No Citation Needed (0)}} & \multicolumn{3}{c}{\textbf{Citation Needed (1)}} \\
\cmidrule(lr){2-5}\cmidrule(l){6-8}\cmidrule(l){9-11}
 & \textbf{Resource} & \makecell{\textbf{N Featured} \\ \textbf{Articles}} & \makecell{\textbf{Language Group} \\ \textbf{(Family)}} & \textbf{Script} & \textbf{N Claims} & \textbf{Avg Len} & \textbf{\% Numeric} & \textbf{N Claims} & \textbf{Avg Len} & \textbf{\% Numeric} & \makecell{\textbf{Unique} \\ \textbf{Topics}} & \makecell{\textbf{Topic} \\ \textbf{Similarity}} \\
\hline
English (en) & high & 6,827 & Germanic (IE) & Latin & 20,824 & 20.1 & 16.8 & 1,142,036 & 23.4 & 39.3 & 51 & 0.71 \\ 
Portuguese (pt) & high & 3,485 & Romance (IE) & Latin & 7,357 & 22.0 & 31.6 & 499,035 & 25.6 & 34.9 & 48 & 0.94 \\ 
German (de) & high & 2,945 & Germanic (IE) & Latin & 273,180 & 18.4 & 32.6 & 615,757 & 19.1 & 35.2 & 51 & 0.96 \\ 
Russian (ru) & high & 2,094 & Slavic (IE) & Cyrillic & 26,195 & 17.3 & 28.5 & 709,255 & 19.8 & 31.4 & 49 & 0.92 \\ 
Italian (it) & high & 1,166 & Romance (IE) & Latin & 17,030 & 27.6 & 28.5 & 180,235 & 32.2 & 38.3 & 36 & 0.95 \\ 
Vietnamese (vi) & high & 1,044 & Mon-Khmer (AA) & Latin & 10,664 & 28.8 & 25.6 & 183,366 & 32.5 & 40.3 & 37 & 0.93 \\ 
Turkish (tr) & high & 573 & Oghuz (TRK) & Latin & 4,021 & 14.8 & 19.7 & 64,679 & 16.9 & 37.2 & 28 & 0.78 \\ 
Dutch (nl) & high & 380 & Germanic (IE) & Latin & 42,863 & 18.7 & 21.7 & 69,490 & 19.1 & 26.0 & 38 & 0.98 \\ 
\addlinespace[.3em]
\hdashline
\addlinespace[.3em]
Ukrainian (uk) & mid & 1,343 & Slavic (IE) & Cyrillic & 63,100 & 17.5 & 33.5 & 174,986 & 18.7 & 39.1 & 45 & 0.96 \\ 
Romanian (ro) & mid & 694 & Romance (IE) & Latin & 13,104 & 21.8 & 26.8 & 121,558 & 24.7 & 36.9 & 35 & 0.88 \\ 
Indonesian (id) & mid & 650 & Malayo-Polynesian (AN) & Latin & 4,446 & 19.2 & 30.1 & 106,202 & 20.8 & 34.4 & 32 & 0.88 \\ 
Bulgarian (bg) & mid & 190 & Slavic (IE) & Cyrillic & 12,091 & 19.0 & 31.3 & 27,722 & 19.9 & 34.1 & 27 & 0.96 \\ 
Uzbek (uz) & mid & 129 & Karluk (TRK) & Latin & 1,867 & 14.1 & 35.1 & 7,495 & 16.1 & 38.5 & 12 & 0.65 \\ 
\addlinespace[.3em]
\hdashline
\addlinespace[.3em]
Norwegian (no) & low & 1,406 & Germanic (IE) & Latin & 96,118 & 18.2 & 31.0 & 192,893 & 18.6 & 34.6 & 51 & 0.95 \\ 
Azerbaijani (az) & low & 768 & Oghuz (TRK) & Latin & 23,911 & 15.5 & 24.6 & 146,637 & 16.1 & 31.1 & 35 & 0.92 \\ 
Macedonian (mk) & low & 395 & Slavic (IE) & Cyrillic & 33,775 & 19.9 & 27.4 & 60,799 & 21.6 & 33.8 & 37 & 0.95 \\ 
Armenian (hy) & low & 118 & Armenian (IE) & Armenian & 6,873 & 16.5 & 34.1 & 25,090 & 17.6 & 38.9 & 27 & 0.84 \\ 
Albanian (sq) & low & 76 & Albanian (IE) & Latin & 3,990 & 21.8 & 24.8 & 10,664 & 24.6 & 46.0 & 18 & 0.38 \\ 
\midrule
Total/Average &  &  &  &  & 661,409 & 19.5 & 28.0 & 4,337,899 & 21.5 & 36.1 & 76 & 0.86 \\ 
\end{tabular}
    
    \end{adjustbox}
        
    \caption{
    Dataset statistics. MCN covers 18 languages spanning multiple resource levels~\cite{joshi2020}. Claims are sourced from Wikipedia featured articles and cover diverse language families and scripts. 
    }
    \label{tab:desc}
\end{table*}

\paragraph{Language Selection}
We select 18 languages that span linguistic diversity and represent high-, medium-, and low-resource levels~\cite{joshi2020} (Figure~\ref{fig:world_map}). Our selection is constrained by the availability of sufficiently high-quality Wikipedia articles from which we sample claims.

To ensure linguistic diversity, we sample languages from multiple families, groups, and scripts. 
We include several Indo-European (IE) language groups (e.g., Germanic, Slavic, and Albanian), covering three scripts (Armenian, Latin, and Cyrillic). 
To increase family-level diversity, we incorporate languages from the Austronesian, Austroasiatic, and Turkic families. 
Balancing linguistic diversity with resource availability, we select at least three languages per resource level from related language groups, including Germanic, Common Turkic, and Slavic.

\paragraph{Data Collection}
For each Wikipedia language edition, we collect claims from \textit{featured articles} (FAs), which editors designate as the highest-quality content on Wikipedia.\footnote{\url{https://en.wikipedia.org/wiki/Category:Featured\_articles}} 
One FA criterion states that "claims are verifiable against high-quality reliable sources and are supported by inline citations where appropriate."\footnote{\url{https://en.wikipedia.org/wiki/Wikipedia:Featured_article_criteria}} 
Prior analyses of Wikipedia’s citation quality confirm that FAs are consistently well sourced~\cite{baigutanova2023}. 
Consequently, we can derive binary CND labels, whether a claim requires a citation, by leveraging the presence or absence of inline citations. 
Following prior CND work~\cite{redi2019,halitaj2024}, we label a claim as 1 (Citation Needed) if it contains an inline citation or appears in a paragraph with a citation at the end; we label claims without inline citations as 0 (No Citation Needed). 
We further augment each claim with contextual information (the preceding and following sentences, when available) and the article’s topic (e.g., Health), which we obtain using Wikipedia’s ORES topic model~\cite{halfaker2020}.
We remove all citation markers from claims and apply extensive cleaning to ensure that no citation artifacts remain.
Appendix~\ref{app:data} provides additional details on the data collection and cleaning procedures.

\paragraph{Dataset Statistics}
Table~\ref{tab:desc} presents dataset statistics. 
As expected, lower-resource languages contain fewer FAs. 
However, even among languages within the same resource level, the number of available articles varies substantially. 
Across languages, claims that require citations tend to be longer and are more likely to include numeric expressions. 
These claims originate from articles that cover, on average, 76 unique topics (e.g., Health, Geography). Topic Similarity, computed as the cosine similarity between the topic distributions of claims with and without citations, remains relatively high. 
This pattern holds across languages except for Albanian, likely due to its considerably smaller number of available FAs. 
Appendix~\ref{app:data} provides additional descriptive statistics and example claims.



        

\paragraph{Data Quality}
A potential concern is that parsing errors during data collection or incorrect citation retrieval may introduce bias.
To address this issue, we conduct an extensive manual evaluation of data quality, which we report in Appendix~\ref{app:data}.

\paragraph{Data Split}
For each language, we split the data at the article level to ensure that test claims originate from articles unseen during training. 
Across all experiments, we use the same train/dev/test splits of 5k/500/500 claims, balanced by label. 
We report accuracy as the primary evaluation metric and provide F1 scores in Appendix~\ref{app:results}.

\section{Experimental Setup}
We first describe the different SFT variants for adapting SLMs to CND, followed by a description of our experimental setup.


\subsection{SFT Variants}
We specialize SLMs via SFT~\cite{wei2021} using QLoRA~\cite{dettmers2023}, making our experiments reproducible on a single A100 40GB.
We explore three SFT variants, with generative and discriminative objectives: 

\paragraph{SFT Variant 1: Full-token loss (FTL)}
The FTL is the standard loss in most work, and the default in Huggingface's SFTTrainer~\cite{vonwerra2022}. 
Training minimizes the autoregressive next-token loss over \textit{all} tokens of the prompt (i.e., system, user, and assistant messages). 
This strategy has been adapted in prior claim detection work~\cite[e.g.,][]{li2024,bell2025}. 

\paragraph{SFT Variant 2: Target-only loss (TOL)}
TOL differs from FTL by masking system and user tokens so that model weights update only on assistant (i.e., target) tokens. 
When fine-tuning models with QLoRA on the MMLU benchmark, \citet{dettmers2023} show that TOL outperforms FTL. 
The motivation for applying TOL to CND is to encourage the model to 'focus' its learning capacity on predicting the label rather than reconstructing the input prompt.

\paragraph{SFT Variant 3: Encoder-style (ES)}
Rather than using generative SFT, recent work shows that SLMs can be effectively adapted as encoder-style discriminators~\cite[e.g.,][]{bolton2024,li2025}. 
We implement this variant similar to~\citet{suganthan2025} by feeding the last-token hidden state into a feed-forward classifier. 
We fine-tune this variant by concatenating the claim with its context and optimizing it using standard cross-entropy loss.

\begin{table*}[t]

    {
    \renewcommand{\arraystretch}{0.9}
    \centering
    \begin{adjustbox}{max width=\linewidth}\centering

    \begin{tabular}{llCCCCCCCCCCCCCCCCCCCCC}
 & & \multicolumn{9}{c}{\textbf{High Resource}} & \multicolumn{6}{c}{\textbf{Medium Resource}} & \multicolumn{6}{c}{\textbf{Low Resource}} \\\cmidrule(lr){3-11}  \cmidrule(lr){12-17}  \cmidrule(lr){18-23}\textbf{Model} & \textbf{SFT} & \textbf{en} & \textbf{pt} & \textbf{de} & \textbf{ru} & \textbf{it} & \textbf{vi} & \textbf{tr} & \textbf{nl}& \textbf{Avg} & \textbf{uk} & \textbf{ro} & \textbf{id} & \textbf{bg} & \textbf{uz}& \textbf{Avg} & \textbf{no} & \textbf{az} & \textbf{mk} & \textbf{hy} & \textbf{sq}& \textbf{Avg} \\
\toprule
\multirow{3}{*}[-1.5ex]{Aya} &  FTL & 83.28  \scriptsize{$(\pm0.71)$} & 56.47  \scriptsize{$(\pm7.95)$} & 54.08  \scriptsize{$(\pm0.47)$} & 65.73  \scriptsize{$(\pm4.81)$} & 59.99  \scriptsize{$(\pm4.18)$} & 68.72  \scriptsize{$(\pm5.64)$} & 71.53  \scriptsize{$(\pm0.31)$} & 53.56  \scriptsize{$(\pm1.91)$} & \cellcolor{gray!25} 64.17 \scriptsize{$(\pm10.24)$} & 58.48  \scriptsize{$(\pm3.20)$} & 65.27  \scriptsize{$(\pm4.16)$} & 66.79  \scriptsize{$(\pm2.75)$} & 54.11  \scriptsize{$(\pm1.40)$} & 50.23  \scriptsize{$(\pm0.41)$} & \cellcolor{gray!25} 58.98 \scriptsize{$(\pm7.09)$} & 56.93  \scriptsize{$(\pm3.72)$} & 56.20  \scriptsize{$(\pm6.30)$} & 56.41  \scriptsize{$(\pm4.46)$} & 51.08  \scriptsize{$(\pm4.08)$} & \textbf{71.40} \scriptsize{$(\pm 3.65)$} & \cellcolor{gray!25} 58.40 \scriptsize{$(\pm7.64)$} \\
 & TOL & 82.03  \scriptsize{$(\pm1.67)$} & 67.47  \scriptsize{$(\pm0.80)$} & 62.66  \scriptsize{$(\pm4.11)$} & 73.81  \scriptsize{$(\pm0.46)$} & 69.55  \scriptsize{$(\pm1.78)$} & 74.95  \scriptsize{$(\pm5.02)$} & \textbf{74.40} \scriptsize{$(\pm 2.95)$} & 50.20  \scriptsize{$(\pm0.17)$} & \cellcolor{gray!25} 69.38 \scriptsize{$(\pm9.66)$} & 59.52  \scriptsize{$(\pm2.61)$} & 74.60  \scriptsize{$(\pm1.45)$} & 56.71  \scriptsize{$(\pm11.80)$} & 52.14  \scriptsize{$(\pm3.27)$} & 50.00  \scriptsize{$(\pm0.00)$} & \cellcolor{gray!25} 58.59 \scriptsize{$(\pm9.70)$} & 51.80  \scriptsize{$(\pm2.46)$} & 59.13  \scriptsize{$(\pm7.96)$} & 53.12  \scriptsize{$(\pm4.76)$} & 47.84  \scriptsize{$(\pm7.20)$} & 68.40  \scriptsize{$(\pm5.09)$} & \cellcolor{gray!25} 56.06 \scriptsize{$(\pm8.00)$} \\
 & ES & \textbf{89.45} \scriptsize{$(\pm 0.90)$} & \textbf{78.58} \scriptsize{$(\pm 1.29)$} & \textbf{64.33} \scriptsize{$(\pm 1.31)$} & \textbf{78.02} \scriptsize{$(\pm 0.83)$} & \textbf{74.14} \scriptsize{$(\pm 1.80)$} & \textbf{80.83} \scriptsize{$(\pm 1.95)$} & 72.13  \scriptsize{$(\pm0.23)$} & \textbf{62.46} \scriptsize{$(\pm 0.81)$} & \cellcolor{gray!25} \textbf{74.99} \scriptsize{$(\pm8.82)$} & \textbf{70.21} \scriptsize{$(\pm 1.93)$} & \textbf{77.89} \scriptsize{$(\pm 1.49)$} & \textbf{70.81} \scriptsize{$(\pm 1.10)$} & \textbf{57.85} \scriptsize{$(\pm 1.17)$} & \textbf{59.27} \scriptsize{$(\pm 1.72)$} & \cellcolor{gray!25} \textbf{67.20} \scriptsize{$(\pm8.47)$} & \textbf{65.33} \scriptsize{$(\pm 0.99)$} & \textbf{69.20} \scriptsize{$(\pm 0.53)$} & \textbf{63.92} \scriptsize{$(\pm 0.91)$} & \textbf{59.59} \scriptsize{$(\pm 1.45)$} & 70.40  \scriptsize{$(\pm4.78)$} & \cellcolor{gray!25} \textbf{65.69} \scriptsize{$(\pm4.33)$} \\
\midrule
\multirow{3}{*}[-1.5ex]{Llama3} &  FTL & 88.64  \scriptsize{$(\pm1.03)$} & 77.84  \scriptsize{$(\pm1.52)$} & 60.65  \scriptsize{$(\pm1.86)$} & 77.15  \scriptsize{$(\pm0.87)$} & 73.94  \scriptsize{$(\pm1.24)$} & 77.35  \scriptsize{$(\pm0.40)$} & 73.40  \scriptsize{$(\pm0.53)$} & 60.52  \scriptsize{$(\pm0.53)$} & \cellcolor{gray!25} 73.69 \scriptsize{$(\pm9.33)$} & 69.14  \scriptsize{$(\pm2.23)$} & 75.60  \scriptsize{$(\pm2.10)$} & 74.55  \scriptsize{$(\pm0.69)$} & \textbf{60.86} \scriptsize{$(\pm 1.22)$} & \textbf{66.73} \scriptsize{$(\pm 5.43)$} & \cellcolor{gray!25} \textbf{69.38} \scriptsize{$(\pm6.03)$} & 62.07  \scriptsize{$(\pm2.57)$} & 67.47  \scriptsize{$(\pm1.86)$} & 65.93  \scriptsize{$(\pm0.91)$} & 58.59  \scriptsize{$(\pm2.31)$} & \textbf{71.60} \scriptsize{$(\pm 0.92)$} & \cellcolor{gray!25} 65.13 \scriptsize{$(\pm5.00)$} \\
 & TOL & 89.58  \scriptsize{$(\pm1.84)$} & 77.24  \scriptsize{$(\pm0.46)$} & 61.86  \scriptsize{$(\pm1.48)$} & 77.35  \scriptsize{$(\pm0.87)$} & \textbf{74.97} \scriptsize{$(\pm 1.04)$} & \textbf{80.63} \scriptsize{$(\pm 1.22)$} & \textbf{73.87} \scriptsize{$(\pm 0.58)$} & 60.59  \scriptsize{$(\pm1.67)$} & \cellcolor{gray!25} 74.51 \scriptsize{$(\pm9.52)$} & 68.54  \scriptsize{$(\pm1.39)$} & \textbf{76.14} \scriptsize{$(\pm 1.64)$} & 75.15  \scriptsize{$(\pm0.53)$} & 60.05  \scriptsize{$(\pm0.50)$} & 60.00  \scriptsize{$(\pm4.33)$} & \cellcolor{gray!25} 67.98 \scriptsize{$(\pm7.82)$} & 63.80  \scriptsize{$(\pm0.92)$} & 66.11  \scriptsize{$(\pm0.89)$} & 64.99  \scriptsize{$(\pm1.74)$} & 59.46  \scriptsize{$(\pm6.68)$} & 67.07  \scriptsize{$(\pm4.13)$} & \cellcolor{gray!25} 64.28 \scriptsize{$(\pm2.96)$} \\
 & ES & \textbf{89.65} \scriptsize{$(\pm 0.23)$} & \textbf{78.58} \scriptsize{$(\pm 1.03)$} & \textbf{66.00} \scriptsize{$(\pm 0.64)$} & \textbf{78.89} \scriptsize{$(\pm 1.16)$} & 74.83  \scriptsize{$(\pm2.08)$} & 80.09  \scriptsize{$(\pm1.29)$} & 71.47  \scriptsize{$(\pm1.30)$} & \textbf{61.19} \scriptsize{$(\pm 3.18)$} & \cellcolor{gray!25} \textbf{75.09} \scriptsize{$(\pm8.89)$} & \textbf{72.01} \scriptsize{$(\pm 1.01)$} & 73.59  \scriptsize{$(\pm8.09)$} & \textbf{75.55} \scriptsize{$(\pm 1.22)$} & 53.51  \scriptsize{$(\pm4.82)$} & 57.67  \scriptsize{$(\pm4.96)$} & \cellcolor{gray!25} 66.46 \scriptsize{$(\pm10.12)$} & \textbf{65.73} \scriptsize{$(\pm 2.89)$} & \textbf{71.47} \scriptsize{$(\pm 2.27)$} & \textbf{66.47} \scriptsize{$(\pm 1.69)$} & \textbf{60.79} \scriptsize{$(\pm 1.89)$} & 69.80  \scriptsize{$(\pm1.91)$} & \cellcolor{gray!25} \textbf{66.85} \scriptsize{$(\pm4.13)$} \\
\midrule
\multirow{3}{*}[-1.5ex]{Qwen3} &  FTL & 88.11  \scriptsize{$(\pm2.42)$} & 76.77  \scriptsize{$(\pm2.15)$} & 56.85  \scriptsize{$(\pm5.08)$} & 74.62  \scriptsize{$(\pm0.99)$} & 73.46  \scriptsize{$(\pm2.09)$} & 79.02  \scriptsize{$(\pm1.41)$} & 73.93  \scriptsize{$(\pm0.31)$} & 58.52  \scriptsize{$(\pm0.60)$} & \cellcolor{gray!25} 72.66 \scriptsize{$(\pm10.37)$} & 61.55  \scriptsize{$(\pm1.94)$} & 74.73  \scriptsize{$(\pm3.44)$} & \textbf{73.41} \scriptsize{$(\pm 2.82)$} & 60.45  \scriptsize{$(\pm0.31)$} & 58.73  \scriptsize{$(\pm1.29)$} & \cellcolor{gray!25} 65.78 \scriptsize{$(\pm7.65)$} & 62.33  \scriptsize{$(\pm4.39)$} & 67.40  \scriptsize{$(\pm2.21)$} & 58.82  \scriptsize{$(\pm2.53)$} & 57.75  \scriptsize{$(\pm2.76)$} & 69.27  \scriptsize{$(\pm3.13)$} & \cellcolor{gray!25} 63.11 \scriptsize{$(\pm5.10)$} \\
 & TOL & 89.11  \scriptsize{$(\pm1.36)$} & \textbf{77.31} \scriptsize{$(\pm 0.35)$} & 60.72  \scriptsize{$(\pm1.57)$} & 77.42  \scriptsize{$(\pm0.83)$} & 73.53  \scriptsize{$(\pm2.65)$} & 78.09  \scriptsize{$(\pm1.10)$} & \textbf{75.53} \scriptsize{$(\pm 0.31)$} & 60.05  \scriptsize{$(\pm2.03)$} & \cellcolor{gray!25} 73.97 \scriptsize{$(\pm9.57)$} & 67.05  \scriptsize{$(\pm0.74)$} & 76.48  \scriptsize{$(\pm1.31)$} & 71.88  \scriptsize{$(\pm0.95)$} & \textbf{60.99} \scriptsize{$(\pm 1.52)$} & \textbf{72.00} \scriptsize{$(\pm 0.80)$} & \cellcolor{gray!25} \textbf{69.68} \scriptsize{$(\pm5.89)$} & 63.67  \scriptsize{$(\pm1.86)$} & 65.62  \scriptsize{$(\pm2.22)$} & 60.23  \scriptsize{$(\pm1.87)$} & 60.09  \scriptsize{$(\pm0.86)$} & \textbf{71.00} \scriptsize{$(\pm 1.51)$} & \cellcolor{gray!25} 64.12 \scriptsize{$(\pm4.50)$} \\
 & ES & \textbf{89.38} \scriptsize{$(\pm 0.53)$} & 76.97  \scriptsize{$(\pm0.31)$} & \textbf{65.06} \scriptsize{$(\pm 1.14)$} & \textbf{77.49} \scriptsize{$(\pm 0.70)$} & \textbf{75.58} \scriptsize{$(\pm 0.93)$} & \textbf{81.10} \scriptsize{$(\pm 0.12)$} & 71.60  \scriptsize{$(\pm1.91)$} & \textbf{62.26} \scriptsize{$(\pm 2.62)$} & \cellcolor{gray!25} \textbf{74.93} \scriptsize{$(\pm8.68)$} & \textbf{71.08} \scriptsize{$(\pm 2.09)$} & \textbf{77.15} \scriptsize{$(\pm 0.42)$} & 70.88  \scriptsize{$(\pm4.07)$} & 59.39  \scriptsize{$(\pm3.31)$} & 58.60  \scriptsize{$(\pm4.20)$} & \cellcolor{gray!25} 67.42 \scriptsize{$(\pm8.10)$} & \textbf{66.67} \scriptsize{$(\pm 2.27)$} & \textbf{71.47} \scriptsize{$(\pm 1.67)$} & \textbf{63.11} \scriptsize{$(\pm 1.48)$} & \textbf{62.26} \scriptsize{$(\pm 1.50)$} & 68.13  \scriptsize{$(\pm2.00)$} & \cellcolor{gray!25} \textbf{66.33} \scriptsize{$(\pm3.76)$} \\
\bottomrule
\end{tabular}
    
    \end{adjustbox}
    }    
    \caption{Monolingual CND accuracy of SLMs across SFT variants (FTL=full-token loss, TOL=target-only loss, ES=encoder-style). 
    \textbf{Bold} denotes the best score per model–language pair. Parentheses indicate standard deviation. 
    ES SFT consistently outperforms generative SFT, while TOL yields modest gains over FTL in some settings.}
    \label{tab:atl_acc}
\end{table*}

\paragraph{SFT Data}
For generative SFT, we need to format each data sample into a chat-style format.
Given our dataset \(d = \{(c_i, x_i, y_i)\}_{i=1}^{N}\) consisting of \textit{claim} \(c_i\), \textit{context} \(x_i\), and \textit{label} \(y_i\),  
we convert each example into a chat-style prompt with \texttt{system}, \texttt{user}, and \texttt{assistant} messages.  
The \texttt{system} message contains task instructions;
the \texttt{user} message concatenates the claim and its context (if available);
and  the \texttt{assistant} message contains the target label formatted as JSON (e.g., \texttt{\{"label": 1\}}) to enforce consistent output formatting.

\paragraph{Models}
We use Llama-8B~\cite{grattafiori2024}, Qwen3-8B~\cite{yang2025} and Aya-8b~\cite{ustun2024}. We omit smaller sizes which we leave for future work.

\subsection{Experiments}
We design an experiment for each research question.
Appendix~\ref{app:experimental_details} presents experimental details for full reproducibility.

\paragraph{(1) SFT Variants}
To evaluate the effectiveness of different SFT variants for adapting SLMs to CND (\textbf{RQ1}), we fine-tune each model for every combination of SFT variant and language using a hyperparameter grid search.
We use a simple instruction prompt across all runs, as preliminary experiments show little benefit from prompt engineering.

\paragraph{(2) Multilingual CND Benchmark}
To evaluate the performance of different LM families for CND on MCN (\textbf{RQ2}), we benchmark SLMs against encoder-based PLMs and commercial LLMs. For PLMs, we fully fine-tune mBERT~\cite{devlin2019} and XLM-R~\cite{conneau2020} in both base and large variants. 
For LLMs, we evaluate GPT-5.2~\cite{openai2025} and Gemini 2.5 Flash~\cite{comanici2025}, combining zero- and few-shot prompting with both instruction-style and verbose prompts, as LLMs may benefit from more detailed task explanations~\cite{majer2024}.

\paragraph{(3) X-CND}
The ability of trained models to perform cross-lingual CND remains an open question (\textbf{RQ3}). We investigate this question by training mBERT and Llama with all three SFT variants on English as the \textit{source} language. 
We then evaluate their zero-shot and few-shot transfer performance on all medium- and low-resource \textit{target} languages in our corpus.

For zero-shot transfer, we consider two evaluation settings:
(\textit{i}) \textbf{Parallel Test Data}:  
Because claims in MCN are not parallel, we construct synthetic parallel test sets by translating the English test set into all  target languages included in our study.\footnote{We use translation-specialized LLMs via \href{https://cloud.google.com/translate}{Google Cloud Translation}.} 
To address potential translation quality issues~\cite{artetxe2020}, we perform extensive automatic quality checks, which we detail in Appendix~\ref{app:data}.
(\textit{ii}) \textbf{Non-parallel Test Data}:  
Evaluation on non-parallel test data assesses models’ cross-lingual transfer under both language \textit{and} claim distribution shifts (e.g., topics and semantics). 
This setting poses a more challenging yet more realistic evaluation scenario, as models deployed in practice must classify claims that diverge from their training data distribution.

For few-shot transfer, we follow the strategy by~\citet{lauscher2020} for the non-parallel data setting:
each model trained on English is further fine-tuned on a small number of target-language examples ($N < 500$).
While this approach has proven effective for PLMs across a range of NLP tasks~\citep{lauscher2020}, its effectiveness for CND remains unclear and, to our knowledge, it has not previously been applied to SLMs.

\section{Results}

\subsection*{Experiment 1: SFT Variants}

\textbf{Encoder-style tuning mostly outperforms other SFT variants for multilingual CND across language resource levels.}
Table~\ref{tab:all_acc} reports CND accuracy for SLMs across SFT variants and languages of all resource levels. 
Across nearly all model-resource combinations, ES tuning achieves higher average accuracy than generative SFT variants. 
This effect is most pronounced for Aya, where ES tuning outperforms generative variants on almost all individual languages (e.g., by an average of 10 percentage points compared to FTL on high-resource languages). 
For Qwen and Llama, ES tuning also outperforms other variants on both high- and low-resource languages, although the gains are smaller (e.g., an average improvement of 3 percentage points for Llama on low-resource languages).
On medium-resource languages, generative tuning variants achieve higher average performance. 
However, this result is driven by a single outlier (Uzbek) where generative tuning yields accuracy gains of up to 13 percentage points over ES tuning. 
For most other medium-resource languages, ES tuning remains competitive with generative SFT.
Among the generative variants, TOL training yields slightly higher performance than FTL on high-resource languages across models. 
On medium- and low-resource languages, TOL and FTL perform comparably, with differences of only 1–2 percentage points.

\begin{table*}[t]
\centering

    {
    \renewcommand{\arraystretch}{0.8}
    
        \begin{adjustbox}{max width=\linewidth}\centering

\begin{tabular}{lCCCCCCCCCCCCCCCCCCCCCC}
 & \multicolumn{9}{c}{\textbf{High Resource}} & \multicolumn{6}{c}{\textbf{Medium Resource}} & \multicolumn{6}{c}{\textbf{Low Resource}} \\\cmidrule(lr){2-10}  \cmidrule(lr){11-16}  \cmidrule(lr){17-22} & \textbf{en} & \textbf{pt} & \textbf{de} & \textbf{ru} & \textbf{it} & \textbf{vi} & \textbf{tr} & \textbf{nl} & \textbf{Avg} & \textbf{uk} & \textbf{ro} & \textbf{id} & \textbf{bg} & \textbf{uz} & \textbf{Avg} & \textbf{no} & \textbf{az} & \textbf{mk} & \textbf{hy} & \textbf{sq} & \textbf{Avg} & \textbf{Avg} \\
\toprule
\multicolumn{23}{c}{\textbf{Decoder-based LLMs}} \\
\midrule
\textit{Zero-shot} \\
\midrule
GPT-5 (Instruct) & 75.00 & 58.80 & 54.40 & 65.20 & 60.52 & 64.40 & 69.20 & 50.60 & \cellcolor{gray!25} 62.27 \scriptsize{($\pm7.90$)} & 53.20 & 65.00 & 53.60 & 53.00 & 56.60 & \cellcolor{gray!25} 56.28 \scriptsize{($\pm5.09$)} & 51.80 & 59.00 & 56.00 & 50.00 & 53.20 & \cellcolor{gray!25} 54.00 \scriptsize{($\pm3.55$)} & 58.31 \scriptsize{($\pm6.98$)} \\
GPT-5 (Verbose) & 80.60 & 57.00 & 51.60 & 62.40 & 55.20 & 66.40 & 69.40 & 50.40 & \cellcolor{gray!25} 61.62 \scriptsize{($\pm10.23$)} & 52.20 & 64.60 & 55.80 & 53.20 & 54.20 & \cellcolor{gray!25} 56.00 \scriptsize{($\pm4.99$)} & 52.80 & 61.20 & 53.40 & 50.60 & 51.60 & \cellcolor{gray!25} 53.92 \scriptsize{($\pm4.21$)} & 57.92 \scriptsize{($\pm8.09$)} \\
Gemini 2.5 (Instruct) & 75.80 & 59.20 & 56.20 & 65.80 & 64.20 & 63.20 & 62.40 & 51.20 & \cellcolor{gray!25} 62.25 \scriptsize{($\pm7.26$)} & 55.60 & 63.40 & 57.40 & 57.00 & 60.20 & \cellcolor{gray!25} 58.72 \scriptsize{($\pm3.10$)} & 54.00 & 54.60 & 55.00 & 50.60 & 55.60 & \cellcolor{gray!25} 53.96 \scriptsize{($\pm1.97$)} & 58.97 \scriptsize{($\pm6.11$)} \\
Gemini 2.5 (Verbose) & 79.20 & 62.80 & 54.40 & 62.00 & 52.60 & 67.40 & 68.40 & 52.00 & \cellcolor{gray!25} 62.35 \scriptsize{($\pm9.35$)} & 54.80 & 62.40 & 58.40 & 58.00 & 55.20 & \cellcolor{gray!25} 57.76 \scriptsize{($\pm3.05$)} & 53.00 & 58.40 & 53.60 & 49.40 & 51.00 & \cellcolor{gray!25} 53.08 \scriptsize{($\pm3.41$)} & 58.50 \scriptsize{($\pm7.53$)} \\
\midrule
\textit{Few-shot} \\
\midrule
GPT-5 (Instruct) & 82.00 & 58.80 & 54.00 & 68.00 & 68.61 & 69.00 & \underline{72.20} & 53.31 & \cellcolor{gray!25} 65.74 \scriptsize{($\pm9.79$)} & 53.60 & 71.40 & 56.80 & 57.40 & 67.00 & \cellcolor{gray!25} 61.24 \scriptsize{($\pm7.57$)} & 52.60 & 61.20 & 55.40 & 52.80 & 55.00 & \cellcolor{gray!25} 55.40 \scriptsize{($\pm3.48$)} & 61.62 \scriptsize{($\pm8.67$)} \\
GPT-5 (Verbose) & 80.40 & 58.20 & 50.80 & 62.20 & 57.00 & 68.74 & 68.94 & 51.20 & \cellcolor{gray!25} 62.19 \scriptsize{($\pm10.09$)} & 53.31 & 65.80 & 57.40 & 53.60 & 56.00 & \cellcolor{gray!25} 57.22 \scriptsize{($\pm5.09$)} & 52.60 & 61.12 & 53.40 & 51.60 & 50.60 & \cellcolor{gray!25} 53.86 \scriptsize{($\pm4.19$)} & 58.49 \scriptsize{($\pm8.09$)} \\
Gemini 2.5 (Instruct) & 80.60 & 58.00 & 59.80 & 67.54 & 66.60 & 68.20 & 69.20 & 55.00 & \cellcolor{gray!25} 65.62 \scriptsize{($\pm8.05$)} & 55.60 & 62.20 & 62.60 & \textbf{63.00} & \underline{67.20} & \cellcolor{gray!25} 62.12 \scriptsize{($\pm4.16$)} & 54.60 & 62.20 & 57.00 & 59.00 & 64.40 & \cellcolor{gray!25} 59.44 \scriptsize{($\pm3.93$)} & 62.93 \scriptsize{($\pm6.45$)} \\
Gemini 2.5 (Verbose) & 81.00 & 63.00 & 50.40 & 61.60 & 53.60 & 71.00 & 67.60 & 53.00 & \cellcolor{gray!25} 62.65 \scriptsize{($\pm10.40$)} & 55.60 & 62.40 & 63.20 & 55.00 & 61.40 & \cellcolor{gray!25} 59.52 \scriptsize{($\pm3.91$)} & 53.80 & 58.80 & 56.20 & 53.40 & 54.40 & \cellcolor{gray!25} 55.32 \scriptsize{($\pm2.22$)} & 59.74 \scriptsize{($\pm7.68$)} \\
\midrule
\multicolumn{23}{c}{\textbf{Encoder-based PLMs}} \\
\midrule
XLM-R-base & 87.91 \scriptsize{($\pm0.31$)} & 73.49 \scriptsize{($\pm2.41$)} & 63.79 \scriptsize{($\pm1.68$)} & 77.02 \scriptsize{($\pm1.14$)} & 68.59 \scriptsize{($\pm2.88$)} & 73.15 \scriptsize{($\pm1.45$)} & \textbf{72.87} \scriptsize{($\pm0.58$)} & 55.11 \scriptsize{($\pm5.22$)} & \cellcolor{gray!25} 71.49 \scriptsize{($\pm9.60$)} & 65.06 \scriptsize{($\pm2.14$)} & 71.84 \scriptsize{($\pm5.25$)} & 64.80 \scriptsize{($\pm3.41$)} & 53.57 \scriptsize{($\pm3.07$)} & 50.00 \scriptsize{($\pm0.00$)} & \cellcolor{gray!25} 61.05 \scriptsize{($\pm9.01$)} & 63.07 \scriptsize{($\pm0.99$)} & 62.73 \scriptsize{($\pm0.83$)} & 62.84 \scriptsize{($\pm1.14$)} & \underline{62.93} \scriptsize{($\pm0.57$)} & 66.60 \scriptsize{($\pm7.35$)} & \cellcolor{gray!25} 63.63 \scriptsize{($\pm1.66$)} & 66.48 \scriptsize{($\pm9.21$)} \\
XLM-R-large & \textbf{89.65} \scriptsize{($\pm0.31$)} & 77.04 \scriptsize{($\pm2.54$)} & 61.06 \scriptsize{($\pm5.73$)} & 77.62 \scriptsize{($\pm0.64$)} & 73.18 \scriptsize{($\pm0.52$)} & 76.82 \scriptsize{($\pm0.83$)} & 71.40 \scriptsize{($\pm1.91$)} & 54.24 \scriptsize{($\pm7.35$)} & \cellcolor{gray!25} 72.63 \scriptsize{($\pm10.86$)} & 62.86 \scriptsize{($\pm6.84$)} & \textbf{79.50} \scriptsize{($\pm1.53$)} & 68.14 \scriptsize{($\pm3.12$)} & 55.98 \scriptsize{($\pm0.93$)} & 50.00 \scriptsize{($\pm0.00$)} & \cellcolor{gray!25} 63.30 \scriptsize{($\pm11.36$)} & \underline{66.80} \scriptsize{($\pm1.11$)} & 64.80 \scriptsize{($\pm1.40$)} & 63.85 \scriptsize{($\pm1.11$)} & 62.32 \scriptsize{($\pm1.31$)} & 66.33 \scriptsize{($\pm1.45$)} & \cellcolor{gray!25} 64.82 \scriptsize{($\pm1.83$)} & 67.86 \scriptsize{($\pm10.18$)} \\
mBert & 88.71 \scriptsize{($\pm0.61$)} & \textbf{78.85} \scriptsize{($\pm0.61$)} & 64.73 \scriptsize{($\pm1.64$)} & 76.55 \scriptsize{($\pm0.72$)} & 73.39 \scriptsize{($\pm1.13$)} & 78.02 \scriptsize{($\pm2.45$)} & 71.53 \scriptsize{($\pm1.27$)} & \textbf{63.33} \scriptsize{($\pm1.31$)} & \cellcolor{gray!25} 74.39 \scriptsize{($\pm8.17$)} & 70.61 \scriptsize{($\pm2.82$)} & \underline{78.29} \scriptsize{($\pm1.57$)} & \underline{73.15} \scriptsize{($\pm0.87$)} & \underline{59.65} \scriptsize{($\pm1.50$)} & \textbf{68.13} \scriptsize{($\pm14.20$)} & \cellcolor{gray!25} \textbf{69.97} \scriptsize{($\pm6.88$)} & \textbf{69.27} \scriptsize{($\pm2.21$)} & \underline{71.30} \scriptsize{($\pm0.42$)} & 63.31 \scriptsize{($\pm0.81$)} & \textbf{63.53} \scriptsize{($\pm2.26$)} & \textbf{70.47} \scriptsize{($\pm1.50$)} & \cellcolor{gray!25} \textbf{67.57} \scriptsize{($\pm3.86$)} & \textbf{71.27} \scriptsize{($\pm7.72$)} \\
\midrule
\multicolumn{23}{c}{\textbf{Decoder-based SLMs}} \\
\midrule
Aya (ES) & \underline{89.45} \scriptsize{($\pm0.90$)} & \underline{78.58} \scriptsize{($\pm1.29$)} & 64.33 \scriptsize{($\pm1.31$)} & \underline{78.02} \scriptsize{($\pm0.83$)} & 74.14 \scriptsize{($\pm1.80$)} & \underline{80.83} \scriptsize{($\pm1.95$)} & 72.13 \scriptsize{($\pm0.23$)} & \underline{62.46} \scriptsize{($\pm0.81$)} & \cellcolor{gray!25} \underline{74.99} \scriptsize{($\pm8.82$)} & 70.21 \scriptsize{($\pm1.93$)} & 77.89 \scriptsize{($\pm1.49$)} & 70.81 \scriptsize{($\pm1.10$)} & 57.85 \scriptsize{($\pm1.17$)} & 59.27 \scriptsize{($\pm1.72$)} & \cellcolor{gray!25} 67.20 \scriptsize{($\pm8.47$)} & 65.33 \scriptsize{($\pm0.99$)} & 69.20 \scriptsize{($\pm0.53$)} & \underline{63.92} \scriptsize{($\pm0.91$)} & 59.59 \scriptsize{($\pm1.45$)} & \underline{70.40} \scriptsize{($\pm4.78$)} & \cellcolor{gray!25} 65.69 \scriptsize{($\pm4.33$)} & 70.24 \scriptsize{($\pm8.48$)} \\
Llama3 (ES) & \textbf{89.65} \scriptsize{($\pm0.23$)} & \underline{78.58} \scriptsize{($\pm1.03$)} & \textbf{66.00} \scriptsize{($\pm0.64$)} & \textbf{78.89} \scriptsize{($\pm1.16$)} & \underline{74.83} \scriptsize{($\pm2.08$)} & 80.09 \scriptsize{($\pm1.29$)} & 71.47 \scriptsize{($\pm1.30$)} & 61.19 \scriptsize{($\pm3.18$)} & \cellcolor{gray!25} \textbf{75.09} \scriptsize{($\pm8.89$)} & \textbf{72.01} \scriptsize{($\pm1.01$)} & 73.59 \scriptsize{($\pm8.09$)} & \textbf{75.55} \scriptsize{($\pm1.22$)} & 53.51 \scriptsize{($\pm4.82$)} & 57.67 \scriptsize{($\pm4.96$)} & \cellcolor{gray!25} 66.46 \scriptsize{($\pm10.12$)} & 65.73 \scriptsize{($\pm2.89$)} & \textbf{71.47} \scriptsize{($\pm2.27$)} & \textbf{66.47} \scriptsize{($\pm1.69$)} & 60.79 \scriptsize{($\pm1.89$)} & 69.80 \scriptsize{($\pm1.91$)} & \cellcolor{gray!25} \underline{66.85} \scriptsize{($\pm4.13$)} & \underline{70.72} \scriptsize{($\pm8.82$)} \\
Qwen3 (ES) & 89.38 \scriptsize{($\pm0.53$)} & 76.97 \scriptsize{($\pm0.31$)} & \underline{65.06} \scriptsize{($\pm1.14$)} & 77.49 \scriptsize{($\pm0.70$)} & \textbf{75.58} \scriptsize{($\pm0.93$)} & \textbf{81.10} \scriptsize{($\pm0.12$)} & 71.60 \scriptsize{($\pm1.91$)} & 62.26 \scriptsize{($\pm2.62$)} & \cellcolor{gray!25} 74.93 \scriptsize{($\pm8.68$)} & \underline{71.08} \scriptsize{($\pm2.09$)} & 77.15 \scriptsize{($\pm0.42$)} & 70.88 \scriptsize{($\pm4.07$)} & 59.39 \scriptsize{($\pm3.31$)} & 58.60 \scriptsize{($\pm4.20$)} & \cellcolor{gray!25} 67.42 \scriptsize{($\pm8.10$)} & 66.67 \scriptsize{($\pm2.27$)} & \textbf{71.47} \scriptsize{($\pm1.67$)} & 63.11 \scriptsize{($\pm1.48$)} & 62.26 \scriptsize{($\pm1.50$)} & 68.13 \scriptsize{($\pm2.00$)} & \cellcolor{gray!25} 66.33 \scriptsize{($\pm3.76$)} & 70.45 \scriptsize{($\pm8.21$)} \\
\bottomrule
\end{tabular}
        
        \end{adjustbox}
    }
    \caption{Monolingual CND accuracy across LM families (ES=encoder-style; Verbose and Instruct denote prompt types). 
    \textbf{Bold} and \underline{underlined} scores mark the best and second-best results, respectively. Parentheses indicate standard deviation. 
    Fine-tuned encoder-based PLMs and decoder-based SLMs substantially outperform LLMs across language resource levels.}
    \label{tab:all_acc}
\end{table*}

\subsection*{Experiment 2: Multilingual CND Benchmark}

\textbf{SLMs considerably outperform LLMs on CND across most languages.}
Table~\ref{tab:all_acc} reports accuracy benchmarking results for all LMs in our study. 
For SLMs, we report results for the ES variant. 
Across resource levels, SLMs outperform LLMs by an average margin of 8–12\%. 
Although LLMs achieve competitive results for Turkish and Bulgarian, ranking second best and best, SLMs consistently outperform LLMs even on most individual languages across all resource levels.
LLM performance degrades as language resources decrease. 
The best-performing LLM, Gemini~2.5, achieves only 59.44\% accuracy on low-resource languages, compared to 66.85\% for Qwen. 
A similar trend appears for SLMs and PLMs when moving from high- to medium-resource languages.
However, performance remains relatively stable from medium- to low-resource languages (e.g., Llama achieves 66.46\% accuracy on medium-resource languages and 66.85\% on low-resource languages).
With detection accuracy ranging between 53–70\% for medium- and low-resource languages, these results highlight the need for further research focused on low-resource language communities. 
We note that performance differences across languages stem from both language-specific factors and variations in claim distributions.\footnote{A notable outlier is German. We provide a qualitative analysis of misclassified instances in Appendix~\ref{app:results}.} 
We are able to analyze these effects in the X-CND experiments presented in the next section.
\textbf{Encoder-based PLMs remain a competitive baseline for multilingual CND.}
Encoder-based PLMs continue to provide strong baselines for multilingual CND. 
mBERT achieves the highest average accuracy on medium- and low-resource languages, with moderate gains of 1–2 percentage points over SLMs. 
Compared to XLM-RoBERTa, including its larger variants, SLMs achieve higher accuracy by 1–4 percentage points.\footnote{This difference is partly attributable to XLM-RoBERTa failing on Uzbek. Despite extensive exploration of random seeds and hyperparameter configurations, the model continued to fail on this language.} 
mBERT’s strong performance may stem from its exclusive pretraining on multilingual Wikipedia data, which provides near-perfect domain alignment.

\subsection*{Experiment 3: X-CND}

\begin{figure}[H]
    \centering
        \includegraphics[width=\linewidth]{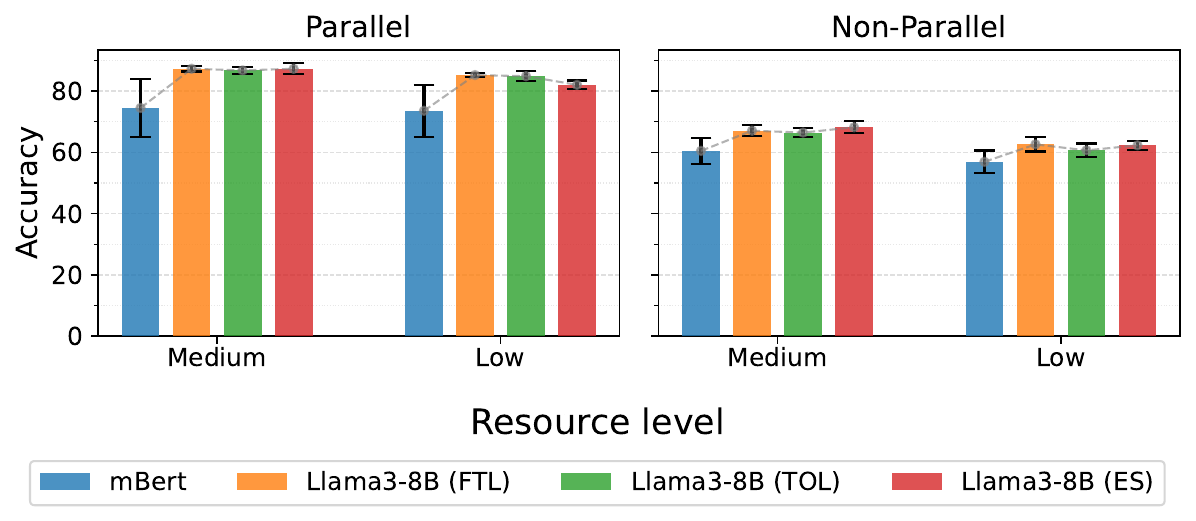}
    \caption{Zero-shot CND accuracy by test data setting, averaged across language resource groups. 
    SLMs across SFT variants exhibit strong zero-shot capabilities.}
    \label{fig:zero_shot_acc}
\end{figure}

\textbf{Fine-tuned models demonstrate strong zero-shot capabilities, but performance drops by up to 20 percentage points under real-world conditions. SLMs outperform PLMs by about 10 percentage points.}
Figure~\ref{fig:zero_shot_acc} reports accuracy averaged across test sets and language resource levels. 
On parallel data, fine-tuned models exhibit strong zero-shot generalization: SLMs achieve up to 87\% accuracy, while mBERT reaches approximately 74\% on medium-resource languages, with only marginal declines for low-resource languages. 
However, on non-parallel data, accuracy drops by 15–20 percentage points across all models. 
Across all settings, SLMs outperform mBERT by 5–10 percentage points and exhibit lower variance, as indicated by the error bars. 
Notably, performance differences across SFT variants remain small (within 2 percentage points), suggesting that the choice of SFT variant matters less for zero-shot transfer than for monolingual CND.
\textbf{QLoRA-fine-tuned SLMs require slower learning to achieve few-shot gains.}
For few-shot transfer, we continue fine-tuning the English-trained models on small amounts of target-language data, following \citet{lauscher2020}. 
However, we find that few-shot adaptation does not reliably improve SLM performance when using the higher learning rates commonly adopted in QLoRA fine-tuning~\citep{dettmers2023}. We therefore treat the learning rate as a tunable hyperparameter during this second-stage adaptation. Appendix~\ref{app:results} reports the full results and shows that substantially lower learning rates are necessary to obtain consistent few-shot gains for QLoRA-fine-tuned SLMs.
\begin{table*}[t]
\centering

    {
        \renewcommand{\arraystretch}{0.9}
        \begin{adjustbox}{max width=\linewidth}\centering

\begin{tabular}{lccccccccccccc}
 & & \multicolumn{6}{c}{\textbf{Medium-Resource}} & \multicolumn{6}{c}{\textbf{Low-Resource}} \\ 
\cmidrule(lr){3-8} \cmidrule(lr){9-14} 
\textbf{Model} & \textbf{Shots} & \textbf{uk} & \textbf{ro} & \textbf{id} & \textbf{bg} & \textbf{uz} & $\Delta$ \textbf{Avg} & \textbf{no} & \textbf{az} & \textbf{mk} & \textbf{hy} & \textbf{sq} & $\Delta$ \textbf{Avg}  \\ 
\toprule
\multirow{5}{*}{Llama3 (FTL)} & \cellcolor{gray!25} 0 &  60.45 \scriptsize{($\pm$0.70)}&  74.38 \scriptsize{($\pm$2.02)}&  66.87 \scriptsize{($\pm$1.95)}&  62.53 \scriptsize{($\pm$1.74)}&  71.13 \scriptsize{($\pm$3.00)} &  &  58.40 \scriptsize{($\pm$1.78)}&  64.86 \scriptsize{($\pm$1.86)}&  58.75 \scriptsize{($\pm$1.21)}&  62.88 \scriptsize{($\pm$4.22)}&  68.07 \scriptsize{($\pm$2.57)} &  \\
 & $\Delta$ \; 50&  \posneg{1.14} \scriptsize{($\pm$0.83)}&  \posneg{-0.54} \scriptsize{($\pm$0.94)}&  \posneg{0.87} \scriptsize{($\pm$1.75)}&  \posneg{-0.47} \scriptsize{($\pm$2.05)}&  \posneg{-0.33} \scriptsize{($\pm$2.60)} &  \posneg{0.13} &  \posneg{-0.40} \scriptsize{($\pm$1.39)}&  \posneg{0.07} \scriptsize{($\pm$1.51)}&  \posneg{1.88} \scriptsize{($\pm$0.91)}&  \posneg{0.00} \scriptsize{($\pm$2.73)}&  \posneg{-0.40} \scriptsize{($\pm$3.41)} &  \posneg{0.23} \\
 & $\Delta$ 100&  \posneg{-0.47} \scriptsize{($\pm$1.73)}&  \posneg{-0.67} \scriptsize{($\pm$0.30)}&  \posneg{2.00} \scriptsize{($\pm$1.36)}&  \posneg{1.20} \scriptsize{($\pm$0.80)}&  \posneg{0.60} \scriptsize{($\pm$2.02)} &  \posneg{0.53} &  \posneg{0.60} \scriptsize{($\pm$1.60)}&  \posneg{-0.80} \scriptsize{($\pm$1.10)}&  \posneg{1.61} \scriptsize{($\pm$2.13)}&  \posneg{-0.76} \scriptsize{($\pm$3.03)}&  \posneg{-1.13} \scriptsize{($\pm$2.50)} &  \posneg{-0.10} \\
 & $\Delta$ 250&  \posneg{-0.94} \scriptsize{($\pm$1.06)}&  \posneg{-0.14} \scriptsize{($\pm$1.34)}&  \posneg{3.07} \scriptsize{($\pm$2.09)}&  \posneg{0.47} \scriptsize{($\pm$0.46)}&  \posneg{-0.20} \scriptsize{($\pm$1.29)} &  \posneg{0.45} &  \posneg{0.33} \scriptsize{($\pm$0.46)}&  \posneg{-0.47} \scriptsize{($\pm$1.50)}&  \posneg{-1.14} \scriptsize{($\pm$1.83)}&  \posneg{0.00} \scriptsize{($\pm$2.27)}&  \posneg{0.20} \scriptsize{($\pm$2.21)} &  \posneg{-0.21} \\
 & $\Delta$ 500&  \posneg{0.13} \scriptsize{($\pm$2.84)}&  \posneg{-0.66} \scriptsize{($\pm$0.42)}&  \posneg{2.61} \scriptsize{($\pm$1.63)}&  \posneg{0.94} \scriptsize{($\pm$0.50)}&  \posneg{0.51} \scriptsize{($\pm$2.62)} &  \posneg{0.71} &  \posneg{0.27} \scriptsize{($\pm$1.10)}&  \posneg{-0.20} \scriptsize{($\pm$2.25)}&  \posneg{0.94} \scriptsize{($\pm$1.21)}&  \posneg{5.56} \scriptsize{($\pm$1.91)}&  \posneg{0.33} \scriptsize{($\pm$2.46)} &  \posneg{1.38} \\
\cmidrule(lr){2-14}
\multirow{5}{*}{Llama3 (TOL)} & \cellcolor{gray!25} 0 &  57.78 \scriptsize{($\pm$1.97)}&  73.79 \scriptsize{($\pm$0.73)}&  68.14 \scriptsize{($\pm$1.97)}&  62.73 \scriptsize{($\pm$1.84)}&  69.80 \scriptsize{($\pm$1.11)} &  &  58.07 \scriptsize{($\pm$1.03)}&  64.89 \scriptsize{($\pm$1.67)}&  57.95 \scriptsize{($\pm$2.10)}&  56.30 \scriptsize{($\pm$5.54)}&  66.07 \scriptsize{($\pm$0.90)} &  \\
 & $\Delta$ \; 50&  \posneg{0.63} \scriptsize{($\pm$0.99)}&  \posneg{0.20} \scriptsize{($\pm$0.29)}&  \posneg{-1.10} \scriptsize{($\pm$0.99)}&  \posneg{0.20} \scriptsize{($\pm$0.57)}&  \posneg{0.80} \scriptsize{($\pm$1.41)} &  \posneg{0.15} &  \posneg{-0.27} \scriptsize{($\pm$2.51)}&  \posneg{-0.47} \scriptsize{($\pm$1.93)}&  \posneg{0.94} \scriptsize{($\pm$0.76)}&  \posneg{0.05} \scriptsize{($\pm$5.05)}&  \posneg{-0.07} \scriptsize{($\pm$3.22)} &  \posneg{0.04} \\
 & $\Delta$ 100&  \posneg{0.73} \scriptsize{($\pm$1.42)}&  \posneg{0.10} \scriptsize{($\pm$2.42)}&  \posneg{0.30} \scriptsize{($\pm$0.71)}&  \posneg{0.60} \scriptsize{($\pm$0.28)}&  \posneg{1.20} \scriptsize{($\pm$1.13)} &  \posneg{0.59} &  \posneg{-0.47} \scriptsize{($\pm$2.03)}&  \posneg{-0.73} \scriptsize{($\pm$2.64)}&  \posneg{1.01} \scriptsize{($\pm$2.44)}&  \posneg{0.29} \scriptsize{($\pm$4.78)}&  \posneg{0.73} \scriptsize{($\pm$3.83)} &  \posneg{0.17} \\
 & $\Delta$ 250&  \posneg{1.54} \scriptsize{($\pm$0.57)}&  \posneg{1.11} \scriptsize{($\pm$0.14)}&  \posneg{-1.10} \scriptsize{($\pm$1.84)}&  \posneg{1.20} \scriptsize{($\pm$1.98)}&  \posneg{-1.80} \scriptsize{($\pm$1.13)} &  \posneg{0.19} &  \posneg{-0.47} \scriptsize{($\pm$1.97)}&  \posneg{-1.56} \scriptsize{($\pm$1.71)}&  \posneg{0.60} \scriptsize{($\pm$1.01)}&  \posneg{2.94} \scriptsize{($\pm$2.14)}&  \posneg{1.27} \scriptsize{($\pm$3.21)} &  \posneg{0.56} \\
 & $\Delta$ 500&  \posneg{1.14} \scriptsize{($\pm$0.28)}&  \posneg{1.11} \scriptsize{($\pm$0.71)}&  \posneg{0.20} \scriptsize{($\pm$3.12)}&  \posneg{-0.50} \scriptsize{($\pm$2.41)}&  \posneg{1.30} \scriptsize{($\pm$2.97)} &  \posneg{0.65} &  \posneg{0.20} \scriptsize{($\pm$2.14)}&  \posneg{-1.16} \scriptsize{($\pm$1.51)}&  \posneg{0.67} \scriptsize{($\pm$1.18)}&  \posneg{4.53} \scriptsize{($\pm$2.67)}&  \posneg{1.67} \scriptsize{($\pm$1.92)} &  \posneg{1.18} \\
\cmidrule(lr){2-14}
\multirow{5}{*}{Llama3 (ES)} & \cellcolor{gray!25} 0 &  64.93 \scriptsize{($\pm$1.51)}&  75.20 \scriptsize{($\pm$1.40)}&  71.68 \scriptsize{($\pm$1.75)}&  64.40 \scriptsize{($\pm$3.64)}&  64.73 \scriptsize{($\pm$1.62)} &  &  60.80 \scriptsize{($\pm$1.11)}&  65.73 \scriptsize{($\pm$2.23)}&  61.90 \scriptsize{($\pm$0.81)}&  53.71 \scriptsize{($\pm$1.75)}&  69.33 \scriptsize{($\pm$1.70)} &  \\
 & $\Delta$ \; 50&  \posneg{0.53} \scriptsize{($\pm$1.21)}&  \posneg{0.60} \scriptsize{($\pm$0.92)}&  \posneg{-0.40} \scriptsize{($\pm$0.70)}&  \posneg{2.54} \scriptsize{($\pm$1.64)}&  \posneg{2.60} \scriptsize{($\pm$5.25)} &  \posneg{1.18} &  \posneg{0.20} \scriptsize{($\pm$1.11)}&  \posneg{-0.67} \scriptsize{($\pm$1.33)}&  \posneg{-1.07} \scriptsize{($\pm$2.07)}&  \posneg{5.88} \scriptsize{($\pm$1.45)}&  \posneg{0.73} \scriptsize{($\pm$1.86)} &  \posneg{1.01} \\
 & $\Delta$ 100&  \posneg{-0.20} \scriptsize{($\pm$1.22)}&  \posneg{-0.13} \scriptsize{($\pm$0.51)}&  \posneg{1.27} \scriptsize{($\pm$1.64)}&  \posneg{2.07} \scriptsize{($\pm$2.55)}&  \posneg{5.07} \scriptsize{($\pm$3.94)} &  \posneg{1.61} &  \posneg{-0.13} \scriptsize{($\pm$1.10)}&  \posneg{-0.40} \scriptsize{($\pm$1.22)}&  \posneg{-0.34} \scriptsize{($\pm$1.41)}&  \posneg{3.54} \scriptsize{($\pm$0.61)}&  \posneg{0.73} \scriptsize{($\pm$1.75)} &  \posneg{0.68} \\
 & $\Delta$ 250&  \posneg{0.53} \scriptsize{($\pm$1.17)}&  \posneg{0.13} \scriptsize{($\pm$0.62)}&  \posneg{2.61} \scriptsize{($\pm$1.33)}&  \posneg{1.67} \scriptsize{($\pm$2.21)}&  \posneg{5.33} \scriptsize{($\pm$4.24)} &  \posneg{2.06} &  \posneg{1.27} \scriptsize{($\pm$1.96)}&  \posneg{0.40} \scriptsize{($\pm$2.02)}&  \posneg{-2.08} \scriptsize{($\pm$1.90)}&  \posneg{4.48} \scriptsize{($\pm$1.67)}&  \posneg{1.80} \scriptsize{($\pm$0.95)} &  \posneg{1.17} \\
 & $\Delta$ 500&  \posneg{3.21} \scriptsize{($\pm$0.80)}&  \posneg{-0.54} \scriptsize{($\pm$0.93)}&  \posneg{2.94} \scriptsize{($\pm$1.81)}&  \posneg{1.40} \scriptsize{($\pm$1.82)}&  \posneg{7.73} \scriptsize{($\pm$3.60)} &  \posneg{2.95} &  \posneg{2.33} \scriptsize{($\pm$1.15)}&  \posneg{2.07} \scriptsize{($\pm$1.60)}&  \posneg{-1.14} \scriptsize{($\pm$1.92)}&  \posneg{4.81} \scriptsize{($\pm$1.40)}&  \posneg{2.60} \scriptsize{($\pm$1.67)} &  \posneg{2.13} \\
\cmidrule(lr){2-14}
\multirow{5}{*}{mBert} & \cellcolor{gray!25} 0 &  55.04 \scriptsize{($\pm$1.33)}&  65.79 \scriptsize{($\pm$6.87)}&  62.66 \scriptsize{($\pm$4.32)}&  54.98 \scriptsize{($\pm$2.32)}&  63.93 \scriptsize{($\pm$5.91)} &  &  54.00 \scriptsize{($\pm$2.16)}&  58.93 \scriptsize{($\pm$4.35)}&  53.66 \scriptsize{($\pm$2.82)}&  56.65 \scriptsize{($\pm$3.33)}&  61.33 \scriptsize{($\pm$5.83)} &  \\
 & $\Delta$ \; 50&  \posneg{-0.73} \scriptsize{($\pm$0.53)}&  \posneg{6.05} \scriptsize{($\pm$1.16)}&  \posneg{-2.20} \scriptsize{($\pm$5.47)}&  \posneg{2.94} \scriptsize{($\pm$4.74)}&  \posneg{2.47} \scriptsize{($\pm$6.30)} &  \posneg{1.70} &  \posneg{-1.60} \scriptsize{($\pm$3.86)}&  \posneg{3.27} \scriptsize{($\pm$0.60)}&  \posneg{2.28} \scriptsize{($\pm$3.43)}&  \posneg{-1.60} \scriptsize{($\pm$2.06)}&  \posneg{7.93} \scriptsize{($\pm$3.63)} &  \posneg{2.06} \\
 & $\Delta$ 100&  \posneg{0.73} \scriptsize{($\pm$0.61)}&  \posneg{6.59} \scriptsize{($\pm$5.09)}&  \posneg{3.87} \scriptsize{($\pm$5.95)}&  \posneg{2.47} \scriptsize{($\pm$3.25)}&  \posneg{0.60} \scriptsize{($\pm$4.98)} &  \posneg{2.85} &  \posneg{-0.40} \scriptsize{($\pm$2.91)}&  \posneg{1.33} \scriptsize{($\pm$2.53)}&  \posneg{3.49} \scriptsize{($\pm$1.22)}&  \posneg{2.07} \scriptsize{($\pm$4.67)}&  \posneg{10.27} \scriptsize{($\pm$3.86)} &  \posneg{3.35} \\
 & $\Delta$ 250&  \posneg{1.67} \scriptsize{($\pm$1.31)}&  \posneg{8.80} \scriptsize{($\pm$1.12)}&  \posneg{3.81} \scriptsize{($\pm$5.62)}&  \posneg{4.28} \scriptsize{($\pm$2.69)}&  \posneg{2.87} \scriptsize{($\pm$7.30)} &  \posneg{4.28} &  \posneg{0.33} \scriptsize{($\pm$0.42)}&  \posneg{3.27} \scriptsize{($\pm$1.74)}&  \posneg{1.88} \scriptsize{($\pm$1.26)}&  \posneg{4.74} \scriptsize{($\pm$2.45)}&  \posneg{4.00} \scriptsize{($\pm$5.35)} &  \posneg{2.84} \\
 & $\Delta$ 500&  \posneg{3.27} \scriptsize{($\pm$4.34)}&  \posneg{3.76} \scriptsize{($\pm$4.74)}&  \posneg{-0.33} \scriptsize{($\pm$2.84)}&  \posneg{3.34} \scriptsize{($\pm$2.11)}&  \posneg{6.27} \scriptsize{($\pm$8.62)} &  \posneg{3.26} &  \posneg{3.00} \scriptsize{($\pm$1.06)}&  \posneg{4.00} \scriptsize{($\pm$1.86)}&  \posneg{1.88} \scriptsize{($\pm$1.05)}&  \posneg{5.88} \scriptsize{($\pm$3.11)}&  \posneg{11.47} \scriptsize{($\pm$1.93)} &  \posneg{5.24} \\
\bottomrule
\end{tabular}
        
        \end{adjustbox}
    }
        
    \caption{Few-shot CND accuracy on non-parallel data by model and number of shots. 
    $\Delta$ denotes gains over zero-shot accuracy. Parentheses report standard deviations. 
    All models are fine-tuned on English claims only. 
    Only Llama (ES) and mBERT benefit from few-shot fine-tuning. 
    Notably, Llama (ES) outperforms LLMs (Table~\ref{tab:all_acc}) even with no or minimal target-language supervision.}
    \label{tab:few_acc}
\end{table*}
\textbf{Encoder-style SFT yields the most reliable few-shot gains among SLMs, while mBERT benefits the most from few-shot tuning overall.}
Table~\ref{tab:few_acc} reports few-shot accuracy results on non-parallel data. 
For each model, we report results obtained with the optimal learning rate, as described above. 
Among the SFT variants, only ES tuning exhibits consistent and substantial gains from few-shot adaptation. 
In contrast, continued fine-tuning with FTL and TOL proves largely ineffective on average, yielding only marginal improvements (less than 1\% accuracy). 
By comparison, mBERT consistently benefits from few-shot tuning, particularly on lower-resource languages (e.g., a gain of 11.47 percentage points on Albanian with only 500 samples). 
Overall, few-shot tuning improves mBERT more than it improves ES-tuned SLMs.

\textbf{English-only ES-fine-tuned SLMs surpass LLMs, even with zero or minimal target-language supervision.}
Compared to the monolingual evaluation of Experiment~2, ES-tuned Llama outperforms prompted LLMs on nearly all target languages, even under zero-shot or minimal target-language supervision. 
For example, on Uzbek, Llama (ES) achieves 64.73\% accuracy in the zero-shot setting and 72.46\% with $k{=}500$ target-language examples, compared to 67.20\% for the best-performing LLM (Gemini~2.5, few-shot). 
This result holds for all languages except Armenian.
Notably, Llama is not only considerably smaller but also fine-tuned exclusively on English claims for this cross-lingual task.

\subsection*{Are results generalizable to non-FA articles?}

\begin{table}[t]
\centering
    \begin{adjustbox}{max width=1\linewidth}\centering

    \begin{tabular}{lcccccc}
 & \textbf{en} & \textbf{it} & \textbf{uk} & \textbf{ro} & \textbf{no} & \textbf{az} \\
\toprule
RA & 62.63 & 61.00 & 71.43 & 64.65 & 70.00 & 60.61 \\
RA (corrected labels) & 86.87 & 70.00 & 72.45 & 68.68 & 68.00 & 64.65 \\
\hdashline
FA baseline & 89.65 & 74.83 & 72.01 & 73.59 & 65.73 & 71.47 \\
\midrule
RA Distribution (0/1) & 14/85 & 11/89 & 6/92 & 7/92 & 8/92 & 2/97 \\
\bottomrule
\end{tabular}
    
    \end{adjustbox}
        
\caption{Monolingual CND accuracy on the Random Articles (RA) test set ($n=100$ per language) using Llama~3 (ES). Due to under-referencing in random articles, we manually corrected the labels.}
    \label{tab:general_acc}
\end{table}

A potential concern is that models trained on claims from FAs may not generalize to non-FA articles, for example due to less standardized editorial styles.
To assess this, we collected claims from randomly sampled Wikipedia articles (RA), selecting two languages per resource level, and constructed balanced test sets ($n=100$ per language).
Because non-FA articles are often under-referenced, leading to a high number of false negatives, we manually corrected all 600 labels.

Table~\ref{tab:general_acc} reports results for (i) the originally sampled claims, (ii) the label-corrected claims, and (iii) the FA baseline from Table~\ref{tab:all_acc}.
We find that, for most languages, model performance on RA claims is comparable to, or slightly lower than, performance on FA claims.
The discrepancy between results on uncorrected and corrected RA labels is explained by the high number of false negatives in the original annotations.
After correction, the label distribution becomes more imbalanced; nevertheless, model performance remains close to FA-level detection.
Overall, these results indicate that models trained on FA data generalize well to claims from randomly sampled Wikipedia articles, supporting their practical utility for the Wikipedia community.

\section{Discussion}

\textbf{Choosing the right SFT variant for SLMs can yield substantial performance gains for multilingual CND (RQ1).}
Experiment~1 shows that ES fine-tuning consistently outperforms generative variants for monolingual CND, although TOL can benefit certain model–language combinations. 
Recent work explores adaptations of decoder-based models for encoder-like tasks~\cite{xu2025,suganthan2025} and demonstrates that encoder-style training can be effective~\cite{bolton2024,li2025}. 
Our results align with these findings and show that ES training performs well for CND across languages. 
For practitioners, ES tuning requires only minimal modifications to the base model while delivering reliable performance gains. 
In contrast, under generative SFT, the optimal variant depends on the specific model-language combination. 

\textbf{Supervised models substantially outperform LLMs (RQ2).}
Both fine-tuned PLMs and SLMs consistently outperform LLMs on multilingual CND. 
This result extends prior findings on English CWD~\cite{majer2024,bell2025,li2024} to the multilingual setting and aligns with evidence from broader NLP classification tasks~\cite[e.g.,][]{bucher2024}. 
Our findings show that supervised models are not only more compact and cost-effective than LLMs but also deliver substantially stronger performance for CND. 
Their relatively low hardware requirements make them particularly well suited for deployment in resource-constrained environments such as Wikipedia. 
Nonetheless, the substantial performance variation across resource levels (53-75\%) indicates that multilingual CND remains an open and challenging research problem.

\textbf{Encoder-style SLMs exhibit strong zero-shot cross-lingual transfer and outperform LLMs with little to no target-language supervision (RQ3).}
Across SFT variants, SLMs demonstrate robust zero-shot generalization on parallel data. 
However, performance drops by up to 20 percentage points under real-world conditions, exposing a critical limitation for practical deployment. 
This finding highlights the need for future X-CND research to develop more robust strategies that address simultaneous language and claim-distribution shifts.

Crucially, ES-fine-tuned SLMs trained exclusively on English claims outperform prompted LLMs even with no or minimal target-language adaptation. 
This result has direct implications for real-world deployment: cross-lingual transfer of specialized SLMs trained on high-resource languages emerges as a viable strategy when target-language data is scarce. 
For practitioners, this approach offers an affordable deployment pathway to support low-resource communities in CND.

\section{Conclusion}
We introduce MCN, a multilingual citation needed detection (CND) dataset that covers Wikipedia claims from 18 linguistically diverse languages across three resource levels. 
We study decoder-based small language models (SLMs) as alternatives to LLMs for CND, specialized through supervised fine-tuning. 
We demonstrate that SLMs outperform LLMs on monolingual CND across languages and that encoder-style tuning enables strong zero- and few-shot cross-lingual CND transfer. 
Our findings have important implications for supporting low-resource language communities in CND.

\section*{Limitations}
Our work has several limitations.

\paragraph{Language selection concentrates on Indo-European languages.}
Our selected languages are 70\% Indo-European. While we vary language groups and also cover other families, future work can focus on language families such as Sino-Tibetan or Semitic. 

\paragraph{Insights into SLM performance are limited to QLoRA-based SFT.}
To ensure practical relevance, we fine-tune SLMs exclusively using QLoRA.
As a result, our insights into the effects of SFT strategies and the cross-lingual capabilities of SLMs for CND are limited to this PEFT setting.
Performance dynamics may differ under alternative parameter-efficient fine-tuning approaches.
Nevertheless, QLoRA is particularly well suited to this study due to its simplicity and efficiency under quantisation.
Moreover, we restrict our analysis to 8B-parameter SLMs.
Future work should explore a broader range of model sizes and PEFT strategies to assess the generality of our findings.


\paragraph{Limited generalizability to other domains.}
CWD is inherently domain-dependent, as domain-specific criteria determine what constitutes a check-worthy claim.
Accordingly, our findings are limited to the Wikipedia domain and may not directly transfer to other common CWD settings, such as political debates.
Nevertheless, the strong monolingual and cross-lingual CND performance observed on the challenging Wikipedia domain suggests that comparable gains may be achievable in other application domains.

\paragraph{Real-world citation practices deviate from labeling heuristics.}

We label claims as requiring a citation or not based on the presence or absence of inline citations, a heuristic following prior CND work~\cite{redi2019,halitaj2024,wmf_mcd}.
Nevertheless, in practice, real-world editorial practices deviate from that heuristic.
For example, a citation may appear only in the final paragraph of a multi-paragraph passage when it is clear to support all paragraphs.
We illustrate this challenge for German Wikipedia in Appendix~\ref{app:results}.
We believe that devising more robust labeling heuristics, for example through the use of soft labels, is an important direction for future work.

\section*{Ethics Statement}

\paragraph{Biases}
Although our language selection spans a wide range of language families and writing systems, it is skewed toward Indo-European languages.
In addition, Featured Wikipedia articles, while covering diverse topics, tend to emphasize subjects that are of particular interest to the corresponding language communities at a given time, which may introduce topical biases.

\paragraph{Potential Harms}
We are not aware of any direct harms arising from our dataset or experiments.
Our work is motivated by the goal of supporting Wikipedia editors, particularly in low-resource language communities, by improving automated citation needed detection.

\paragraph{Licensing}
We collect only publicly available Wikipedia content, which is licensed under CC BY-SA.
We release all collected data, including translated text obtained via the Google Translate API, under the same CC BY-SA license to support future research.

\paragraph{Reproducibility}
We provide full details required to reproduce our results in Appendix~\ref{app:experimental_details}, including hyperparameters, hardware requirements, and fine-tuning procedures.
All models are trained three times with different random seeds, and we report mean performance and standard deviations to enhance the reliability of our findings.

\paragraph{AI Assistants}
We use LLMs to proofread our writing by prompting it to correct grammar and spelling.

\bibliography{custom}

\clearpage

\begin{table*}[t]
    \centering
    \footnotesize
    \begin{adjustbox}{width=.8\textwidth}
  
    \begin{tabular}{l p{0.18\linewidth} p{0.14\linewidth} p{0.22\linewidth} p{0.22\linewidth} p{0.22\linewidth} c}
    \textbf{Language} & \textbf{Title}  & \textbf{Section} & \textbf{Claim} & \textbf{Previous Sentence} &  \textbf{Subsequent Sentence} & \textbf{Label} \\
    \toprule
    \multirow{2}{*}{en} & Spanish conquest of Guatemala & San Marcos: Province of Tecusitlán and Lacandón & De León renamed the city as San Pedro Sacatepéquez in honour of his friar, Pedro de Angulo. & De León marched to a Maya city named Quezalli by his Nahuatl-speaking allies with a force of fifty Spaniards; his Mexican allies also referred to the city by the name Sacatepequez. & The Spanish founded a village nearby at Candacuchex in April that year, renaming it as San Marcos. & 1 \\
    
    \cmidrule(lr){2-7}
    
     & Aliens (film) & Plot & Ripley meets the Colonial Marines aboard the spaceship Sulaco but distrusts their android, Bishop, because the Nostromo's android, Ash, betrayed its crew to protect the alien on company orders. & Still traumatized by her alien encounter, she agrees on the condition that they exterminate the creatures. & - & 0 \\
    
    \midrule
    
    \multirow{2}{*}{it} & Francesco Guccini & Il debutto (1967–1971) & In autunno iniziò le registrazioni di un nuovo disco, e così a undici mesi da Due anni dopo fu pubblicato L'isola non trovata. & - & Il titolo dell'album, che è anche quello di una canzone, è un riferimento a Guido Gozzano; altra citazione letteraria presente nel disco fu quella di J. D. Salinger in La collina. & 1 \\

    \cmidrule(lr){2-7}
    
     & Alchimia & Alchimia indiana & Gianluca Magi nota come: & L'alchimia giocò un ruolo di spicco fin dalle origini del pensiero indiano. & $<$\textit{citation}$>$ & 0 \\
    \bottomrule
\end{tabular}


    \end{adjustbox}
    \caption{Dataset Examples.}
    \label{tab:data_examples}
\end{table*}

\begin{figure*}[t]
    \centering
        \begin{adjustbox}{max width=\textwidth}
            \includegraphics{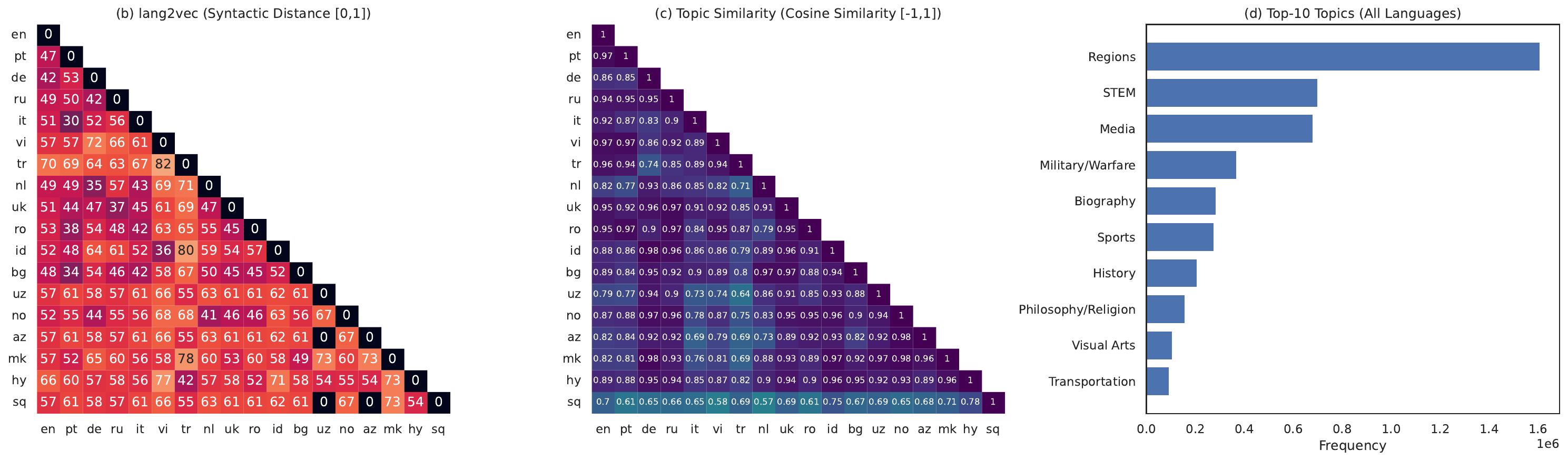}
        \end{adjustbox}
    \caption{Descriptive statistics.}
    \label{fig:desc}
\end{figure*}

\appendix

\section{Data}
\label{app:data}

\subsection{Data collection}
We collect claims from Featured Articles, which represent the highest-quality articles on Wikipedia.\footnote{\url{https://en.wikipedia.org/wiki/Wikipedia:Featured_articles}}
For languages other than German, English, and Russian, where Featured Articles are abundant, we additionally include Good Articles\footnote{\url{https://en.wikipedia.org/wiki/Wikipedia:Good_articles}} to increase data coverage.
Like Featured Articles, Good Articles are required to contain factually accurate and verifiable information, although they are typically less comprehensive.

For all selected articles across languages, we query the MediaWiki API\footnote{\url{https://www.mediawiki.org/wiki/MediaWiki}} to obtain their full HTML representations.
We then parse each article into paragraphs (excluding the lead section) and split paragraphs into individual claims (i.e., sentences) using Stanza.\footnote{\url{https://stanfordnlp.github.io/stanza/}}
Cleaning primarily involves correcting formatting issues (e.g., spaces before punctuation), removing extraneous whitespace, and discarding claims shorter than 15 characters, among other steps.
These procedures are applied consistently to all retrieved elements, including the topic, the claim itself, and the preceding and following sentences.

Because artifacts introduced during parsing or preprocessing may create spurious correlations and bias our results, we additionally conduct a manual data quality check.

\subsection{Data Labelling}
We largely follow prior work in labeling claims~\cite{redi2019,halitaj2024}, with one minor modification.
Consistent with \citet{redi2019}, we label a claim as not requiring a citation only if it appears in a paragraph that contains no citations.
Conversely, if a target claim does not contain an inline citation but appears in a paragraph where other claims are cited, we label it as citation needed.
This design choice is motivated by our observation that, in practice, claims without inline citations are often implicitly supported by citations in adjacent sentences, either preceding or following the claim, even when no explicit inline citation is present.
We provide a qualitative analysis of the general limitations of this heuristic on the German Wikipedia in Appendix~\ref{app:results}.

\subsection{Descriptive Statistics}

\paragraph{Dataset examples}
Table~\ref{tab:data_examples} illustrates two representative examples in English and Italian from our dataset.
In the English examples, the first claim contains multiple factual assertions (e.g., the naming of a city and the reason for its name) related to a historical event, each of which requires a citation to verify that these events occurred.
The second English example represents a canonical case of a claim that does not require a citation under Wikipedia’s verifiability policy, as readers can reasonably infer that the source of the information is the film itself.\footnote{\url{https://en.wikipedia.org/wiki/Wikipedia:When_to_cite}}

For the Italian examples, the first claim translates to “In the fall, he began recording a new album, and thus, eleven months after \textit{Due anni dopo}, \textit{L’isola non trovata} was released.”
This claim refers to the recording and release of a new album and therefore requires a reliable external source to substantiate the information.
The second Italian example introduces a citation to Gianluca Magi. 
While the cited text itself requires a citation, the surrounding sentence does not, illustrating a case where the claim is not citation-worthy.

\paragraph{Descriptive statistics}
Figure~\ref{fig:desc} presents descriptive statistics on syntactic distance (a), cross-lingual topic similarity (b), and the ten most frequent topics across languages (c).
Our language selection covers a broad range of linguistic diversity, as measured by syntactic distance derived from lang2vec vectors.\footnote{\url{https://github.com/antonisa/lang2vec}}
We cover syntactic distance relationships ranging from distant language pairs (e.g., Indonesian and Albanian) to closely related ones (e.g., Ukrainian and Russian).

Panel~(b) illustrates cross-lingual topic similarity measured via cosine similarity between topic vectors.
Overall, Wikipedia language editions exhibit relatively high topical overlap.
A notable exception is Albanian (\textit{sq}), for which topic similarity is lower, likely due to the substantially smaller number of high-quality Albanian Wikipedia articles in our sample.

Finally, Panel~(c) shows the ten most frequent topics across languages, indicating that claims span a diverse set of Wikipedia categories, including Regions, STEM, and Media.

\subsection{Manual Quality Evaluation of MCN}
\label{app:manual_eval}

\begin{table}[H]
    \centering
    \begin{adjustbox}{width=\linewidth}
    
       \begin{tabular}{lccc}
    \textbf{Language} & \textbf{Valid Claim (\%)} & \textbf{Label Accuracy (\%)} & \textbf{Valid Context (\%)} \\
    \toprule
    \multicolumn{4}{l}{\textbf{High Resource}} \\
    \midrule
    en & 0.93 & 1.00 & 0.93 \\
    pt & 1.00 & 1.00 & 0.97 \\
    de & 0.97 & 1.00 & 0.93 \\
    ru & 0.93 & 0.90 & 0.93 \\
    it & 1.00 & 1.00 & 1.00 \\
    vi & 1.00 & 1.00 & 0.93 \\
    tr & 0.93 & 0.93 & 0.93 \\
    nl & 1.00 & 1.00 & 1.00 \\
    \midrule
    \multicolumn{4}{l}{\textbf{Medium Resource}} \\
    \midrule
    uk & 1.00 & 0.97 & 1.00 \\
    ro & 0.93 & 0.97 & 0.93 \\
    id & 0.97 & 1.00 & 0.97 \\
    bg & 0.90 & 1.00 & 0.97 \\
    uz & 1.00 & 1.00 & 1.00 \\
    \midrule
    \multicolumn{4}{l}{\textbf{Low Resource}} \\
    \midrule
    no & 1.00 & 1.00 & 1.00 \\
    az & 0.97 & 0.97 & 0.90 \\
    mk & 1.00 & 1.00 & 1.00 \\
    hy & 1.00 & 1.00 & 1.00 \\
    sq & 0.93 & 1.00 & 0.83 \\
    \bottomrule
\end{tabular}
        
    \end{adjustbox}
    
    \caption{Manual data quality evaluation. We assess 540 data (30 per language) examples for the correctness of the claim, its label, and the surrounding context.}
    \label{tab:data_eval}
\end{table}

To ensure the quality of our data, we conduct a manual evaluation to assess the correctness of the labels, extracted claims, and context on a representative sample of our data.  
We randomly sample 540 instances, balanced across languages and labels, and manually evaluate the following:  
(1) whether the claim is correctly extracted and free from parsing errors (e.g., truncated sentences);  
(2) whether the assigned citation label is accurate based on our heuristic (see above); and  
(3) whether the surrounding context is correctly retrieved and free from parsing errors.  
We conduct these checks by identifying the original claim and its surrounding context on the source Wikipedia page and then manually verifying whether the original instances match our retrieved instances.
If they do not match, we mark the corresponding metric as incorrect.

Table~\ref{tab:data_eval} presents the data quality evaluation results.
Overall, data quality is consistently high across all three criteria and across language resource levels.
Most common sources of errors are missing conclusion of certain HTML tags which lead to incorrect claims and labels, and incorrect sentence splitting.
However, we also observe discrepancies between the labeling heuristic, also used in prior work~\cite{redi2019,halitaj2024}, and real-world editorial citation practices.
We discuss this issue in more detail below.

\subsection{Translation Quality Evaluation}

\begin{table}[H]
    \centering
    \begin{adjustbox}{width=\linewidth}

    \begin{tabular}{lcccc}
    \textbf{Language} & \textbf{BLEU} & \textbf{ROUGE-1} & \textbf{ROUGE-2} & \textbf{BERT Score} \\
    \toprule
    \textit{Medium-resource}  &  &  &  &  \\
    \midrule
    uk & 50.97 & 0.62 & 0.36 & 0.97 \\
    ro & 63.05 & 0.62 & 0.29 & 0.98 \\
    id & 58.31 & 0.57 & 0.33 & 0.98 \\
    bg & 58.64 & 0.62 & 0.29 & 0.97 \\
    uz & 39.87 & 0.59 & 0.27 & 0.96 \\
    \midrule
    \textit{Low-resource}  &  &  &  &  \\
    \midrule
    no & 63.92 & 0.71 & 0.67 & 0.98 \\
    az & 42.31 & 0.53 & 0.31 & 0.96 \\
    mk & 48.93 & 0.53 & 0.29 & 0.94 \\
    hy & 46.31 & 0.67 & 0.31 & 0.97 \\
    sq & 57.90 & 0.37 & 0.14 & 0.97 \\
    \bottomrule
    \end{tabular}

    \end{adjustbox}
    \caption{Backtranslation quality evaluation.}
    \label{tab:trans_qual}
\end{table}

We conduct an extensive automatic evaluation of the translated parallel test sets used for zero-shot X-CND.
Specifically, we back-translate claims that were translated from English into the target languages back into English in order to compute standard translation quality metrics.
Automatic evaluation via back-translation is a common procedure when gold reference translations are unavailable~\cite{chatzikoumi2020}.
We use the DeepL API for back-translation.\footnote{\url{https://www.deepl.com/en/pro-api}}
We report standard machine translation metrics, including BLEU~\cite{papineni2002} and ROUGE~\cite{lin2004} as n-gram overlap measures, and BERTScore~\cite{zhang2020} to assess semantic similarity.

Table~\ref{tab:trans_qual} presents the results of the automatic translation quality evaluation.
Across all languages, all metrics indicate strong semantic preservation, with BERTScore values ranging from 0.94 to 0.98.
BLEU and ROUGE scores likewise suggest successful reconstruction, although they exhibit greater variance.
In particular, Macedonian and Uzbek show lower BLEU and ROUGE scores, likely reflecting greater typological distance from English rather than translation errors, as semantic similarity remains high.
Overall, these results provide strong evidence that translation quality is high across languages of all resource levels.

\section{Experimental Details}
\label{app:experimental_details}
We design all experiments to be runnable on a single A100 GPU (40 GB).
All experiments are implemented using PyTorch and the Hugging Face \texttt{transformers} library\footnote{https://github.com/huggingface/transformers}.
While each experiment fits on a single GPU, we run experiments on an HPC cluster with multiple GPUs to accelerate training.

\subsection{SLMs}

\begin{table}[H]
    \centering
    \begin{adjustbox}{width=.8\linewidth}
        \begin{tabular}{l l}
        \textbf{Hyperparameter} & \textbf{Value(s)} \\
        \toprule
        Epochs & $\{1,2,3,4\}$ \\
        Learning rate & $\{5\times10^{-4},\,2\times10^{-4},\,5\times10^{-5}\}$ \\
        Batch size & 16 \\
        Max gradient norm & $\{0.5,\,1.0\}$ \\
        Weight decay & 0.01 \\
        Seed & 42 \\
        \midrule
        LoRA rank ($r$) & 16 \\
        LoRA scaling ($\alpha$) & 32 \\
        LoRA dropout & 0.05 \\
        \bottomrule
        \end{tabular}
    \end{adjustbox}
    \caption{Hyperparameter grid configuration for SFT.}
    \label{tab:hps_slm}
\end{table}

We use SFT~\cite{wei2021} with QLoRA~\cite{dettmers2023} to adapt SLMs to CND.
For each language–model combination, we perform a hyperparameter grid search using the hyperparameters shown in Table~\ref{tab:hps_slm}.
In cases where we encounter out-of-memory errors, we reduce the batch size to 8 and use a gradient accumulation step of 2.
We use the \texttt{peft}\footnote{\url{https://github.com/huggingface/peft}} library to implement QLoRA.

Preliminary experiments with different prompts revealed only marginal performance differences.
Therefore, we restrict our experiments to a single prompt.
We use the simple instruction prompt shown below for all experiments involving SLMs:

\begin{lstlisting}[language={},basicstyle=\ttfamily\small]
Instruct Prompt
----------------
You are a multilingual classifier.
Decide whether the given {lang} claim requires a citation.
Use 1 if the claim needs a citation. Use 0 if the claim does not need a citation.

Return only JSON in the format: {"label": 0} or {"label": 1}.
No explanations or extra text.
\end{lstlisting}

To enhance the reliability of our results, we select the optimal hyperparameter configuration and repeat training twice using different random seeds.
We report the mean performance and standard deviations across these runs.

\subsection{PLMs}

\begin{table}[H]
    \centering
    \begin{adjustbox}{width=.8\linewidth}
        \begin{tabular}{l l}
        \hline
        \textbf{Hyperparameter} & \textbf{Value(s)} \\
        \hline
        Epochs & {1,2,3} \\
        Learning rate & $\{5\times10^{-5},\,1\times10^{-5},\,5\times10^{-6}\}$ \\
        Batch size & $\{16,\,32\}$ \\
        Max gradient norm & 1.0 \\
        Weight decay & 0.01 \\
        Seed & 42 \\
        \hline
        \end{tabular}

    \end{adjustbox}
    
\caption{Hyperparameter grid configuration for PLMs.}
\label{tab:hps_plm}
\end{table}

For PLMs, we fully fine-tune each model-language combination with hyperparameter grid search using the hyperparameters displayed in Table~\ref{tab:hps_plm}.
We run reliability checks identical to the SLM setup.

\subsection{LLMs}

For LLMs, we use the same \textsc{instruct} prompt applied to SLMs as well as a more verbose prompt, as \citet{majer2024} show that prompt verbosity can influence performance.
The verbose prompt provides additional task instructions and incorporates the citation reason taxonomy proposed by \citet{redi2019}:

\begin{lstlisting}[language={},basicstyle=\ttfamily\small]
Verbose Prompt
----------------
You are a multilingual Wikipedia citation classifier.
You are provided with a {lang} claim and its context.
Your task is to analyze the claim and the context to decide whether the claim needs a citation.
On Wikipedia, there are various reasons why a claim may or may not require a citation.
The reasons are listed below:

Reasons why citations are needed (Label 1):
Quotation - The statement is a direct quotation or close paraphrase of a source.
Statistics - The statement contains statistics or quantitative data.
Controversial - The statement makes surprising or potentially controversial claims.
Opinion - The statement expresses a person's subjective opinion or belief.
Private Life - The statement contains claims about a person's private life.
Scientific - The statement includes technical or scientific claims.
Historical - The statement makes general or historical claims that are not common knowledge.
Other (Needs Citation) - The statement requires a citation for other reasons.

Reasons why citations are not needed (Label 0):
Common Knowledge - The statement contains well-known or widely established facts.
Plot - The statement describes the plot or characters of a work that is the subject of the article.
Other (No Citation Needed) - The statement does not require a citation for other reasons.

Based on these reasons, think step-by-step to decide in which category the claim falls.
Return only JSON in the format: {"label": 0} or {"label": 1}.
No extra text.
\end{lstlisting}

We use OpenRouter\footnote{\url{https://openrouter.ai/}} to run inference with the selected LLMs.
All LLM experiments are conducted with a low temperature (0.1) which is why we do not perform additional reliability runs.

For few-shot prompting, we randomly sample two examples per label.
As shown in Table~\ref{tab:all_acc}, the inclusion of random shots often improves performance, particularly for the \textsc{instruct} prompt variant.

\section{Results}
\label{app:results}

In this section, we report the results of our main experiments using F1 score instead of accuracy.
We further present additional analyses on the effect of different learning rates for few-shot tuning of SLMs.

\subsection{Experiment 1: SFT Variants}

Table~\ref{tab:atl_f1} presents the results of our SFT variant experiments using F1 score as the main evaluation metric.

\begin{table*}[t]
    \centering
        \begin{adjustbox}{max width=.9\linewidth}\centering

    \begin{tabular}{llCCCCCCCCCCCCCCCCCCCCC}
 & & \multicolumn{9}{c}{\textbf{High Resource}} & \multicolumn{6}{c}{\textbf{Medium Resource}} & \multicolumn{6}{c}{\textbf{Low Resource}} \\\cmidrule(lr){3-11}  \cmidrule(lr){12-17}  \cmidrule(lr){18-23}\textbf{Model} & \textbf{SFT} & \textbf{en} & \textbf{pt} & \textbf{de} & \textbf{ru} & \textbf{it} & \textbf{vi} & \textbf{tr} & \textbf{nl}& \textbf{Avg} & \textbf{uk} & \textbf{ro} & \textbf{id} & \textbf{bg} & \textbf{uz}& \textbf{Avg} & \textbf{no} & \textbf{az} & \textbf{mk} & \textbf{hy} & \textbf{sq}& \textbf{Avg} \\
\toprule
\multirow{3}{*}[-1.8ex]{Aya-8b} &  FTL & 85.39  \scriptsize{$(\pm1.05)$} & 69.63  \scriptsize{$(\pm3.94)$} & 61.29  \scriptsize{$(\pm1.64)$} & 70.89  \scriptsize{$(\pm3.56)$} & 67.20  \scriptsize{$(\pm2.00)$} & 73.53  \scriptsize{$(\pm1.56)$} & 76.08  \scriptsize{$(\pm0.31)$} & \textbf{66.76} \scriptsize{$(\pm 2.10)$} & \cellcolor{gray!25} 71.35 \scriptsize{$(\pm7.24)$} & 66.72  \scriptsize{$(\pm4.84)$} & 70.84  \scriptsize{$(\pm2.10)$} & 73.13  \scriptsize{$(\pm1.52)$} & \textbf{67.20} \scriptsize{$(\pm 0.20)$} & 66.79  \scriptsize{$(\pm0.21)$} & \cellcolor{gray!25} 68.94 \scriptsize{$(\pm2.91)$} & 64.02  \scriptsize{$(\pm3.14)$} & 67.72  \scriptsize{$(\pm0.97)$} & 61.53  \scriptsize{$(\pm8.31)$} & 54.45  \scriptsize{$(\pm3.94)$} & 69.02  \scriptsize{$(\pm1.69)$} & \cellcolor{gray!25} 63.35 \scriptsize{$(\pm5.79)$} \\
 & TOL & 84.03  \scriptsize{$(\pm1.59)$} & 74.48  \scriptsize{$(\pm0.36)$} & \textbf{66.39} \scriptsize{$(\pm 0.99)$} & 76.03  \scriptsize{$(\pm0.81)$} & 70.78  \scriptsize{$(\pm0.96)$} & 79.17  \scriptsize{$(\pm3.09)$} & \textbf{78.50} \scriptsize{$(\pm 1.47)$} & 66.76  \scriptsize{$(\pm0.00)$} & \cellcolor{gray!25} 74.52 \scriptsize{$(\pm6.22)$} & 60.05  \scriptsize{$(\pm7.09)$} & 78.27  \scriptsize{$(\pm0.52)$} & 68.41  \scriptsize{$(\pm3.17)$} & 66.90  \scriptsize{$(\pm0.14)$} & 66.67  \scriptsize{$(\pm0.00)$} & \cellcolor{gray!25} 68.06 \scriptsize{$(\pm6.56)$} & \textbf{67.03} \scriptsize{$(\pm 0.37)$} & 68.69  \scriptsize{$(\pm2.92)$} & \textbf{66.55} \scriptsize{$(\pm 0.26)$} & 62.35  \scriptsize{$(\pm5.32)$} & 66.12  \scriptsize{$(\pm0.51)$} & \cellcolor{gray!25} 66.15 \scriptsize{$(\pm2.34)$} \\
 & ES & \textbf{89.42} \scriptsize{$(\pm 0.42)$} & \textbf{81.45} \scriptsize{$(\pm 0.34)$} & 65.57  \scriptsize{$(\pm2.89)$} & \textbf{77.45} \scriptsize{$(\pm 0.42)$} & \textbf{75.45} \scriptsize{$(\pm 3.12)$} & \textbf{82.17} \scriptsize{$(\pm 0.72)$} & 76.83  \scriptsize{$(\pm0.27)$} & 62.88  \scriptsize{$(\pm2.11)$} & \cellcolor{gray!25} \textbf{76.40} \scriptsize{$(\pm8.71)$} & \textbf{70.39} \scriptsize{$(\pm 1.69)$} & \textbf{79.49} \scriptsize{$(\pm 0.71)$} & \textbf{75.41} \scriptsize{$(\pm 0.59)$} & 64.86  \scriptsize{$(\pm1.53)$} & \textbf{69.30} \scriptsize{$(\pm 1.19)$} & \cellcolor{gray!25} \textbf{71.89} \scriptsize{$(\pm5.67)$} & 64.67  \scriptsize{$(\pm0.60)$} & \textbf{71.63} \scriptsize{$(\pm 1.17)$} & 65.03  \scriptsize{$(\pm2.41)$} & \textbf{63.12} \scriptsize{$(\pm 2.57)$} & \textbf{73.80} \scriptsize{$(\pm 1.12)$} & \cellcolor{gray!25} \textbf{67.65} \scriptsize{$(\pm4.74)$} \\
\midrule
\multirow{3}{*}[-1.8ex]{Llama3-8B} &  FTL & 88.93  \scriptsize{$(\pm1.05)$} & 79.79  \scriptsize{$(\pm1.05)$} & 62.58  \scriptsize{$(\pm4.81)$} & \textbf{78.85} \scriptsize{$(\pm 0.71)$} & 75.20  \scriptsize{$(\pm0.25)$} & 81.10  \scriptsize{$(\pm1.61)$} & 76.38  \scriptsize{$(\pm0.21)$} & \textbf{62.60} \scriptsize{$(\pm 1.01)$} & \cellcolor{gray!25} 75.68 \scriptsize{$(\pm9.06)$} & 69.68  \scriptsize{$(\pm2.27)$} & \textbf{76.20} \scriptsize{$(\pm 2.72)$} & 76.75  \scriptsize{$(\pm0.83)$} & 66.52  \scriptsize{$(\pm1.65)$} & \textbf{74.08} \scriptsize{$(\pm 2.67)$} & \cellcolor{gray!25} \textbf{72.65} \scriptsize{$(\pm4.41)$} & 62.40  \scriptsize{$(\pm4.23)$} & 69.29  \scriptsize{$(\pm2.77)$} & 63.24  \scriptsize{$(\pm1.10)$} & 50.31  \scriptsize{$(\pm2.78)$} & 71.83  \scriptsize{$(\pm0.48)$} & \cellcolor{gray!25} 63.41 \scriptsize{$(\pm8.34)$} \\
 & TOL & 89.42  \scriptsize{$(\pm1.85)$} & 79.45  \scriptsize{$(\pm0.99)$} & 64.85  \scriptsize{$(\pm5.89)$} & 77.97  \scriptsize{$(\pm0.73)$} & \textbf{75.39} \scriptsize{$(\pm 0.64)$} & \textbf{82.53} \scriptsize{$(\pm 1.13)$} & \textbf{77.44} \scriptsize{$(\pm 0.21)$} & 61.76  \scriptsize{$(\pm1.39)$} & \cellcolor{gray!25} \textbf{76.10} \scriptsize{$(\pm9.01)$} & 70.28  \scriptsize{$(\pm1.28)$} & 75.03  \scriptsize{$(\pm1.43)$} & 77.47  \scriptsize{$(\pm0.78)$} & 65.76  \scriptsize{$(\pm0.60)$} & 69.67  \scriptsize{$(\pm2.27)$} & \cellcolor{gray!25} 71.64 \scriptsize{$(\pm4.63)$} & 64.34  \scriptsize{$(\pm2.77)$} & 69.06  \scriptsize{$(\pm1.85)$} & 64.38  \scriptsize{$(\pm1.41)$} & 49.80  \scriptsize{$(\pm11.73)$} & 71.00  \scriptsize{$(\pm2.73)$} & \cellcolor{gray!25} 63.72 \scriptsize{$(\pm8.31)$} \\
 & ES & \textbf{89.87} \scriptsize{$(\pm 0.29)$} & \textbf{80.75} \scriptsize{$(\pm 0.88)$} & \textbf{66.00} \scriptsize{$(\pm 2.97)$} & 78.34  \scriptsize{$(\pm1.56)$} & 73.32  \scriptsize{$(\pm2.01)$} & 82.23  \scriptsize{$(\pm0.86)$} & 76.61  \scriptsize{$(\pm1.01)$} & 58.06  \scriptsize{$(\pm4.14)$} & \cellcolor{gray!25} 75.65 \scriptsize{$(\pm9.91)$} & \textbf{72.38} \scriptsize{$(\pm 1.08)$} & 69.50  \scriptsize{$(\pm8.30)$} & \textbf{77.85} \scriptsize{$(\pm 1.48)$} & \textbf{66.76} \scriptsize{$(\pm 0.00)$} & 69.61  \scriptsize{$(\pm2.01)$} & \cellcolor{gray!25} 71.22 \scriptsize{$(\pm4.21)$} & \textbf{66.37} \scriptsize{$(\pm 0.63)$} & \textbf{72.12} \scriptsize{$(\pm 2.85)$} & \textbf{66.58} \scriptsize{$(\pm 1.51)$} & \textbf{65.07} \scriptsize{$(\pm 1.83)$} & \textbf{73.76} \scriptsize{$(\pm 2.00)$} & \cellcolor{gray!25} \textbf{68.78} \scriptsize{$(\pm3.89)$} \\
\midrule
\multirow{3}{*}[-1.8ex]{Qwen3-8B} &  FTL & 88.26  \scriptsize{$(\pm2.57)$} & 79.92  \scriptsize{$(\pm0.97)$} & 58.39  \scriptsize{$(\pm6.59)$} & 75.58  \scriptsize{$(\pm0.86)$} & 72.98  \scriptsize{$(\pm2.18)$} & 81.69  \scriptsize{$(\pm0.93)$} & 77.09  \scriptsize{$(\pm0.48)$} & 60.86  \scriptsize{$(\pm0.76)$} & \cellcolor{gray!25} 74.35 \scriptsize{$(\pm10.18)$} & 62.25  \scriptsize{$(\pm2.67)$} & 76.71  \scriptsize{$(\pm1.30)$} & 75.01  \scriptsize{$(\pm0.85)$} & \textbf{65.77} \scriptsize{$(\pm 0.89)$} & 70.12  \scriptsize{$(\pm0.43)$} & \cellcolor{gray!25} 69.97 \scriptsize{$(\pm6.08)$} & 61.69  \scriptsize{$(\pm1.56)$} & 71.15  \scriptsize{$(\pm2.10)$} & 59.22  \scriptsize{$(\pm4.02)$} & 59.58  \scriptsize{$(\pm1.19)$} & 70.87  \scriptsize{$(\pm1.32)$} & \cellcolor{gray!25} 64.50 \scriptsize{$(\pm6.02)$} \\
 & TOL & 89.24  \scriptsize{$(\pm1.49)$} & \textbf{80.10} \scriptsize{$(\pm 0.94)$} & 62.97  \scriptsize{$(\pm5.27)$} & \textbf{78.02} \scriptsize{$(\pm 0.83)$} & 73.16  \scriptsize{$(\pm1.75)$} & 79.72  \scriptsize{$(\pm1.06)$} & \textbf{77.96} \scriptsize{$(\pm 0.09)$} & 61.37  \scriptsize{$(\pm2.53)$} & \cellcolor{gray!25} 75.32 \scriptsize{$(\pm9.27)$} & 68.54  \scriptsize{$(\pm1.75)$} & 75.71  \scriptsize{$(\pm2.09)$} & 74.25  \scriptsize{$(\pm0.57)$} & 63.18  \scriptsize{$(\pm1.50)$} & \textbf{77.20} \scriptsize{$(\pm 0.45)$} & \cellcolor{gray!25} 71.78 \scriptsize{$(\pm5.82)$} & 63.61  \scriptsize{$(\pm2.18)$} & 68.12  \scriptsize{$(\pm3.70)$} & 59.95  \scriptsize{$(\pm2.34)$} & 62.95  \scriptsize{$(\pm1.45)$} & \textbf{73.66} \scriptsize{$(\pm 1.38)$} & \cellcolor{gray!25} 65.66 \scriptsize{$(\pm5.35)$} \\
 & ES & \textbf{89.25} \scriptsize{$(\pm 0.69)$} & 79.55  \scriptsize{$(\pm0.05)$} & \textbf{66.75} \scriptsize{$(\pm 4.01)$} & 77.75  \scriptsize{$(\pm0.76)$} & \textbf{75.02} \scriptsize{$(\pm 2.00)$} & \textbf{82.82} \scriptsize{$(\pm 0.08)$} & 77.17  \scriptsize{$(\pm0.91)$} & \textbf{63.12} \scriptsize{$(\pm 2.93)$} & \cellcolor{gray!25} \textbf{76.43} \scriptsize{$(\pm8.37)$} & \textbf{70.87} \scriptsize{$(\pm 2.69)$} & \textbf{78.03} \scriptsize{$(\pm 1.56)$} & \textbf{75.80} \scriptsize{$(\pm 2.69)$} & 64.71  \scriptsize{$(\pm1.83)$} & 70.03  \scriptsize{$(\pm2.00)$} & \cellcolor{gray!25} \textbf{71.89} \scriptsize{$(\pm5.22)$} & \textbf{68.26} \scriptsize{$(\pm 2.58)$} & \textbf{72.74} \scriptsize{$(\pm 1.63)$} & \textbf{63.29} \scriptsize{$(\pm 1.34)$} & \textbf{67.02} \scriptsize{$(\pm 1.95)$} & 70.48  \scriptsize{$(\pm1.98)$} & \cellcolor{gray!25} \textbf{68.36} \scriptsize{$(\pm3.58)$} \\
\bottomrule
\end{tabular}
    
    \end{adjustbox}
        
    \caption{Monolingual CND F1 scores of SLMs across SFT variants (FTL=full-token loss, TOL=target-only loss, ES=encoder-style). 
    \textbf{Bold} denotes the best score per model–language pair. Parentheses indicate standard deviations.}
    \label{tab:atl_f1}
\end{table*}

\subsection{Experiment 2: Multilingual CND Benchmark}

Table~\ref{tab:all_f1} presents the multilingual CND benchmarking results for all LM families, using F1 score as the main evaluation metric.

\begin{table*}[t]
\centering
    \begin{adjustbox}{max width=.9\linewidth}\centering

    \begin{tabular}{lCCCCCCCCCCCCCCCCCCCCCC}
 & \multicolumn{9}{c}{\textbf{High Resource}} & \multicolumn{6}{c}{\textbf{Medium Resource}} & \multicolumn{6}{c}{\textbf{Low Resource}} \\\cmidrule(lr){2-10}  \cmidrule(lr){11-16}  \cmidrule(lr){17-22} & \textbf{en} & \textbf{pt} & \textbf{de} & \textbf{ru} & \textbf{it} & \textbf{vi} & \textbf{tr} & \textbf{nl} & \textbf{Avg} & \textbf{uk} & \textbf{ro} & \textbf{id} & \textbf{bg} & \textbf{uz} & \textbf{Avg} & \textbf{no} & \textbf{az} & \textbf{mk} & \textbf{hy} & \textbf{sq} & \textbf{Avg} & \textbf{Avg} \\
\toprule
\multicolumn{23}{c}{\textbf{Decoder-based LLMs}} \\
\midrule
\textit{Zero-shot} \\
\midrule
GPT-5 (Instruct) & 79.61 & 70.06 & \underline{68.07} & 73.56 & 70.99 & 73.11 & 75.63 & 66.30 & \cellcolor{gray!25} 72.17 \scriptsize{($\pm4.26$)} & 67.41 & 73.76 & 67.51 & 67.50 & 68.78 & \cellcolor{gray!25} 68.99 \scriptsize{($\pm2.73$)} & 66.67 & 69.54 & \textbf{68.93} & 66.03 & 65.49 & \cellcolor{gray!25} 67.33 \scriptsize{($\pm1.80$)} & 69.94 \scriptsize{($\pm3.82$)} \\
GPT-5 (Verbose) & 83.36 & 69.33 & 66.57 & 72.11 & 68.54 & 74.39 & 76.06 & \underline{66.30} & \cellcolor{gray!25} 72.08 \scriptsize{($\pm5.76$)} & 67.13 & 73.54 & 68.92 & 67.68 & 68.59 & \cellcolor{gray!25} 69.17 \scriptsize{($\pm2.55$)} & 67.49 & 71.13 & 67.50 & 66.76 & 66.30 & \cellcolor{gray!25} 67.84 \scriptsize{($\pm1.91$)} & 70.09 \scriptsize{($\pm4.43$)} \\
Gemini 2.5 (Instruct) & 77.96 & 69.73 & 65.51 & 72.55 & 71.08 & 72.29 & 70.62 & 63.14 & \cellcolor{gray!25} 70.36 \scriptsize{($\pm4.52$)} & 66.47 & 72.23 & 66.14 & 68.89 & 69.71 & \cellcolor{gray!25} 68.69 \scriptsize{($\pm2.50$)} & 65.15 & 66.37 & 67.06 & 65.65 & 65.20 & \cellcolor{gray!25} 65.89 \scriptsize{($\pm0.82$)} & 68.65 \scriptsize{($\pm3.70$)} \\
Gemini 2.5 (Verbose) & 81.09 & 70.85 & 66.17 & 71.04 & 66.67 & 74.17 & 74.43 & 64.81 & \cellcolor{gray!25} 71.15 \scriptsize{($\pm5.40$)} & 67.62 & 71.52 & 68.48 & 68.75 & 68.18 & \cellcolor{gray!25} 68.91 \scriptsize{($\pm1.52$)} & 66.28 & 68.29 & 66.95 & 65.29 & 64.75 & \cellcolor{gray!25} 66.31 \scriptsize{($\pm1.40$)} & 69.19 \scriptsize{($\pm4.15$)} \\
\midrule
\textit{Few-shot} \\
\midrule
GPT-5 (Instruct) & 84.04 & 70.14 & 67.42 & 74.44 & 74.43 & 75.82 & 77.02 & \textbf{66.67} & \cellcolor{gray!25} 73.75 \scriptsize{($\pm5.67$)} & 66.28 & 74.51 & 69.14 & \underline{69.35} & \underline{73.68} & \cellcolor{gray!25} 70.59 \scriptsize{($\pm3.43$)} & 66.67 & 70.52 & \underline{67.73} & 66.67 & 65.54 & \cellcolor{gray!25} 67.42 \scriptsize{($\pm1.89$)} & 71.12 \scriptsize{($\pm4.92$)} \\
GPT-5 (Verbose) & 83.04 & 69.84 & 65.83 & 71.49 & 69.15 & 75.62 & 75.74 & \textbf{66.67} & \cellcolor{gray!25} 72.18 \scriptsize{($\pm5.71$)} & 67.41 & 73.81 & 69.70 & 67.60 & 69.10 & \cellcolor{gray!25} 69.53 \scriptsize{($\pm2.59$)} & 67.04 & 70.96 & 67.23 & 66.94 & 65.45 & \cellcolor{gray!25} 67.52 \scriptsize{($\pm2.05$)} & 70.15 \scriptsize{($\pm4.48$)} \\
Gemini 2.5 (Instruct) & 80.87 & 69.74 & \textbf{68.64} & 69.55 & 69.58 & 73.54 & 66.38 & 62.81 & \cellcolor{gray!25} 70.14 \scriptsize{($\pm5.31$)} & 64.42 & 47.35 & 68.57 & \textbf{71.14} & 69.96 & \cellcolor{gray!25} 64.29 \scriptsize{($\pm9.80$)} & 61.72 & 63.44 & 65.27 & 58.59 & 66.54 & \cellcolor{gray!25} 63.11 \scriptsize{($\pm3.12$)} & 66.56 \scriptsize{($\pm6.90$)} \\
Gemini 2.5 (Verbose) & 81.90 & 71.67 & 64.47 & 71.34 & 66.67 & 75.30 & 72.35 & 62.88 & \cellcolor{gray!25} 70.82 \scriptsize{($\pm6.18$)} & 64.08 & 67.13 & 70.89 & 68.18 & 70.26 & \cellcolor{gray!25} 68.11 \scriptsize{($\pm2.72$)} & 65.78 & 68.11 & 66.67 & 66.08 & 63.92 & \cellcolor{gray!25} 66.11 \scriptsize{($\pm1.52$)} & 68.76 \scriptsize{($\pm4.71$)} \\
\midrule
\multicolumn{23}{c}{\textbf{Encoder-based PLMs}} \\
\midrule
XLM-R-base & 88.38 \scriptsize{($\pm0.26$)} & 77.47 \scriptsize{($\pm1.08$)} & 62.67 \scriptsize{($\pm3.12$)} & 77.04 \scriptsize{($\pm1.44$)} & 70.56 \scriptsize{($\pm3.64$)} & 76.47 \scriptsize{($\pm0.61$)} & \underline{77.03} \scriptsize{($\pm0.90$)} & 64.51 \scriptsize{($\pm3.49$)} & \cellcolor{gray!25} 74.27 \scriptsize{($\pm8.22$)} & 67.38 \scriptsize{($\pm3.14$)} & 72.46 \scriptsize{($\pm4.72$)} & 69.52 \scriptsize{($\pm3.35$)} & 64.45 \scriptsize{($\pm0.76$)} & 66.67 \scriptsize{($\pm0.00$)} & \cellcolor{gray!25} 68.10 \scriptsize{($\pm3.04$)} & 66.41 \scriptsize{($\pm0.78$)} & 66.61 \scriptsize{($\pm1.07$)} & 64.90 \scriptsize{($\pm0.88$)} & \textbf{67.76} \scriptsize{($\pm0.74$)} & 64.64 \scriptsize{($\pm2.92$)} & \cellcolor{gray!25} 66.06 \scriptsize{($\pm1.29$)} & 70.32 \scriptsize{($\pm6.87$)} \\
XLM-R-large & \textbf{90.06} \scriptsize{($\pm0.22$)} & 79.34 \scriptsize{($\pm1.80$)} & 52.23 \scriptsize{($\pm23.10$)} & 77.42 \scriptsize{($\pm1.53$)} & 74.59 \scriptsize{($\pm1.16$)} & 79.86 \scriptsize{($\pm0.51$)} & 75.80 \scriptsize{($\pm1.65$)} & 44.61 \scriptsize{($\pm38.63$)} & \cellcolor{gray!25} 71.74 \scriptsize{($\pm15.27$)} & 67.09 \scriptsize{($\pm0.84$)} & \textbf{79.76} \scriptsize{($\pm0.68$)} & 71.18 \scriptsize{($\pm3.65$)} & 60.45 \scriptsize{($\pm7.15$)} & 66.67 \scriptsize{($\pm0.00$)} & \cellcolor{gray!25} 69.03 \scriptsize{($\pm7.12$)} & \textbf{70.28} \scriptsize{($\pm1.61$)} & 69.16 \scriptsize{($\pm0.98$)} & 65.79 \scriptsize{($\pm0.72$)} & 64.08 \scriptsize{($\pm4.93$)} & 66.54 \scriptsize{($\pm1.30$)} & \cellcolor{gray!25} 67.17 \scriptsize{($\pm2.52$)} & 69.69 \scriptsize{($\pm13.89$)} \\
mBert & 89.26 \scriptsize{($\pm0.58$)} & \underline{80.87} \scriptsize{($\pm0.70$)} & 62.13 \scriptsize{($\pm1.66$)} & 75.91 \scriptsize{($\pm0.53$)} & 73.52 \scriptsize{($\pm1.63$)} & 79.07 \scriptsize{($\pm1.21$)} & 74.70 \scriptsize{($\pm0.93$)} & 63.04 \scriptsize{($\pm2.67$)} & \cellcolor{gray!25} 74.81 \scriptsize{($\pm8.99$)} & 69.72 \scriptsize{($\pm2.82$)} & 79.11 \scriptsize{($\pm1.37$)} & \underline{75.89} \scriptsize{($\pm0.59$)} & 63.92 \scriptsize{($\pm3.01$)} & \textbf{75.49} \scriptsize{($\pm7.88$)} & \cellcolor{gray!25} \textbf{72.83} \scriptsize{($\pm6.02$)} & 67.83 \scriptsize{($\pm2.31$)} & \underline{72.13} \scriptsize{($\pm0.58$)} & 64.89 \scriptsize{($\pm1.22$)} & 66.30 \scriptsize{($\pm2.19$)} & \textbf{73.85} \scriptsize{($\pm0.93$)} & \cellcolor{gray!25} \textbf{69.00} \scriptsize{($\pm3.84$)} & 72.66 \scriptsize{($\pm7.39$)} \\
\midrule
\multicolumn{23}{c}{\textbf{Decoder-based SLMs}} \\
\midrule
Aya-8b (ES) & 89.42 \scriptsize{($\pm0.42$)} & \textbf{81.45} \scriptsize{($\pm0.34$)} & 65.57 \scriptsize{($\pm2.89$)} & 77.45 \scriptsize{($\pm0.42$)} & \textbf{75.45} \scriptsize{($\pm3.12$)} & 82.17 \scriptsize{($\pm0.72$)} & 76.83 \scriptsize{($\pm0.27$)} & 62.88 \scriptsize{($\pm2.11$)} & \cellcolor{gray!25} \underline{76.40} \scriptsize{($\pm8.71$)} & 70.39 \scriptsize{($\pm1.69$)} & \underline{79.49} \scriptsize{($\pm0.71$)} & 75.41 \scriptsize{($\pm0.59$)} & 64.86 \scriptsize{($\pm1.53$)} & 69.30 \scriptsize{($\pm1.19$)} & \cellcolor{gray!25} 71.89 \scriptsize{($\pm5.67$)} & 64.67 \scriptsize{($\pm0.60$)} & 71.63 \scriptsize{($\pm1.17$)} & 65.03 \scriptsize{($\pm2.41$)} & 63.12 \scriptsize{($\pm2.57$)} & \underline{73.80} \scriptsize{($\pm1.12$)} & \cellcolor{gray!25} 67.65 \scriptsize{($\pm4.74$)} & \underline{72.72} \scriptsize{($\pm7.60$)} \\
Llama3-8B (ES) & \underline{89.87} \scriptsize{($\pm0.29$)} & 80.75 \scriptsize{($\pm0.88$)} & 66.00 \scriptsize{($\pm2.97$)} & \textbf{78.34} \scriptsize{($\pm1.56$)} & 73.32 \scriptsize{($\pm2.01$)} & \underline{82.23} \scriptsize{($\pm0.86$)} & 76.61 \scriptsize{($\pm1.01$)} & 58.06 \scriptsize{($\pm4.14$)} & \cellcolor{gray!25} 75.65 \scriptsize{($\pm9.91$)} & \textbf{72.38} \scriptsize{($\pm1.08$)} & 69.50 \scriptsize{($\pm8.30$)} & \textbf{77.85} \scriptsize{($\pm1.48$)} & 66.76 \scriptsize{($\pm0.00$)} & 69.61 \scriptsize{($\pm2.01$)} & \cellcolor{gray!25} 71.22 \scriptsize{($\pm4.21$)} & 66.37 \scriptsize{($\pm0.63$)} & 72.12 \scriptsize{($\pm2.85$)} & 66.58 \scriptsize{($\pm1.51$)} & 65.07 \scriptsize{($\pm1.83$)} & 73.76 \scriptsize{($\pm2.00$)} & \cellcolor{gray!25} \underline{68.78} \scriptsize{($\pm3.89$)} & 72.62 \scriptsize{($\pm7.78$)} \\
Qwen3-8B (ES) & 89.25 \scriptsize{($\pm0.69$)} & 79.55 \scriptsize{($\pm0.05$)} & 66.75 \scriptsize{($\pm4.01$)} & \underline{77.75} \scriptsize{($\pm0.76$)} & \underline{75.02} \scriptsize{($\pm2.00$)} & \textbf{82.82} \scriptsize{($\pm0.08$)} & \textbf{77.17} \scriptsize{($\pm0.91$)} & 63.12 \scriptsize{($\pm2.93$)} & \cellcolor{gray!25} \textbf{76.43} \scriptsize{($\pm8.37$)} & \underline{70.87} \scriptsize{($\pm2.69$)} & 78.03 \scriptsize{($\pm1.56$)} & 75.80 \scriptsize{($\pm2.69$)} & 64.71 \scriptsize{($\pm1.83$)} & 70.03 \scriptsize{($\pm2.00$)} & \cellcolor{gray!25} 71.89 \scriptsize{($\pm5.22$)} & \underline{68.26} \scriptsize{($\pm2.58$)} & \textbf{72.74} \scriptsize{($\pm1.63$)} & 63.29 \scriptsize{($\pm1.34$)} & \underline{67.02} \scriptsize{($\pm1.95$)} & 70.48 \scriptsize{($\pm1.98$)} & \cellcolor{gray!25} 68.36 \scriptsize{($\pm3.58$)} & \textbf{72.93} \scriptsize{($\pm7.17$)} \\
\bottomrule
\end{tabular}
    
    \end{adjustbox}
        
    \caption{Monolingual CND F1 scores across LM families (ES=encoder-style; Verbose and Instruct denote prompt types). 
    \textbf{Bold} and \underline{underlined} scores mark the best and second-best results, respectively. Parentheses indicate standard deviation.}
    \label{tab:all_f1}
\end{table*}

\subsubsection*{Error Analysis of the German Wikipedia}

We observe comparatively low evaluation scores for the German Wikipedia.
To better understand this effect, we manually analyze model misclassifications.
We find that real-world citation practices of Wikipedia editors occasionally diverge from the labeling heuristic we adopt, which is also used in prior work~\cite{redi2019,halitaj2024}.
We illustrate these discrepancies with two representative examples.

\paragraph{Example 1}
The first deviation arises from citations that appear only at the end of multiple consecutive paragraphs.
In the German Wikipedia article on the city of "Augsburg, the section "Gewässer" consists of six paragraphs.
While the first five paragraphs contain no citations, the final paragraph includes a single citation that appears to support the entire multi-paragraph.
We observe several such cases, suggesting that editors sometimes choose to include only one citation to cover a broader textual span.

Under the established labeling heuristic, claims in all but the final paragraph would be incorrectly labeled as not requiring a citation.
One potential remedy would be to treat all claims preceding the cited paragraph as citation-worthy.
However, this assumption does not always hold, as the final citation does not necessarily reference all preceding paragraphs in every multi-paragraph setting.

\paragraph{Example 2}
A second issue arises when a citation attached to an earlier claim implicitly supports subsequent claims without being repeated, provided that the connection is obvious.
For example, in the article \emph{Useless}, section \emph{Video and Vinyl}:

\begin{quote}
2004 veröffentlichte das Bochumer Label Dirty Faces Records das Album \emph{Useless} als Schallplatte und zusätzlich auf farbig gesprenkeltem Vinyl in limitierter Auflage von 525 Stück.\,[18] Eine weitere Ausgabe des Albums in pinkfarbenem Vinyl erschien im Jahr 2017 bei Drumming Monkey Records.

(\textit{English translation}) In 2004, the Bochum-based label Dirty Faces Records released the album \emph{Useless} on vinyl and additionally on colored splatter vinyl in a limited edition of 525 copies.\,[18] Another edition of the album on pink vinyl was released in 2017 by Drumming Monkey Records.
\end{quote}

In this context, the citation~[18] clearly also supports the final sentence.
To account for this editorial practice, we modify existing labeling heuristics~\cite{redi2019,halitaj2024} to also label the final sentence as citation needed, which is correct in this case.
However, this assumption does not hold for all sentences without explicit citations.

Real-world editorial citation practices are difficult to capture automatically, as they vary across Wikipedia language communities.
Overall, these examples highlight the limitations of current heuristic labeling approaches for CND on Wikipedia.
Designing more robust labeling strategies therefore remains an important direction for future research.

\subsection{Experiment 3: X-CND}

\subsubsection*{Zero-shot X-CND}

\begin{figure}[H]
    \centering
        \includegraphics[width=.9\linewidth]{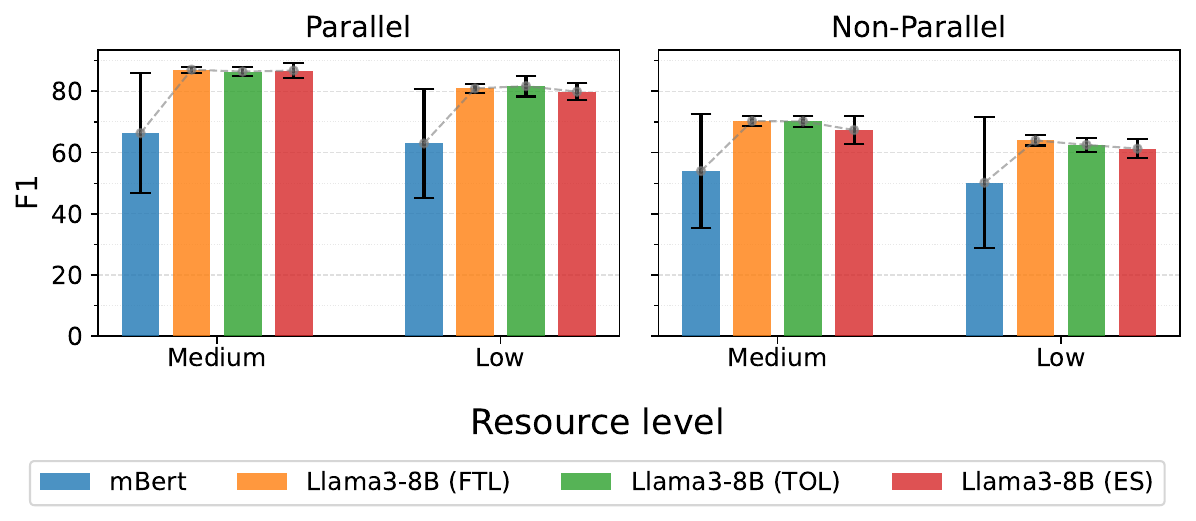}
    \caption{Zero-shot CND F1 scores by test data setting, averaged across language resource groups.}
    \label{fig:zero_shot_f1}
\end{figure}

Figure~\ref{fig:zero_shot_f1} presents zero-shot X-CND results with F1 score as the main metric.

\subsubsection*{Few-shot performance by learning rates}
\begin{table}[t]
\centering
    \begin{adjustbox}{max width=.9\linewidth}\centering

    \begin{tabular}{llcccccc}
 & & \multicolumn{5}{c}{\textbf{Shots}} \\ 
\cmidrule(lr){3-7} 
\textbf{Model}   & \textbf{LR} & $k=50$ & $k=100$ & $k=250$ & $k=500$ & \textbf{Avg} $\Delta$ \\ \toprule
\textbf{Medium Resource }  &  &  &  &  &  &  \\
\midrule
\multirow{4}{*}{Llama 3-8B (FTL)} & 1e-04 & \makecell{66.57 \\[-3pt] \small{($\pm${4.64})}}  \small{(\posneg{-0.50})} & \makecell{66.24 \\[-3pt] \small{($\pm${4.94})}}  \small{(\posneg{-0.83})} & \makecell{65.19 \\[-3pt] \small{($\pm${5.98})}}  \small{(\posneg{-1.88})} & \makecell{66.78 \\[-3pt] \small{($\pm${5.35})}}  \small{(\posneg{-0.29})} & -0.88 \\ 
 & 1e-05 & \makecell{67.20 \\[-3pt] \small{($\pm${5.19})}}  \small{(\posneg{0.13})} & \makecell{67.60 \\[-3pt] \small{($\pm${5.39})}}  \small{(\posneg{0.53})} & \makecell{67.53 \\[-3pt] \small{($\pm${5.73})}}  \small{(\posneg{0.45})} & \makecell{67.78 \\[-3pt] \small{($\pm${5.39})}}  \small{(\posneg{0.71})} & \textbf{0.46} \\ 
 & 1e-06 & \makecell{66.91 \\[-3pt] \small{($\pm${5.33})}}  \small{(\posneg{-0.16})} & \makecell{67.58 \\[-3pt] \small{($\pm${5.27})}}  \small{(\posneg{0.51})} & \makecell{67.38 \\[-3pt] \small{($\pm${5.18})}}  \small{(\posneg{0.31})} & \makecell{68.02 \\[-3pt] \small{($\pm${5.59})}}  \small{(\posneg{0.95})} & 0.40 \\ 
 & 1e-07 & \makecell{66.91 \\[-3pt] \small{($\pm${5.42})}}  \small{(\posneg{-0.16})} & \makecell{67.47 \\[-3pt] \small{($\pm${5.21})}}  \small{(\posneg{0.40})} & \makecell{67.23 \\[-3pt] \small{($\pm${5.27})}}  \small{(\posneg{0.16})} & \makecell{68.07 \\[-3pt] \small{($\pm${5.41})}}  \small{(\posneg{1.00})} & 0.35 \\ 
\cmidrule(lr){2-7}
\multirow{4}{*}{Llama 3-8B (TOL)} & 1e-04 & \makecell{60.90 \\[-3pt] \small{($\pm${4.81})}}  \small{(\posneg{-5.55})} & \makecell{60.52 \\[-3pt] \small{($\pm${7.95})}}  \small{(\posneg{-5.93})} & \makecell{66.08 \\[-3pt] \small{($\pm${9.82})}}  \small{(\posneg{-0.36})} & \makecell{70.03 \\[-3pt] \small{($\pm${8.31})}}  \small{(\posneg{3.59})} & -2.06 \\ 
 & 1e-05 & \makecell{65.83 \\[-3pt] \small{($\pm${6.13})}}  \small{(\posneg{-0.62})} & \makecell{66.81 \\[-3pt] \small{($\pm${5.99})}}  \small{(\posneg{0.36})} & \makecell{65.14 \\[-3pt] \small{($\pm${5.63})}}  \small{(\posneg{-1.31})} & \makecell{65.95 \\[-3pt] \small{($\pm${5.47})}}  \small{(\posneg{-0.50})} & -0.52 \\ 
 & 1e-06 & \makecell{66.32 \\[-3pt] \small{($\pm${5.92})}}  \small{(\posneg{-0.13})} & \makecell{66.82 \\[-3pt] \small{($\pm${5.99})}}  \small{(\posneg{0.38})} & \makecell{66.28 \\[-3pt] \small{($\pm${5.30})}}  \small{(\posneg{-0.16})} & \makecell{67.26 \\[-3pt] \small{($\pm${5.54})}}  \small{(\posneg{0.82})} & 0.23 \\ 
 & 1e-07 & \makecell{66.59 \\[-3pt] \small{($\pm${5.84})}}  \small{(\posneg{0.15})} & \makecell{67.03 \\[-3pt] \small{($\pm${5.89})}}  \small{(\posneg{0.59})} & \makecell{66.64 \\[-3pt] \small{($\pm${5.49})}}  \small{(\posneg{0.19})} & \makecell{67.10 \\[-3pt] \small{($\pm${6.35})}}  \small{(\posneg{0.65})} & \textbf{0.39} \\ 
\cmidrule(lr){2-7}
\multirow{4}{*}{Llama 3-8B (ES)} & 1e-04 & \makecell{65.55 \\[-3pt] \small{($\pm${6.58})}}  \small{(\posneg{-2.64})} & \makecell{64.07 \\[-3pt] \small{($\pm${5.90})}}  \small{(\posneg{-4.12})} & \makecell{69.13 \\[-3pt] \small{($\pm${6.00})}}  \small{(\posneg{0.94})} & \makecell{72.38 \\[-3pt] \small{($\pm${5.73})}}  \small{(\posneg{4.19})} & -0.41 \\ 
 & 1e-05 & \makecell{69.36 \\[-3pt] \small{($\pm${4.45})}}  \small{(\posneg{1.18})} & \makecell{69.80 \\[-3pt] \small{($\pm${4.44})}}  \small{(\posneg{1.61})} & \makecell{70.24 \\[-3pt] \small{($\pm${4.63})}}  \small{(\posneg{2.06})} & \makecell{71.14 \\[-3pt] \small{($\pm${4.08})}}  \small{(\posneg{2.95})} & \textbf{1.95} \\ 
 & 1e-06 & \makecell{68.39 \\[-3pt] \small{($\pm${4.56})}}  \small{(\posneg{0.20})} & \makecell{68.67 \\[-3pt] \small{($\pm${4.54})}}  \small{(\posneg{0.48})} & \makecell{69.23 \\[-3pt] \small{($\pm${4.46})}}  \small{(\posneg{1.04})} & \makecell{69.84 \\[-3pt] \small{($\pm${4.26})}}  \small{(\posneg{1.65})} & 0.84 \\ 
 & 1e-07 & \makecell{68.19 \\[-3pt] \small{($\pm${4.88})}}  \small{(\posneg{-0.00})} & \makecell{68.25 \\[-3pt] \small{($\pm${4.84})}}  \small{(\posneg{0.07})} & \makecell{68.33 \\[-3pt] \small{($\pm${4.88})}}  \small{(\posneg{0.15})} & \makecell{68.43 \\[-3pt] \small{($\pm${4.63})}}  \small{(\posneg{0.24})} & 0.11 \\ 
\cmidrule(lr){2-7}
\multirow{5}{*}{mBert} & 1e-04 & \makecell{55.39 \\[-3pt] \small{($\pm${7.49})}}  \small{(\posneg{-5.09})} & \makecell{59.83 \\[-3pt] \small{($\pm${9.79})}}  \small{(\posneg{-0.66})} & \makecell{59.62 \\[-3pt] \small{($\pm${7.34})}}  \small{(\posneg{-0.86})} & \makecell{58.11 \\[-3pt] \small{($\pm${7.21})}}  \small{(\posneg{-2.37})} & -2.25 \\ 
 & 5e-05 & \makecell{62.18 \\[-3pt] \small{($\pm${7.42})}}  \small{(\posneg{1.70})} & \makecell{63.33 \\[-3pt] \small{($\pm${7.32})}}  \small{(\posneg{2.85})} & \makecell{64.77 \\[-3pt] \small{($\pm${7.50})}}  \small{(\posneg{4.28})} & \makecell{63.74 \\[-3pt] \small{($\pm${6.89})}}  \small{(\posneg{3.26})} & \textbf{3.03} \\ 
 & 1e-05 & \makecell{60.96 \\[-3pt] \small{($\pm${5.93})}}  \small{(\posneg{0.48})} & \makecell{61.31 \\[-3pt] \small{($\pm${6.02})}}  \small{(\posneg{0.83})} & \makecell{61.52 \\[-3pt] \small{($\pm${5.40})}}  \small{(\posneg{1.04})} & \makecell{62.63 \\[-3pt] \small{($\pm${5.75})}}  \small{(\posneg{2.15})} & 1.13 \\ 
 & 1e-06 & \makecell{60.45 \\[-3pt] \small{($\pm${6.17})}}  \small{(\posneg{-0.03})} & \makecell{60.52 \\[-3pt] \small{($\pm${6.17})}}  \small{(\posneg{0.04})} & \makecell{60.48 \\[-3pt] \small{($\pm${6.14})}}  \small{(\posneg{-0.00})} & \makecell{60.44 \\[-3pt] \small{($\pm${6.11})}}  \small{(\posneg{-0.04})} & -0.01 \\ 
 & 1e-07 & \makecell{60.49 \\[-3pt] \small{($\pm${6.11})}}  \small{(\posneg{0.01})} & \makecell{60.51 \\[-3pt] \small{($\pm${6.14})}}  \small{(\posneg{0.03})} & \makecell{60.47 \\[-3pt] \small{($\pm${6.21})}}  \small{(\posneg{-0.01})} & \makecell{60.49 \\[-3pt] \small{($\pm${6.14})}}  \small{(\posneg{0.01})} & 0.01 \\ 
\midrule
\textbf{Low Resource }  &  &  &  &  &  &  \\
\midrule
\multirow{4}{*}{Llama 3-8B (FTL)} & 1e-04 & \makecell{62.61 \\[-3pt] \small{($\pm${4.74})}}  \small{(\posneg{0.02})} & \makecell{61.88 \\[-3pt] \small{($\pm${5.27})}}  \small{(\posneg{-0.71})} & \makecell{61.42 \\[-3pt] \small{($\pm${5.24})}}  \small{(\posneg{-1.17})} & \makecell{62.71 \\[-3pt] \small{($\pm${6.01})}}  \small{(\posneg{0.12})} & -0.44 \\ 
 & 1e-05 & \makecell{62.92 \\[-3pt] \small{($\pm${4.17})}}  \small{(\posneg{0.32})} & \makecell{62.35 \\[-3pt] \small{($\pm${3.96})}}  \small{(\posneg{-0.24})} & \makecell{62.60 \\[-3pt] \small{($\pm${5.12})}}  \small{(\posneg{0.01})} & \makecell{63.72 \\[-3pt] \small{($\pm${4.97})}}  \small{(\posneg{1.13})} & 0.31 \\ 
 & 1e-06 & \makecell{62.90 \\[-3pt] \small{($\pm${4.12})}}  \small{(\posneg{0.30})} & \makecell{62.47 \\[-3pt] \small{($\pm${3.30})}}  \small{(\posneg{-0.12})} & \makecell{62.44 \\[-3pt] \small{($\pm${4.35})}}  \small{(\posneg{-0.15})} & \makecell{63.65 \\[-3pt] \small{($\pm${4.39})}}  \small{(\posneg{1.06})} & 0.27 \\ 
 & 1e-07 & \makecell{62.82 \\[-3pt] \small{($\pm${3.93})}}  \small{(\posneg{0.23})} & \makecell{62.50 \\[-3pt] \small{($\pm${3.43})}}  \small{(\posneg{-0.10})} & \makecell{62.38 \\[-3pt] \small{($\pm${4.28})}}  \small{(\posneg{-0.21})} & \makecell{63.97 \\[-3pt] \small{($\pm${4.58})}}  \small{(\posneg{1.38})} & \textbf{0.32} \\ 
\cmidrule(lr){2-7}
\multirow{4}{*}{Llama 3-8B (TOL)} & 1e-04 & \makecell{57.76 \\[-3pt] \small{($\pm${4.73})}}  \small{(\posneg{-2.89})} & \makecell{58.43 \\[-3pt] \small{($\pm${5.02})}}  \small{(\posneg{-2.22})} & \makecell{60.74 \\[-3pt] \small{($\pm${6.08})}}  \small{(\posneg{0.08})} & \makecell{63.22 \\[-3pt] \small{($\pm${5.57})}}  \small{(\posneg{2.57})} & -0.61 \\ 
 & 1e-05 & \makecell{60.96 \\[-3pt] \small{($\pm${5.24})}}  \small{(\posneg{0.31})} & \makecell{60.48 \\[-3pt] \small{($\pm${5.13})}}  \small{(\posneg{-0.18})} & \makecell{60.10 \\[-3pt] \small{($\pm${4.72})}}  \small{(\posneg{-0.55})} & \makecell{61.31 \\[-3pt] \small{($\pm${4.50})}}  \small{(\posneg{0.65})} & 0.06 \\ 
 & 1e-06 & \makecell{60.58 \\[-3pt] \small{($\pm${4.97})}}  \small{(\posneg{-0.08})} & \makecell{60.88 \\[-3pt] \small{($\pm${4.45})}}  \small{(\posneg{0.23})} & \makecell{61.16 \\[-3pt] \small{($\pm${3.91})}}  \small{(\posneg{0.50})} & \makecell{61.69 \\[-3pt] \small{($\pm${3.89})}}  \small{(\posneg{1.04})} & \textbf{0.42} \\ 
 & 1e-07 & \makecell{61.19 \\[-3pt] \small{($\pm${3.90})}}  \small{(\posneg{0.53})} & \makecell{60.66 \\[-3pt] \small{($\pm${6.07})}}  \small{(\posneg{0.00})} & \makecell{60.53 \\[-3pt] \small{($\pm${4.82})}}  \small{(\posneg{-0.13})} & \makecell{61.49 \\[-3pt] \small{($\pm${4.69})}}  \small{(\posneg{0.84})} & 0.31 \\ 
\cmidrule(lr){2-7}
\multirow{4}{*}{Llama 3-8B (ES)} & 1e-04 & \makecell{59.41 \\[-3pt] \small{($\pm${4.29})}}  \small{(\posneg{-2.88})} & \makecell{58.25 \\[-3pt] \small{($\pm${5.23})}}  \small{(\posneg{-4.05})} & \makecell{62.44 \\[-3pt] \small{($\pm${6.33})}}  \small{(\posneg{0.14})} & \makecell{64.40 \\[-3pt] \small{($\pm${5.21})}}  \small{(\posneg{2.11})} & -1.17 \\ 
 & 1e-05 & \makecell{63.31 \\[-3pt] \small{($\pm${4.21})}}  \small{(\posneg{1.01})} & \makecell{62.98 \\[-3pt] \small{($\pm${4.66})}}  \small{(\posneg{0.68})} & \makecell{63.47 \\[-3pt] \small{($\pm${5.05})}}  \small{(\posneg{1.17})} & \makecell{64.43 \\[-3pt] \small{($\pm${5.20})}}  \small{(\posneg{2.13})} & \textbf{1.25} \\ 
 & 1e-06 & \makecell{62.68 \\[-3pt] \small{($\pm${5.59})}}  \small{(\posneg{0.39})} & \makecell{62.79 \\[-3pt] \small{($\pm${5.61})}}  \small{(\posneg{0.49})} & \makecell{62.96 \\[-3pt] \small{($\pm${5.56})}}  \small{(\posneg{0.67})} & \makecell{63.85 \\[-3pt] \small{($\pm${4.93})}}  \small{(\posneg{1.55})} & 0.77 \\ 
 & 1e-07 & \makecell{62.30 \\[-3pt] \small{($\pm${5.64})}}  \small{(\posneg{0.00})} & \makecell{62.34 \\[-3pt] \small{($\pm${5.62})}}  \small{(\posneg{0.04})} & \makecell{62.56 \\[-3pt] \small{($\pm${5.52})}}  \small{(\posneg{0.27})} & \makecell{62.52 \\[-3pt] \small{($\pm${5.51})}}  \small{(\posneg{0.23})} & 0.13 \\ 
\cmidrule(lr){2-7}
\multirow{5}{*}{mBert} & 1e-04 & \makecell{53.92 \\[-3pt] \small{($\pm${6.44})}}  \small{(\posneg{-2.99})} & \makecell{55.84 \\[-3pt] \small{($\pm${7.25})}}  \small{(\posneg{-1.07})} & \makecell{57.31 \\[-3pt] \small{($\pm${7.45})}}  \small{(\posneg{0.40})} & \makecell{58.08 \\[-3pt] \small{($\pm${8.44})}}  \small{(\posneg{1.17})} & -0.62 \\ 
 & 5e-05 & \makecell{58.97 \\[-3pt] \small{($\pm${6.77})}}  \small{(\posneg{2.06})} & \makecell{60.27 \\[-3pt] \small{($\pm${6.87})}}  \small{(\posneg{3.35})} & \makecell{59.76 \\[-3pt] \small{($\pm${4.93})}}  \small{(\posneg{2.84})} & \makecell{62.16 \\[-3pt] \small{($\pm${6.50})}}  \small{(\posneg{5.24})} & \textbf{3.37} \\ 
 & 1e-05 & \makecell{57.09 \\[-3pt] \small{($\pm${4.42})}}  \small{(\posneg{0.17})} & \makecell{57.37 \\[-3pt] \small{($\pm${4.33})}}  \small{(\posneg{0.45})} & \makecell{57.49 \\[-3pt] \small{($\pm${4.10})}}  \small{(\posneg{0.58})} & \makecell{58.04 \\[-3pt] \small{($\pm${4.23})}}  \small{(\posneg{1.12})} & 0.58 \\ 
 & 1e-06 & \makecell{56.93 \\[-3pt] \small{($\pm${4.47})}}  \small{(\posneg{0.01})} & \makecell{56.90 \\[-3pt] \small{($\pm${4.46})}}  \small{(\posneg{-0.01})} & \makecell{56.93 \\[-3pt] \small{($\pm${4.42})}}  \small{(\posneg{0.01})} & \makecell{56.93 \\[-3pt] \small{($\pm${4.42})}}  \small{(\posneg{0.01})} & 0.01 \\ 
 & 1e-07 & \makecell{56.87 \\[-3pt] \small{($\pm${4.45})}}  \small{(\posneg{-0.04})} & \makecell{56.90 \\[-3pt] \small{($\pm${4.47})}}  \small{(\posneg{-0.01})} & \makecell{56.83 \\[-3pt] \small{($\pm${4.43})}}  \small{(\posneg{-0.08})} & \makecell{56.91 \\[-3pt] \small{($\pm${4.47})}}  \small{(\posneg{0.00})} & -0.03 \\ 
\bottomrule
\end{tabular}
    
    \end{adjustbox}
        
    \caption{Few-shot accuracy for X-CND. LR=learning rate. Parentheses indicate standard deviations. We observe that generative SFT yields consistent improvements only when using substantially reduced learning rates, whereas mBERT and ES-based tuning remain effective under higher learning rates.}
    \label{tab:lr_acc}
\end{table}

\begin{table}[t]
\centering
    \begin{adjustbox}{max width=.9\linewidth}\centering

    \begin{tabular}{llcccccc}
 & & \multicolumn{5}{c}{\textbf{Shots}} \\ 
\cmidrule(lr){3-7} 
\textbf{Model}   & \textbf{LR} & $k=50$ & $k=100$ & $k=250$ & $k=500$ & \textbf{Avg} $\Delta$ \\ \toprule
\textbf{Medium Resource }  &  &  &  &  &  &  \\
\midrule
\multirow{4}{*}{Llama 3-8B (FTL)} & 1e-04 & \makecell{67.41 \\[-3pt] \small{($\pm${3.41})}}  \small{(\posneg{-2.85})} & \makecell{64.82 \\[-3pt] \small{($\pm${4.34})}}  \small{(\posneg{-5.45})} & \makecell{65.07 \\[-3pt] \small{($\pm${6.59})}}  \small{(\posneg{-5.20})} & \makecell{66.67 \\[-3pt] \small{($\pm${5.47})}}  \small{(\posneg{-3.60})} & -4.27 \\ 
 & 1e-05 & \makecell{70.19 \\[-3pt] \small{($\pm${3.98})}}  \small{(\posneg{-0.07})} & \makecell{70.35 \\[-3pt] \small{($\pm${4.14})}}  \small{(\posneg{0.09})} & \makecell{69.61 \\[-3pt] \small{($\pm${4.66})}}  \small{(\posneg{-0.66})} & \makecell{69.06 \\[-3pt] \small{($\pm${4.57})}}  \small{(\posneg{-1.20})} & -0.46 \\ 
 & 1e-06 & \makecell{70.16 \\[-3pt] \small{($\pm${4.07})}}  \small{(\posneg{-0.11})} & \makecell{70.70 \\[-3pt] \small{($\pm${3.87})}}  \small{(\posneg{0.43})} & \makecell{70.34 \\[-3pt] \small{($\pm${4.14})}}  \small{(\posneg{0.08})} & \makecell{70.93 \\[-3pt] \small{($\pm${4.22})}}  \small{(\posneg{0.66})} & 0.27 \\ 
 & 1e-07 & \makecell{70.16 \\[-3pt] \small{($\pm${4.15})}}  \small{(\posneg{-0.10})} & \makecell{70.65 \\[-3pt] \small{($\pm${3.81})}}  \small{(\posneg{0.38})} & \makecell{70.31 \\[-3pt] \small{($\pm${4.17})}}  \small{(\posneg{0.05})} & \makecell{71.18 \\[-3pt] \small{($\pm${3.95})}}  \small{(\posneg{0.92})} & \textbf{0.31} \\ 
\cmidrule(lr){2-7}
\multirow{4}{*}{Llama 3-8B (TOL)} & 1e-04 & \makecell{63.68 \\[-3pt] \small{($\pm${6.51})}}  \small{(\posneg{-6.49})} & \makecell{60.20 \\[-3pt] \small{($\pm${7.88})}}  \small{(\posneg{-9.97})} & \makecell{67.72 \\[-3pt] \small{($\pm${8.67})}}  \small{(\posneg{-2.46})} & \makecell{71.60 \\[-3pt] \small{($\pm${7.58})}}  \small{(\posneg{1.42})} & -4.37 \\ 
 & 1e-05 & \makecell{67.71 \\[-3pt] \small{($\pm${4.51})}}  \small{(\posneg{-2.47})} & \makecell{67.81 \\[-3pt] \small{($\pm${4.36})}}  \small{(\posneg{-2.36})} & \makecell{65.49 \\[-3pt] \small{($\pm${5.13})}}  \small{(\posneg{-4.69})} & \makecell{66.40 \\[-3pt] \small{($\pm${5.47})}}  \small{(\posneg{-3.78})} & -3.32 \\ 
 & 1e-06 & \makecell{69.95 \\[-3pt] \small{($\pm${4.20})}}  \small{(\posneg{-0.22})} & \makecell{70.31 \\[-3pt] \small{($\pm${4.43})}}  \small{(\posneg{0.13})} & \makecell{69.45 \\[-3pt] \small{($\pm${3.92})}}  \small{(\posneg{-0.72})} & \makecell{69.36 \\[-3pt] \small{($\pm${4.18})}}  \small{(\posneg{-0.82})} & -0.41 \\ 
 & 1e-07 & \makecell{69.77 \\[-3pt] \small{($\pm${4.65})}}  \small{(\posneg{-0.40})} & \makecell{70.13 \\[-3pt] \small{($\pm${4.82})}}  \small{(\posneg{-0.05})} & \makecell{69.97 \\[-3pt] \small{($\pm${4.21})}}  \small{(\posneg{-0.20})} & \makecell{70.28 \\[-3pt] \small{($\pm${4.61})}}  \small{(\posneg{0.10})} & \textbf{-0.14} \\ 
\cmidrule(lr){2-7}
\multirow{4}{*}{Llama 3-8B (ES)} & 1e-04 & \makecell{69.06 \\[-3pt] \small{($\pm${4.44})}}  \small{(\posneg{1.72})} & \makecell{63.22 \\[-3pt] \small{($\pm${11.34})}}  \small{(\posneg{-4.12})} & \makecell{69.40 \\[-3pt] \small{($\pm${6.72})}}  \small{(\posneg{2.06})} & \makecell{73.46 \\[-3pt] \small{($\pm${5.80})}}  \small{(\posneg{6.11})} & 1.44 \\ 
 & 1e-05 & \makecell{70.67 \\[-3pt] \small{($\pm${3.82})}}  \small{(\posneg{3.32})} & \makecell{70.49 \\[-3pt] \small{($\pm${4.32})}}  \small{(\posneg{3.15})} & \makecell{71.03 \\[-3pt] \small{($\pm${4.99})}}  \small{(\posneg{3.69})} & \makecell{71.90 \\[-3pt] \small{($\pm${4.48})}}  \small{(\posneg{4.55})} & \textbf{3.68} \\ 
 & 1e-06 & \makecell{67.84 \\[-3pt] \small{($\pm${7.85})}}  \small{(\posneg{0.50})} & \makecell{68.56 \\[-3pt] \small{($\pm${6.69})}}  \small{(\posneg{1.22})} & \makecell{69.40 \\[-3pt] \small{($\pm${5.53})}}  \small{(\posneg{2.06})} & \makecell{70.11 \\[-3pt] \small{($\pm${4.40})}}  \small{(\posneg{2.77})} & 1.64 \\ 
 & 1e-07 & \makecell{67.36 \\[-3pt] \small{($\pm${8.87})}}  \small{(\posneg{0.02})} & \makecell{67.48 \\[-3pt] \small{($\pm${8.74})}}  \small{(\posneg{0.14})} & \makecell{67.60 \\[-3pt] \small{($\pm${8.56})}}  \small{(\posneg{0.26})} & \makecell{67.76 \\[-3pt] \small{($\pm${8.14})}}  \small{(\posneg{0.42})} & 0.21 \\ 
\cmidrule(lr){2-7}
\multirow{5}{*}{mBert} & 1e-04 & \makecell{51.78 \\[-3pt] \small{($\pm${28.71})}}  \small{(\posneg{-2.19})} & \makecell{59.87 \\[-3pt] \small{($\pm${21.01})}}  \small{(\posneg{5.90})} & \makecell{56.43 \\[-3pt] \small{($\pm${21.12})}}  \small{(\posneg{2.47})} & \makecell{48.66 \\[-3pt] \small{($\pm${28.22})}}  \small{(\posneg{-5.30})} & 0.22 \\ 
 & 5e-05 & \makecell{63.05 \\[-3pt] \small{($\pm${16.79})}}  \small{(\posneg{9.09})} & \makecell{66.79 \\[-3pt] \small{($\pm${9.08})}}  \small{(\posneg{12.82})} & \makecell{69.34 \\[-3pt] \small{($\pm${5.88})}}  \small{(\posneg{15.38})} & \makecell{65.41 \\[-3pt] \small{($\pm${12.03})}}  \small{(\posneg{11.45})} & \textbf{12.19} \\ 
 & 1e-05 & \makecell{55.84 \\[-3pt] \small{($\pm${15.93})}}  \small{(\posneg{1.87})} & \makecell{56.60 \\[-3pt] \small{($\pm${15.21})}}  \small{(\posneg{2.63})} & \makecell{57.17 \\[-3pt] \small{($\pm${13.88})}}  \small{(\posneg{3.21})} & \makecell{59.28 \\[-3pt] \small{($\pm${12.47})}}  \small{(\posneg{5.32})} & 3.26 \\ 
 & 1e-06 & \makecell{53.94 \\[-3pt] \small{($\pm${17.70})}}  \small{(\posneg{-0.03})} & \makecell{53.99 \\[-3pt] \small{($\pm${17.72})}}  \small{(\posneg{0.03})} & \makecell{53.99 \\[-3pt] \small{($\pm${17.65})}}  \small{(\posneg{0.03})} & \makecell{53.89 \\[-3pt] \small{($\pm${17.77})}}  \small{(\posneg{-0.08})} & -0.01 \\ 
 & 1e-07 & \makecell{53.93 \\[-3pt] \small{($\pm${17.76})}}  \small{(\posneg{-0.03})} & \makecell{53.95 \\[-3pt] \small{($\pm${17.77})}}  \small{(\posneg{-0.01})} & \makecell{53.94 \\[-3pt] \small{($\pm${17.78})}}  \small{(\posneg{-0.03})} & \makecell{53.99 \\[-3pt] \small{($\pm${17.74})}}  \small{(\posneg{0.02})} & -0.01 \\ 
\midrule
\textbf{Low Resource }  &  &  &  &  &  &  \\
\midrule
\multirow{4}{*}{Llama 3-8B (FTL)} & 1e-04 & \makecell{61.21 \\[-3pt] \small{($\pm${5.43})}}  \small{(\posneg{-2.74})} & \makecell{58.89 \\[-3pt] \small{($\pm${6.74})}}  \small{(\posneg{-5.06})} & \makecell{57.50 \\[-3pt] \small{($\pm${6.23})}}  \small{(\posneg{-6.44})} & \makecell{58.82 \\[-3pt] \small{($\pm${6.57})}}  \small{(\posneg{-5.12})} & -4.84 \\ 
 & 1e-05 & \makecell{64.22 \\[-3pt] \small{($\pm${4.49})}}  \small{(\posneg{0.28})} & \makecell{63.15 \\[-3pt] \small{($\pm${4.50})}}  \small{(\posneg{-0.79})} & \makecell{62.45 \\[-3pt] \small{($\pm${4.48})}}  \small{(\posneg{-1.49})} & \makecell{62.57 \\[-3pt] \small{($\pm${4.73})}}  \small{(\posneg{-1.38})} & -0.85 \\ 
 & 1e-06 & \makecell{64.50 \\[-3pt] \small{($\pm${4.80})}}  \small{(\posneg{0.56})} & \makecell{63.88 \\[-3pt] \small{($\pm${4.35})}}  \small{(\posneg{-0.06})} & \makecell{63.40 \\[-3pt] \small{($\pm${5.46})}}  \small{(\posneg{-0.55})} & \makecell{65.09 \\[-3pt] \small{($\pm${3.45})}}  \small{(\posneg{1.15})} & 0.28 \\ 
 & 1e-07 & \makecell{64.49 \\[-3pt] \small{($\pm${4.83})}}  \small{(\posneg{0.54})} & \makecell{63.96 \\[-3pt] \small{($\pm${4.43})}}  \small{(\posneg{0.02})} & \makecell{63.47 \\[-3pt] \small{($\pm${5.34})}}  \small{(\posneg{-0.47})} & \makecell{65.64 \\[-3pt] \small{($\pm${3.12})}}  \small{(\posneg{1.69})} & \textbf{0.45} \\ 
\cmidrule(lr){2-7}
\multirow{4}{*}{Llama 3-8B (TOL)} & 1e-04 & \makecell{56.74 \\[-3pt] \small{($\pm${7.51})}}  \small{(\posneg{-5.78})} & \makecell{55.48 \\[-3pt] \small{($\pm${7.30})}}  \small{(\posneg{-7.05})} & \makecell{54.39 \\[-3pt] \small{($\pm${13.34})}}  \small{(\posneg{-8.14})} & \makecell{59.38 \\[-3pt] \small{($\pm${9.49})}}  \small{(\posneg{-3.14})} & -6.02 \\ 
 & 1e-05 & \makecell{59.54 \\[-3pt] \small{($\pm${7.81})}}  \small{(\posneg{-2.98})} & \makecell{56.44 \\[-3pt] \small{($\pm${7.51})}}  \small{(\posneg{-6.08})} & \makecell{54.72 \\[-3pt] \small{($\pm${6.46})}}  \small{(\posneg{-7.80})} & \makecell{56.80 \\[-3pt] \small{($\pm${7.39})}}  \small{(\posneg{-5.72})} & -5.65 \\ 
 & 1e-06 & \makecell{62.89 \\[-3pt] \small{($\pm${7.09})}}  \small{(\posneg{0.37})} & \makecell{62.58 \\[-3pt] \small{($\pm${6.44})}}  \small{(\posneg{0.06})} & \makecell{61.70 \\[-3pt] \small{($\pm${6.15})}}  \small{(\posneg{-0.82})} & \makecell{61.25 \\[-3pt] \small{($\pm${5.62})}}  \small{(\posneg{-1.27})} & -0.41 \\ 
 & 1e-07 & \makecell{62.40 \\[-3pt] \small{($\pm${8.03})}}  \small{(\posneg{-0.12})} & \makecell{61.82 \\[-3pt] \small{($\pm${8.67})}}  \small{(\posneg{-0.70})} & \makecell{61.52 \\[-3pt] \small{($\pm${7.05})}}  \small{(\posneg{-1.00})} & \makecell{62.85 \\[-3pt] \small{($\pm${5.33})}}  \small{(\posneg{0.33})} & \textbf{-0.37} \\ 
\cmidrule(lr){2-7}
\multirow{4}{*}{Llama 3-8B (ES)} & 1e-04 & \makecell{61.01 \\[-3pt] \small{($\pm${8.95})}}  \small{(\posneg{-0.26})} & \makecell{51.65 \\[-3pt] \small{($\pm${16.46})}}  \small{(\posneg{-9.62})} & \makecell{61.91 \\[-3pt] \small{($\pm${6.45})}}  \small{(\posneg{0.64})} & \makecell{63.98 \\[-3pt] \small{($\pm${6.84})}}  \small{(\posneg{2.71})} & -1.63 \\ 
 & 1e-05 & \makecell{58.39 \\[-3pt] \small{($\pm${6.97})}}  \small{(\posneg{-2.88})} & \makecell{61.48 \\[-3pt] \small{($\pm${5.37})}}  \small{(\posneg{0.21})} & \makecell{62.42 \\[-3pt] \small{($\pm${4.94})}}  \small{(\posneg{1.15})} & \makecell{64.29 \\[-3pt] \small{($\pm${4.16})}}  \small{(\posneg{3.02})} & 0.37 \\ 
 & 1e-06 & \makecell{61.28 \\[-3pt] \small{($\pm${10.65})}}  \small{(\posneg{0.01})} & \makecell{61.80 \\[-3pt] \small{($\pm${9.36})}}  \small{(\posneg{0.53})} & \makecell{61.57 \\[-3pt] \small{($\pm${7.68})}}  \small{(\posneg{0.30})} & \makecell{62.42 \\[-3pt] \small{($\pm${4.67})}}  \small{(\posneg{1.15})} & \textbf{0.50} \\ 
 & 1e-07 & \makecell{61.14 \\[-3pt] \small{($\pm${11.91})}}  \small{(\posneg{-0.13})} & \makecell{61.31 \\[-3pt] \small{($\pm${11.56})}}  \small{(\posneg{0.04})} & \makecell{61.54 \\[-3pt] \small{($\pm${11.12})}}  \small{(\posneg{0.27})} & \makecell{61.50 \\[-3pt] \small{($\pm${10.80})}}  \small{(\posneg{0.22})} & 0.10 \\ 
\cmidrule(lr){2-7}
\multirow{5}{*}{mBert} & 1e-04 & \makecell{46.34 \\[-3pt] \small{($\pm${30.89})}}  \small{(\posneg{-3.79})} & \makecell{44.50 \\[-3pt] \small{($\pm${29.78})}}  \small{(\posneg{-5.64})} & \makecell{55.81 \\[-3pt] \small{($\pm${18.73})}}  \small{(\posneg{5.67})} & \makecell{55.14 \\[-3pt] \small{($\pm${22.42})}}  \small{(\posneg{5.00})} & 0.31 \\ 
 & 5e-05 & \makecell{57.78 \\[-3pt] \small{($\pm${17.48})}}  \small{(\posneg{7.64})} & \makecell{60.39 \\[-3pt] \small{($\pm${10.91})}}  \small{(\posneg{10.25})} & \makecell{62.30 \\[-3pt] \small{($\pm${7.50})}}  \small{(\posneg{12.16})} & \makecell{65.13 \\[-3pt] \small{($\pm${6.03})}}  \small{(\posneg{14.99})} & \textbf{11.26} \\ 
 & 1e-05 & \makecell{51.42 \\[-3pt] \small{($\pm${17.46})}}  \small{(\posneg{1.28})} & \makecell{52.17 \\[-3pt] \small{($\pm${15.91})}}  \small{(\posneg{2.03})} & \makecell{53.17 \\[-3pt] \small{($\pm${14.55})}}  \small{(\posneg{3.03})} & \makecell{55.03 \\[-3pt] \small{($\pm${12.63})}}  \small{(\posneg{4.89})} & 2.81 \\ 
 & 1e-06 & \makecell{50.15 \\[-3pt] \small{($\pm${18.85})}}  \small{(\posneg{0.01})} & \makecell{50.15 \\[-3pt] \small{($\pm${18.80})}}  \small{(\posneg{0.01})} & \makecell{50.20 \\[-3pt] \small{($\pm${18.76})}}  \small{(\posneg{0.06})} & \makecell{50.21 \\[-3pt] \small{($\pm${18.74})}}  \small{(\posneg{0.07})} & 0.04 \\ 
 & 1e-07 & \makecell{50.10 \\[-3pt] \small{($\pm${18.83})}}  \small{(\posneg{-0.04})} & \makecell{50.10 \\[-3pt] \small{($\pm${18.89})}}  \small{(\posneg{-0.04})} & \makecell{49.99 \\[-3pt] \small{($\pm${18.91})}}  \small{(\posneg{-0.15})} & \makecell{50.14 \\[-3pt] \small{($\pm${18.80})}}  \small{(\posneg{0.00})} & -0.06 \\ 
\bottomrule
\end{tabular}
    
    \end{adjustbox}
        
    \caption{Few-shot F1 scores for X-CND. LR=learning rate. Parentheses indicate standard deviations.}
    \label{tab:lr_f1}
\end{table}

For SLMs fine-tuned with QLoRA, the optimal learning rate is typically relatively higher (2e-4) compared to most other fine-tuning tasks~\cite{dettmers2023}.
This is also the case for CND, where we observe that across nearly all model-language combinations, hyperparameter search consistently identifies high learning rates as optimal.
However, when applying continued fine-tuning for few-shot transfer following~\cite{lauscher2020}, we quickly observe that this strategy is detrimental to the performance of SLMs.

Therefore, for the second-stage fine-tuning on the target language shots, we treat the learning rate as the sole hyperparameter to optimize.
While we could also include other hyperparameters, the connection between high learning rates, small amounts of training shots, and poor model performance seems most obvious to remedy.

Tables~\ref{tab:lr_acc} and~\ref{tab:lr_f1} report the results of our experiments analyzing the effect of learning rate choice on few-shot X-CND.
We observe a consistent pattern: Llama (ES) and mBERT benefit from relatively higher learning rates, whereas generative SFT variants (FTL and TOL) exhibit improvements from few-shot tuning only when using substantially lower learning rates, down to 1e-7.
Even then, the gains achieved by generative SFT through additional few-shot fine-tuning remain modest.
Overall, these results indicate that learning rate selection is critical for continued fine-tuning on small amounts of target-language data for QLoRA-based SLMs, particularly under generative SFT.
In contrast, mBERT performs optimally with learning rates similar to those identified during hyperparameter.

\subsubsection*{Few-shot Performance}

\begin{table*}[t]
\centering
    \begin{adjustbox}{max width=.9\linewidth}\centering

\begin{tabular}{lccccccccccccc}
 & & \multicolumn{12}{c}{\textbf{Non-Parallel Data}} \\ 
\cmidrule(lr){3-14} 
 & & \multicolumn{6}{c}{\textbf{Medium-Resource}} & \multicolumn{6}{c}{\textbf{Low-Resource}} \\ 
\cmidrule(lr){3-8} \cmidrule(lr){9-14} 
\textbf{Model} & \textbf{Shots} & \textbf{uk} & \textbf{ro} & \textbf{id} & \textbf{bg} & \textbf{uz} & $\Delta$ \textbf{Avg} & \textbf{no} & \textbf{az} & \textbf{mk} & \textbf{hy} & \textbf{sq} & $\Delta$ \textbf{Avg}  \\ 
\toprule
\multirow{5}{*}{Llama3-8B (FTL)} & \cellcolor{gray!25} 0 &  65.47 \scriptsize{($\pm$1.25)}&  75.55 \scriptsize{($\pm$1.87)}&  71.91 \scriptsize{($\pm$0.53)}&  65.87 \scriptsize{($\pm$2.76)}&  72.52 \scriptsize{($\pm$1.92)} &  &  65.03 \scriptsize{($\pm$0.39)}&  67.03 \scriptsize{($\pm$2.54)}&  63.11 \scriptsize{($\pm$0.39)}&  56.39 \scriptsize{($\pm$3.19)}&  68.16 \scriptsize{($\pm$1.73)} &  \\
 & $\Delta$ \; 50&  \posneg{0.27} \scriptsize{($\pm$2.18)}&  \posneg{0.21} \scriptsize{($\pm$1.76)}&  \posneg{-0.47} \scriptsize{($\pm$1.23)}&  \posneg{0.25} \scriptsize{($\pm$0.74)}&  \posneg{-0.76} \scriptsize{($\pm$1.80)} &  \posneg{-0.10} &  \posneg{-0.44} \scriptsize{($\pm$1.33)}&  \posneg{0.14} \scriptsize{($\pm$1.21)}&  \posneg{2.04} \scriptsize{($\pm$2.26)}&  \posneg{0.71} \scriptsize{($\pm$5.59)}&  \posneg{0.27} \scriptsize{($\pm$2.70)} &  \posneg{0.54} \\
 & $\Delta$ 100&  \posneg{-0.15} \scriptsize{($\pm$1.04)}&  \posneg{-0.84} \scriptsize{($\pm$0.43)}&  \posneg{0.56} \scriptsize{($\pm$0.54)}&  \posneg{1.61} \scriptsize{($\pm$1.17)}&  \posneg{0.74} \scriptsize{($\pm$1.12)} &  \posneg{0.38} &  \posneg{0.69} \scriptsize{($\pm$1.54)}&  \posneg{-0.93} \scriptsize{($\pm$1.23)}&  \posneg{0.88} \scriptsize{($\pm$2.50)}&  \posneg{0.21} \scriptsize{($\pm$3.83)}&  \posneg{-0.76} \scriptsize{($\pm$1.34)} &  \posneg{0.02} \\
 & $\Delta$ 250&  \posneg{-0.62} \scriptsize{($\pm$1.24)}&  \posneg{0.05} \scriptsize{($\pm$1.40)}&  \posneg{1.08} \scriptsize{($\pm$0.77)}&  \posneg{1.35} \scriptsize{($\pm$1.07)}&  \posneg{-1.64} \scriptsize{($\pm$2.01)} &  \posneg{0.05} &  \posneg{0.55} \scriptsize{($\pm$1.82)}&  \posneg{-0.64} \scriptsize{($\pm$1.76)}&  \posneg{-1.50} \scriptsize{($\pm$2.06)}&  \posneg{-0.69} \scriptsize{($\pm$6.26)}&  \posneg{-0.09} \scriptsize{($\pm$1.73)} &  \posneg{-0.47} \\
 & $\Delta$ 500&  \posneg{0.80} \scriptsize{($\pm$0.45)}&  \posneg{0.23} \scriptsize{($\pm$1.50)}&  \posneg{0.64} \scriptsize{($\pm$1.03)}&  \posneg{1.55} \scriptsize{($\pm$0.56)}&  \posneg{1.37} \scriptsize{($\pm$1.65)} &  \posneg{0.92} &  \posneg{0.06} \scriptsize{($\pm$1.97)}&  \posneg{0.14} \scriptsize{($\pm$2.51)}&  \posneg{0.82} \scriptsize{($\pm$2.42)}&  \posneg{6.80} \scriptsize{($\pm$4.06)}&  \posneg{0.64} \scriptsize{($\pm$1.77)} &  \posneg{1.69} \\
\cmidrule(lr){2-14}
\multirow{5}{*}{Llama3-8B (TOL)} & \cellcolor{gray!25} 0 &  65.33 \scriptsize{($\pm$1.53)}&  75.20 \scriptsize{($\pm$1.76)}&  72.70 \scriptsize{($\pm$0.83)}&  66.21 \scriptsize{($\pm$2.53)}&  71.43 \scriptsize{($\pm$2.66)} &  &  64.91 \scriptsize{($\pm$1.33)}&  67.44 \scriptsize{($\pm$3.26)}&  63.57 \scriptsize{($\pm$2.85)}&  50.53 \scriptsize{($\pm$1.88)}&  66.16 \scriptsize{($\pm$1.68)} &  \\
 & $\Delta$ \; 50&  \posneg{-0.25} \scriptsize{($\pm$2.70)}&  \posneg{0.07} \scriptsize{($\pm$1.94)}&  \posneg{-1.38} \scriptsize{($\pm$2.80)}&  \posneg{-0.36} \scriptsize{($\pm$3.99)}&  \posneg{-0.09} \scriptsize{($\pm$3.88)} &  \posneg{-0.40} &  \posneg{-1.33} \scriptsize{($\pm$0.52)}&  \posneg{-1.33} \scriptsize{($\pm$3.80)}&  \posneg{1.82} \scriptsize{($\pm$3.21)}&  \posneg{0.05} \scriptsize{($\pm$13.86)}&  \posneg{0.20} \scriptsize{($\pm$2.22)} &  \posneg{-0.12} \\
 & $\Delta$ 100&  \posneg{-0.26} \scriptsize{($\pm$0.89)}&  \posneg{0.10} \scriptsize{($\pm$3.64)}&  \posneg{-0.34} \scriptsize{($\pm$1.74)}&  \posneg{-0.21} \scriptsize{($\pm$5.46)}&  \posneg{0.46} \scriptsize{($\pm$2.59)} &  \posneg{-0.05} &  \posneg{-0.56} \scriptsize{($\pm$3.28)}&  \posneg{0.44} \scriptsize{($\pm$2.46)}&  \posneg{-0.68} \scriptsize{($\pm$2.85)}&  \posneg{-4.26} \scriptsize{($\pm$1.74)}&  \posneg{1.56} \scriptsize{($\pm$2.56)} &  \posneg{-0.70} \\
 & $\Delta$ 250&  \posneg{0.84} \scriptsize{($\pm$1.60)}&  \posneg{1.03} \scriptsize{($\pm$1.26)}&  \posneg{-1.07} \scriptsize{($\pm$0.66)}&  \posneg{0.08} \scriptsize{($\pm$2.77)}&  \posneg{-1.90} \scriptsize{($\pm$2.68)} &  \posneg{-0.20} &  \posneg{-3.02} \scriptsize{($\pm$4.93)}&  \posneg{-2.52} \scriptsize{($\pm$4.45)}&  \posneg{-0.63} \scriptsize{($\pm$2.15)}&  \posneg{-0.54} \scriptsize{($\pm$4.61)}&  \posneg{1.69} \scriptsize{($\pm$2.04)} &  \posneg{-1.00} \\
 & $\Delta$ 500&  \posneg{0.07} \scriptsize{($\pm$1.33)}&  \posneg{0.56} \scriptsize{($\pm$1.72)}&  \posneg{-0.02} \scriptsize{($\pm$0.35)}&  \posneg{-0.85} \scriptsize{($\pm$3.39)}&  \posneg{0.76} \scriptsize{($\pm$0.17)} &  \posneg{0.10} &  \posneg{-3.10} \scriptsize{($\pm$3.64)}&  \posneg{-1.24} \scriptsize{($\pm$3.05)}&  \posneg{-0.93} \scriptsize{($\pm$4.11)}&  \posneg{5.60} \scriptsize{($\pm$7.38)}&  \posneg{1.32} \scriptsize{($\pm$2.04)} &  \posneg{0.33} \\
\cmidrule(lr){2-14}
\multirow{5}{*}{Llama3-8B (ES)} & \cellcolor{gray!25} 0 &  61.00 \scriptsize{($\pm$6.35)}&  74.93 \scriptsize{($\pm$1.79)}&  74.08 \scriptsize{($\pm$1.81)}&  56.86 \scriptsize{($\pm$10.82)}&  69.84 \scriptsize{($\pm$2.15)} &  &  66.28 \scriptsize{($\pm$1.45)}&  68.44 \scriptsize{($\pm$1.34)}&  65.27 \scriptsize{($\pm$2.46)}&  39.54 \scriptsize{($\pm$8.23)}&  66.83 \scriptsize{($\pm$2.44)} &  \\
 & $\Delta$ \; 50&  \posneg{6.24} \scriptsize{($\pm$0.43)}&  \posneg{0.68} \scriptsize{($\pm$2.37)}&  \posneg{-2.00} \scriptsize{($\pm$2.56)}&  \posneg{10.35} \scriptsize{($\pm$2.17)}&  \posneg{1.34} \scriptsize{($\pm$3.06)} &  \posneg{3.32} &  \posneg{-0.98} \scriptsize{($\pm$1.88)}&  \posneg{-0.63} \scriptsize{($\pm$0.64)}&  \posneg{-0.41} \scriptsize{($\pm$2.51)}&  \posneg{2.05} \scriptsize{($\pm$6.48)}&  \posneg{0.05} \scriptsize{($\pm$2.37)} &  \posneg{0.01} \\
 & $\Delta$ 100&  \posneg{6.71} \scriptsize{($\pm$0.27)}&  \posneg{-0.59} \scriptsize{($\pm$1.73)}&  \posneg{-0.13} \scriptsize{($\pm$2.30)}&  \posneg{8.41} \scriptsize{($\pm$4.63)}&  \posneg{1.35} \scriptsize{($\pm$2.72)} &  \posneg{3.15} &  \posneg{-1.06} \scriptsize{($\pm$0.96)}&  \posneg{0.18} \scriptsize{($\pm$0.97)}&  \posneg{-1.07} \scriptsize{($\pm$1.36)}&  \posneg{5.03} \scriptsize{($\pm$5.33)}&  \posneg{-0.45} \scriptsize{($\pm$2.85)} &  \posneg{0.53} \\
 & $\Delta$ 250&  \posneg{4.96} \scriptsize{($\pm$0.91)}&  \posneg{0.84} \scriptsize{($\pm$0.82)}&  \posneg{1.87} \scriptsize{($\pm$1.77)}&  \posneg{8.68} \scriptsize{($\pm$2.29)}&  \posneg{2.11} \scriptsize{($\pm$3.01)} &  \posneg{3.69} &  \posneg{-2.66} \scriptsize{($\pm$0.76)}&  \posneg{-0.74} \scriptsize{($\pm$0.91)}&  \posneg{-2.09} \scriptsize{($\pm$0.58)}&  \posneg{8.17} \scriptsize{($\pm$5.01)}&  \posneg{-1.18} \scriptsize{($\pm$2.68)} &  \posneg{0.30} \\
 & $\Delta$ 500&  \posneg{7.36} \scriptsize{($\pm$0.71)}&  \posneg{-0.58} \scriptsize{($\pm$2.05)}&  \posneg{1.64} \scriptsize{($\pm$1.46)}&  \posneg{8.96} \scriptsize{($\pm$1.91)}&  \posneg{5.39} \scriptsize{($\pm$2.87)} &  \posneg{4.55} &  \posneg{-3.91} \scriptsize{($\pm$1.31)}&  \posneg{-0.69} \scriptsize{($\pm$1.73)}&  \posneg{-3.52} \scriptsize{($\pm$0.78)}&  \posneg{15.59} \scriptsize{($\pm$2.65)}&  \posneg{-1.70} \scriptsize{($\pm$2.53)} &  \posneg{1.15} \\
\cmidrule(lr){2-14}
\multirow{5}{*}{mBert} & \cellcolor{gray!25} 0 &  54.97 \scriptsize{($\pm$14.70)}&  59.43 \scriptsize{($\pm$16.59)}&  59.98 \scriptsize{($\pm$16.82)}&  40.36 \scriptsize{($\pm$25.63)}&  55.07 \scriptsize{($\pm$19.87)} &  &  54.95 \scriptsize{($\pm$15.31)}&  45.93 \scriptsize{($\pm$23.11)}&  48.12 \scriptsize{($\pm$22.58)}&  46.23 \scriptsize{($\pm$23.77)}&  55.47 \scriptsize{($\pm$22.38)} &  \\
 & $\Delta$ \; 50&  \posneg{2.51} \scriptsize{($\pm$11.86)}&  \posneg{15.14} \scriptsize{($\pm$2.55)}&  \posneg{10.39} \scriptsize{($\pm$1.52)}&  \posneg{9.35} \scriptsize{($\pm$31.00)}&  \posneg{8.07} \scriptsize{($\pm$16.39)} &  \posneg{9.09} &  \posneg{-5.00} \scriptsize{($\pm$26.22)}&  \posneg{20.95} \scriptsize{($\pm$3.40)}&  \posneg{1.90} \scriptsize{($\pm$25.16)}&  \posneg{7.76} \scriptsize{($\pm$17.85)}&  \posneg{12.59} \scriptsize{($\pm$0.72)} &  \posneg{7.64} \\
 & $\Delta$ 100&  \posneg{4.83} \scriptsize{($\pm$11.24)}&  \posneg{13.43} \scriptsize{($\pm$10.41)}&  \posneg{10.55} \scriptsize{($\pm$1.72)}&  \posneg{20.87} \scriptsize{($\pm$11.09)}&  \posneg{14.43} \scriptsize{($\pm$3.07)} &  \posneg{12.82} &  \posneg{-1.09} \scriptsize{($\pm$15.98)}&  \posneg{19.95} \scriptsize{($\pm$3.96)}&  \posneg{11.37} \scriptsize{($\pm$11.18)}&  \posneg{8.74} \scriptsize{($\pm$13.87)}&  \posneg{12.30} \scriptsize{($\pm$2.79)} &  \posneg{10.25} \\
 & $\Delta$ 250&  \posneg{7.92} \scriptsize{($\pm$5.88)}&  \posneg{16.75} \scriptsize{($\pm$2.74)}&  \posneg{12.26} \scriptsize{($\pm$1.24)}&  \posneg{26.58} \scriptsize{($\pm$3.78)}&  \posneg{13.39} \scriptsize{($\pm$5.37)} &  \posneg{15.38} &  \posneg{-0.28} \scriptsize{($\pm$8.65)}&  \posneg{22.43} \scriptsize{($\pm$0.36)}&  \posneg{14.44} \scriptsize{($\pm$8.40)}&  \posneg{12.45} \scriptsize{($\pm$7.04)}&  \posneg{11.74} \scriptsize{($\pm$0.79)} &  \posneg{12.16} \\
 & $\Delta$ 500&  \posneg{-1.26} \scriptsize{($\pm$19.98)}&  \posneg{9.20} \scriptsize{($\pm$11.08)}&  \posneg{10.79} \scriptsize{($\pm$0.75)}&  \posneg{21.57} \scriptsize{($\pm$8.71)}&  \posneg{16.95} \scriptsize{($\pm$8.25)} &  \posneg{11.45} &  \posneg{9.46} \scriptsize{($\pm$3.41)}&  \posneg{22.45} \scriptsize{($\pm$2.73)}&  \posneg{10.18} \scriptsize{($\pm$11.00)}&  \posneg{19.77} \scriptsize{($\pm$2.77)}&  \posneg{13.09} \scriptsize{($\pm$1.10)} &  \posneg{14.99} \\
\bottomrule
\end{tabular}
    
    \end{adjustbox}
        
    \caption{Few-shot CND F1 scores on non-parallel data by model and number of shots. 
    $\Delta$ denotes gains over zero-shot accuracy. Parentheses report standard deviations. 
    All models are fine-tuned on English claims only.}
    \label{tab:few_main_f1}
\end{table*}

Table~\ref{tab:few_main_f1} reports the results for few-shot CND with F1 score as the main metric on non-parallel data.

\subsubsection*{Generalizability}

\begin{table}[t]
\centering
    \begin{adjustbox}{max width=1\linewidth}\centering

    \begin{tabular}{lcccccc}
 & \textbf{en} & \textbf{it} & \textbf{uk} & \textbf{ro} & \textbf{no} & \textbf{az} \\
\toprule
Random articles & 70.40 & 65.49 & 71.43 & 67.27 & 67.19 & 55.77 \\
Random articles (corrected labels) & 91.93 & 80.00 & 83.23 & 79.49 & 93.26 & 70.20 \\
\hdashline
FA baseline & 89.87 & 73.32 & 72.38 & 69.50 & 66.37 & 72.12 \\
\midrule
RA Distribution (0/1) & 14/85 & 11/89 & 6/92 & 7/92 & 8/92 & 2/97 \\
\end{tabular}

    \end{adjustbox}
        
\caption{Monolingual CND F1 scores on the Random Articles (RA) test set ($n=100$ per language) using Llama~3 (ES). Due to under-referencing in random articles, we manually corrected the labels.}
    \label{tab:general_f1}
\end{table}

\end{document}